\documentclass[sigconf]{acmart}
\usepackage{algorithm}
\usepackage{algpseudocode}
\usepackage{color}
\usepackage{threeparttable}
\usepackage{multirow}

\usepackage{balance}

\newcommand{\etal}{\textit{et al}.}
\newcommand{\ie}{\textit{i}.\textit{e}.}
\newcommand{\eg}{\textit{e}.\textit{g}.}
\newcommand{\etc}{\textit{etc}}

\AtBeginDocument{%
  \providecommand\BibTeX{{%
    \normalfont B\kern-0.5em{\scshape i\kern-0.25em b}\kern-0.8em\TeX}}}

\copyrightyear{2022}
\acmYear{2022}
\setcopyright{acmcopyright}
\acmConference[MM '22]{Proceedings of the 30th ACM International Conference on Multimedia}{October 10--14, 2022}{Lisboa, Portugal}
\acmBooktitle{Proceedings of the 30th ACM International Conference on Multimedia (MM '22), October 10--14, 2022, Lisboa, Portugal}
\acmPrice{15.00}
\acmISBN{978-1-4503-XXXX-X/22/10}

\begin{document}

\title{AesUST: Towards Aesthetic-Enhanced Universal Style Transfer}

\author{Zhizhong Wang}
\authornote{Both authors contributed equally to this research.}
\author{Zhanjie Zhang}
\authornotemark[1]
\email{endywon@zju.edu.cn}
\affiliation{%
	\institution{Zhejiang University}
	\city{}
	\country{}
}

\author{Lei Zhao}
\authornote{Corresponding authors.}
\email{cszhl@zju.edu.cn}
\affiliation{%
	\institution{Zhejiang University}
	\city{}
	\country{}
}

\author{Zhiwen Zuo}
\email{zzwcs@zju.edu.cn}
\affiliation{%
	\institution{ Zhejiang University}
	\city{}
	\country{}
}

\author{Ailin Li}
\email{liailin@zju.edu.cn}
\affiliation{%
	\institution{Zhejiang University}
	\city{}
	\country{}
}

\author{Wei Xing}
\authornotemark[2]
\email{wxing@zju.edu.cn}
\affiliation{%
	\institution{Zhejiang University}
	\city{}
	\country{}
}

\author{Dongming Lu}
\email{ldm@zju.edu.cn}
\affiliation{%
	\institution{Zhejiang University}
	\city{}
	\country{}
}

\renewcommand{\shortauthors}{Zhizhong Wang et al.}
		





\begin{abstract}
  Recent studies have shown remarkable success in universal style transfer which transfers arbitrary visual styles to content images. However, existing approaches suffer from the aesthetic-unrealistic problem that introduces disharmonious patterns and evident artifacts, making the results easy to spot from real paintings. To address this limitation, we propose AesUST, a novel \textbf{Aes}thetic-enhanced \textbf{U}niversal \textbf{S}tyle \textbf{T}ransfer approach that can generate aesthetically more realistic and pleasing results for arbitrary styles. Specifically, our approach introduces an aesthetic discriminator to learn the universal human-delightful aesthetic features from a large corpus of artist-created paintings. Then, the aesthetic features are incorporated to enhance the style transfer process via a novel \textbf{Aes}thetic-aware \textbf{S}tyle-\textbf{A}ttention (AesSA) module. Such an AesSA module enables our AesUST to efficiently and flexibly integrate the style patterns according to the global aesthetic channel distribution of the style image and the local semantic spatial distribution of the content image. Moreover, we also develop a new two-stage transfer training strategy with two aesthetic regularizations to train our model more effectively, further improving stylization performance. Extensive experiments and user studies demonstrate that our approach synthesizes aesthetically more harmonious and realistic results than state of the art, greatly narrowing the disparity with real artist-created paintings. Our code is available at https://github.com/EndyWon/AesUST.
\end{abstract}

\ccsdesc[500]{Applied computing~Fine arts}
\ccsdesc[300]{Computing methodologies~Rendering}
\ccsdesc[300]{Computing methodologies~Image manipulation}

\keywords{image style transfer, aesthetic, universal, harmonious, realistic}


\maketitle

\section{Introduction}
\label{intro}
The task of style transfer is to transfer the artistic style from a reference image to a content image, \eg, transferring the style of Vincent Van Gogh’s {\em Sunflower} to an everyday photograph. This task has experienced significant improvements following the seminal work of Gatys~\etal~\cite{gatys2015neural,gatys2016image,gatys2015texture}, with aspects ranging from efficiency~\cite{johnson2016perceptual,ulyanov2016texture,li2016precomputed}, quality~\cite{li2016combining,sheng2018avatar,park2019arbitrary,wang2021evaluate,lin2021drafting}, generalization~\cite{chen2017stylebank,dumoulin2017learned,huang2017arbitrary,li2017universal,jing2020dynamic}, diversity~\cite{li2017diversified,ulyanov2017improved,wang2020diversified}, and user control~\cite{champandard2016semantic,gatys2017controlling,kolkin2019style}.

\begin{figure*}
	\centering
	\setlength{\tabcolsep}{0.02cm}
	\renewcommand\arraystretch{0.4}
	\begin{tabular}{cccccccc}
		
		\includegraphics[width=0.12\linewidth]{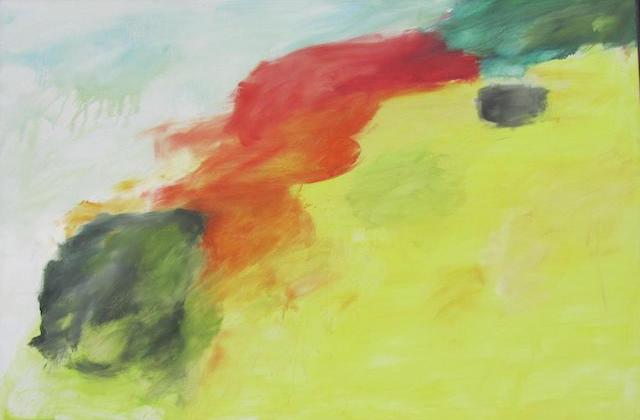}&
		\includegraphics[width=0.12\linewidth]{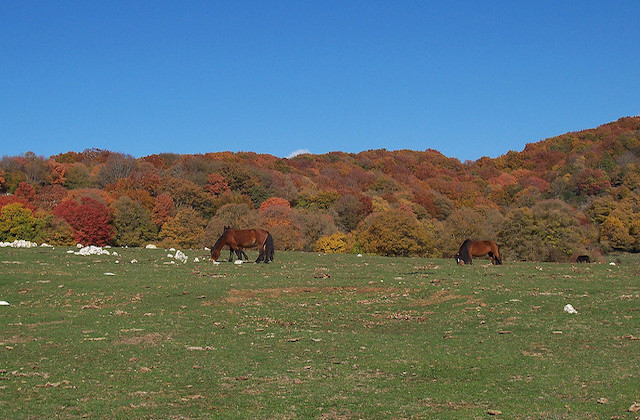}&
		\includegraphics[width=0.12\linewidth]{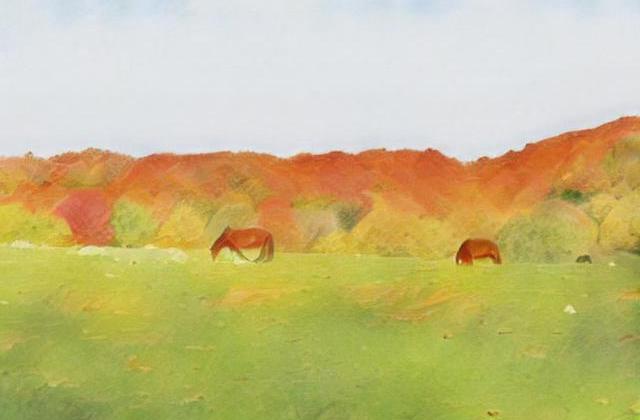} &
		\includegraphics[width=0.12\linewidth]{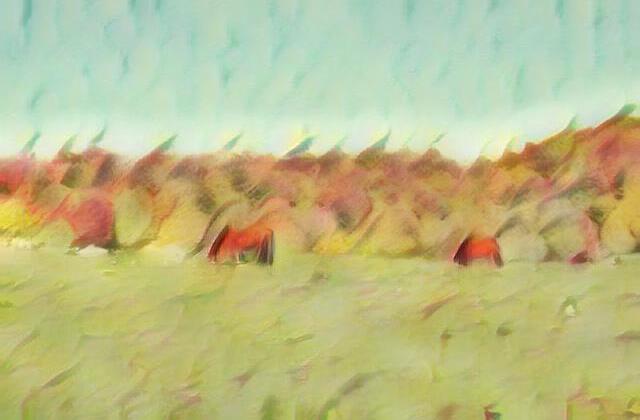}&
		\includegraphics[width=0.12\linewidth]{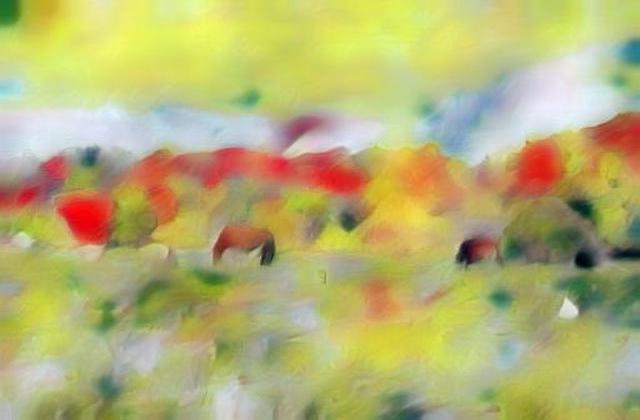}&
		\includegraphics[width=0.12\linewidth]{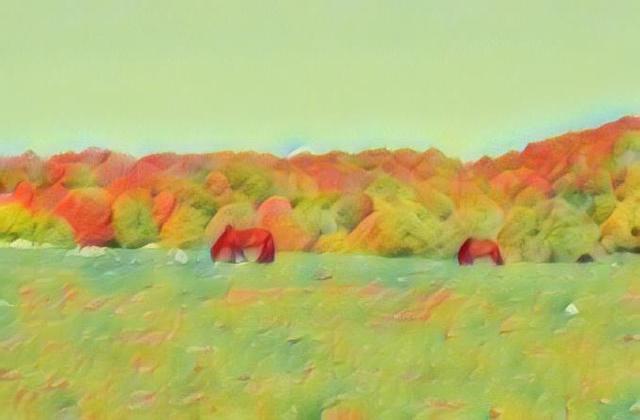}&
		\includegraphics[width=0.12\linewidth]{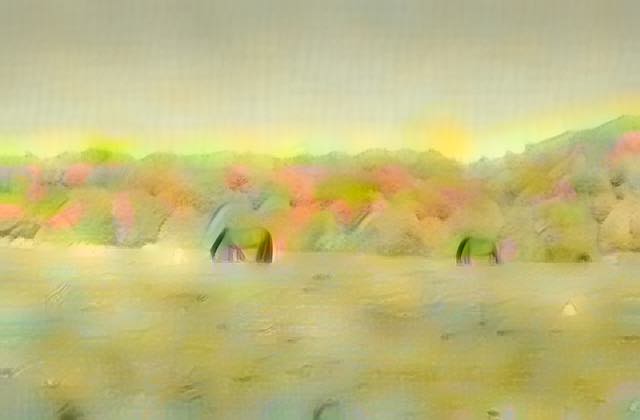}&
		\includegraphics[width=0.12\linewidth]{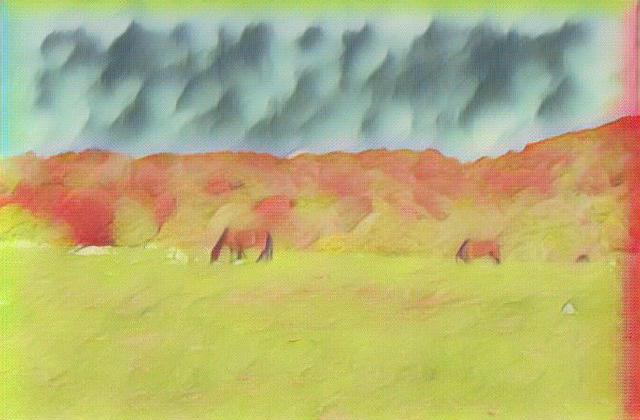}
		\\
		
		\includegraphics[width=0.12\linewidth]{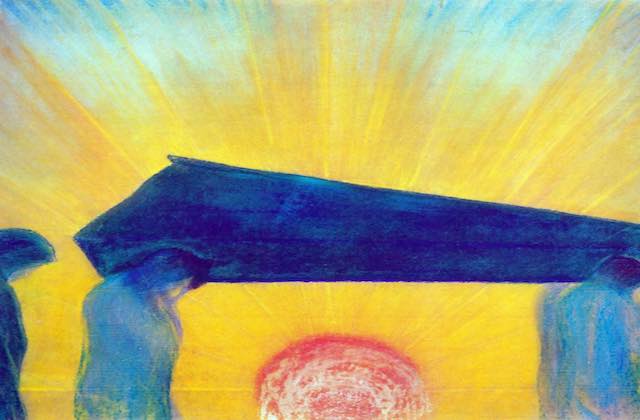}&
		\includegraphics[width=0.12\linewidth]{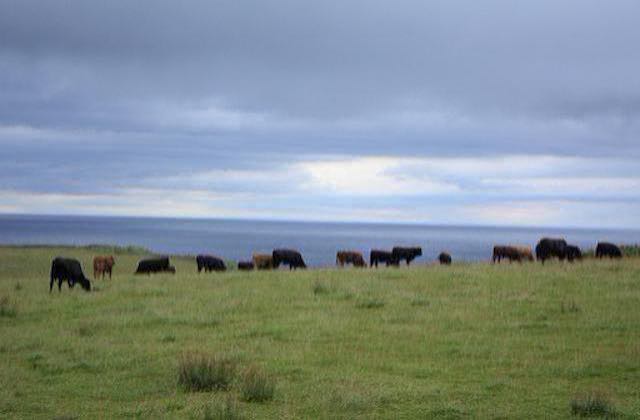}&
		\includegraphics[width=0.12\linewidth]{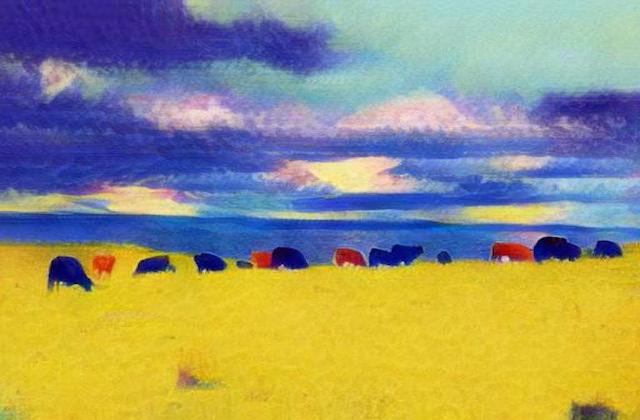} &
		\includegraphics[width=0.12\linewidth]{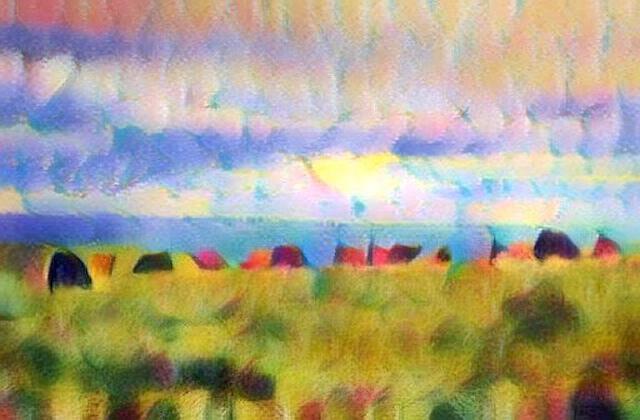}&
		\includegraphics[width=0.12\linewidth]{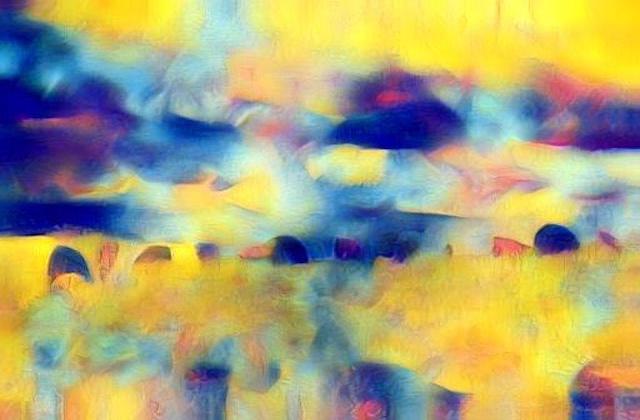}&
		\includegraphics[width=0.12\linewidth]{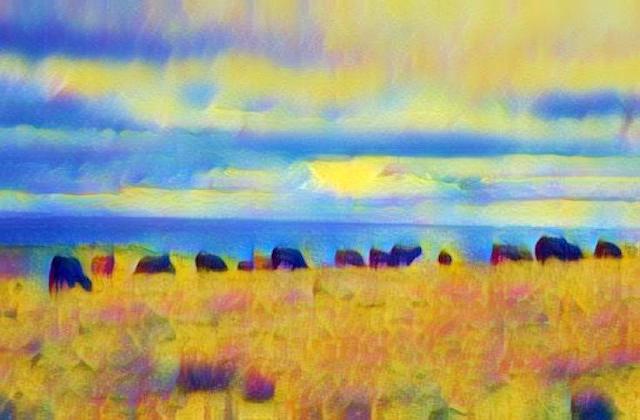}&
		\includegraphics[width=0.12\linewidth]{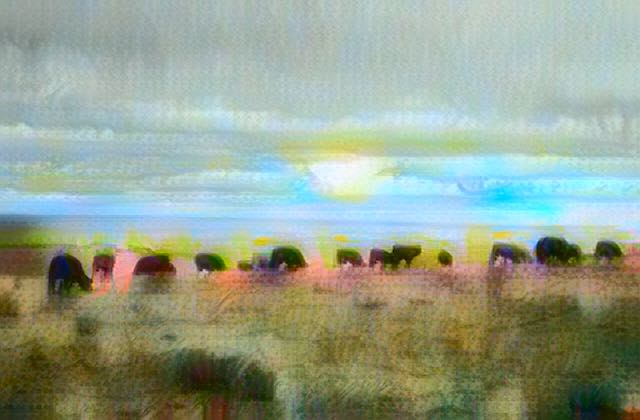}&
		\includegraphics[width=0.12\linewidth]{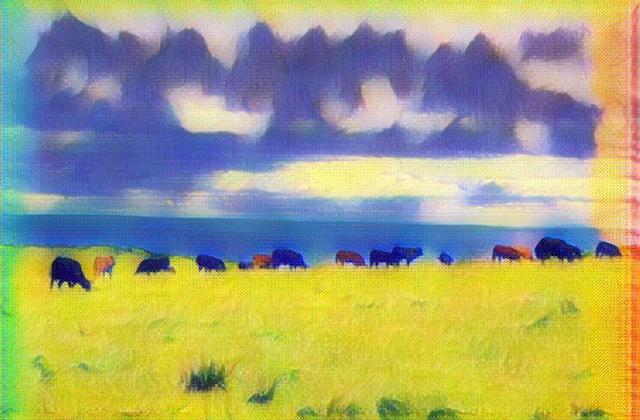}
		\\
			
		Style & Content & {\bf Ours}   &  AdaIN & WCT &  LST & AAST &  ArtFlow
		
		\\
		
		\includegraphics[width=0.12\linewidth]{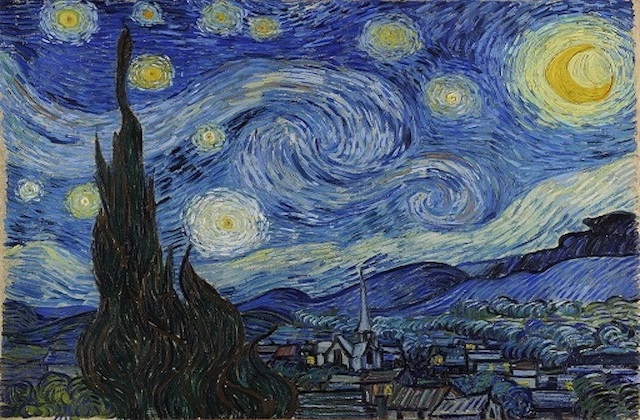}&
		\includegraphics[width=0.12\linewidth]{figs/teaser/local1/c}&
		\includegraphics[width=0.12\linewidth]{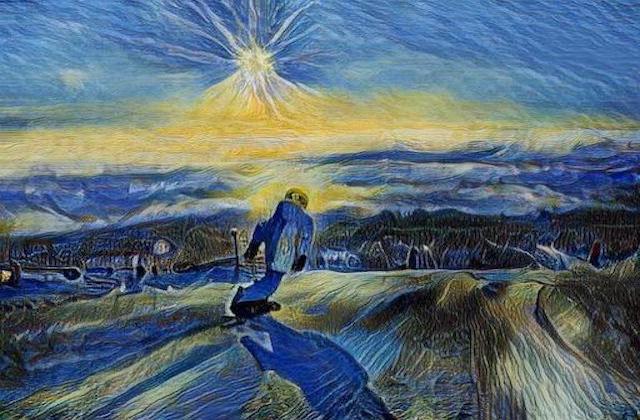} &
		\includegraphics[width=0.12\linewidth]{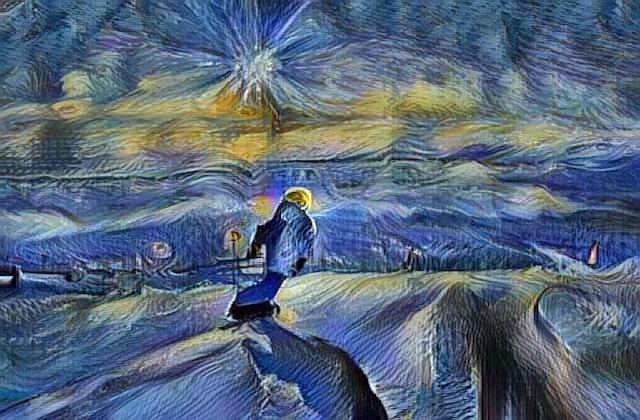}&
		\includegraphics[width=0.12\linewidth]{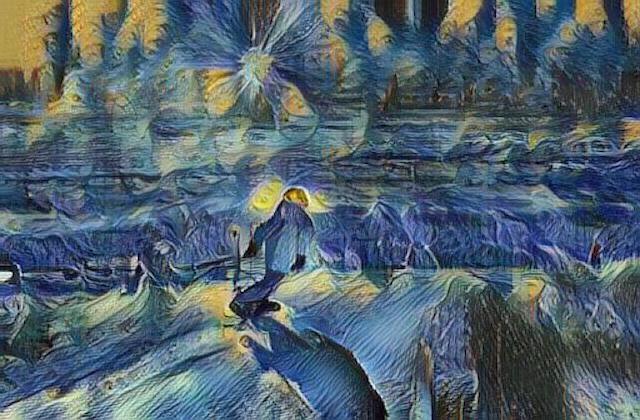}&
		\includegraphics[width=0.12\linewidth]{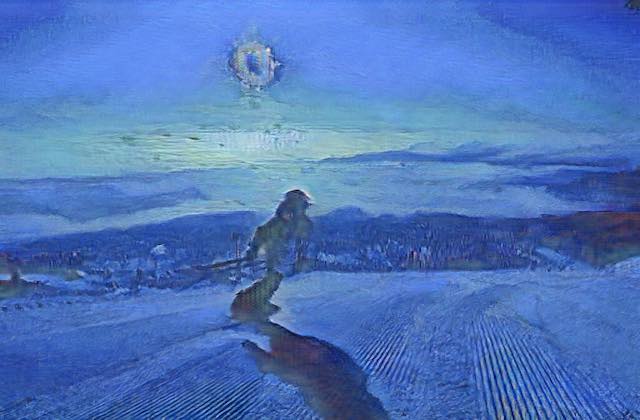}&
		\includegraphics[width=0.12\linewidth]{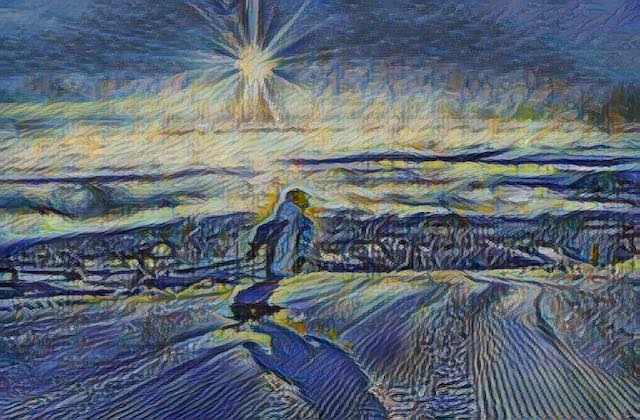}&
		\includegraphics[width=0.12\linewidth]{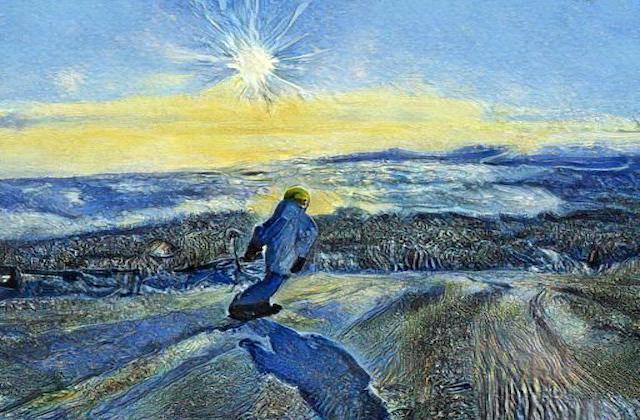}
		\\
		
		\includegraphics[width=0.12\linewidth]{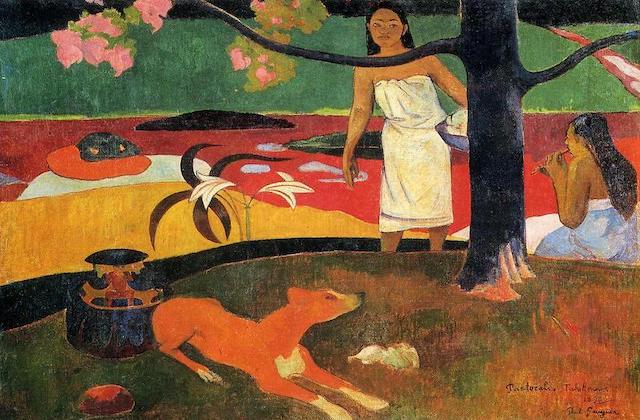}&
		\includegraphics[width=0.12\linewidth]{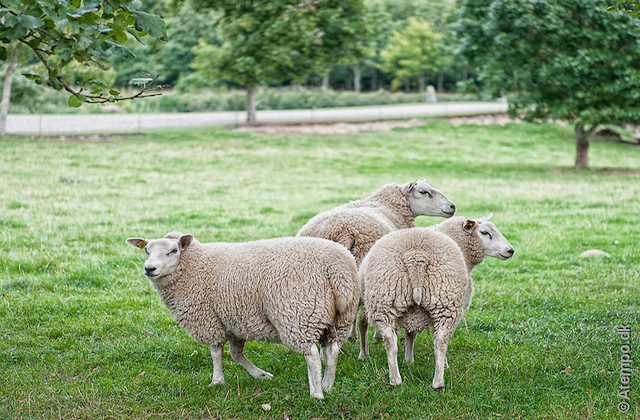}&
		\includegraphics[width=0.12\linewidth]{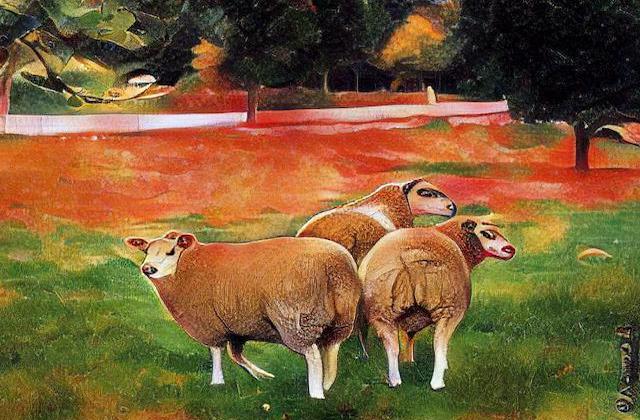} &
		\includegraphics[width=0.12\linewidth]{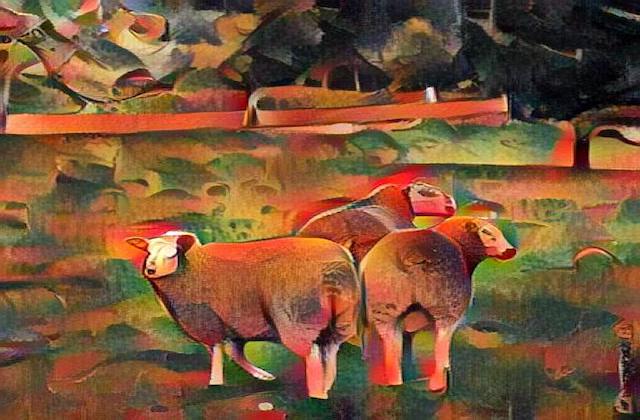}&
		\includegraphics[width=0.12\linewidth]{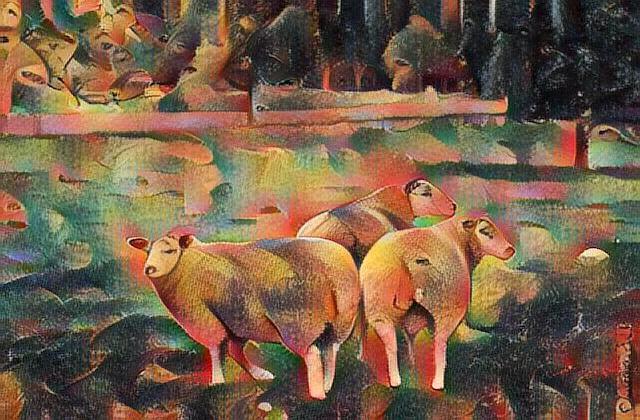}&
		\includegraphics[width=0.12\linewidth]{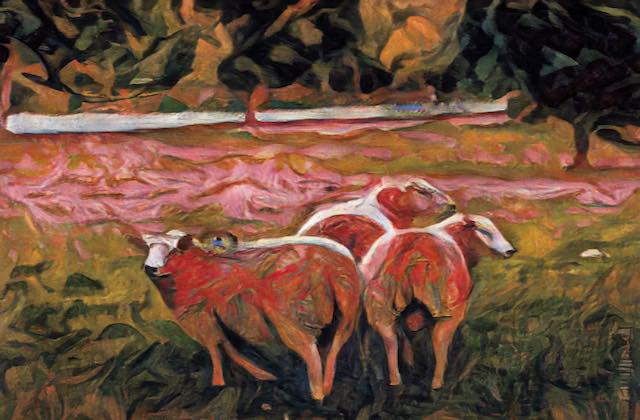}&
		\includegraphics[width=0.12\linewidth]{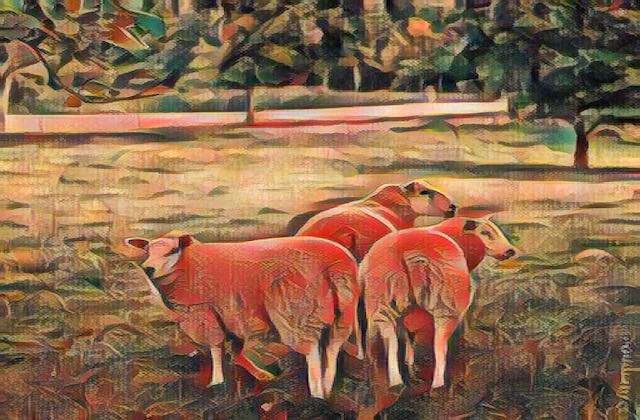}&
		\includegraphics[width=0.12\linewidth]{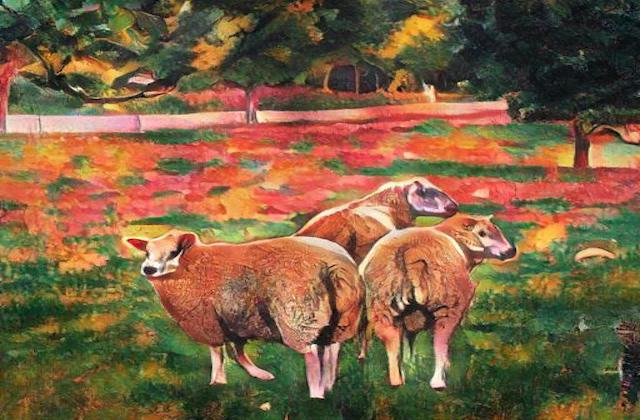}
		\\
			
		Style & Content  & {\bf Ours} &  SANet & MAST &  TPFR  & AdaAttN &  IECAST  
		
	\end{tabular}
    \vspace{-1em}
	\caption{{\bf Exemplar stylization results.} Existing UST approaches (column 4-8) suffer from the aesthetic-unrealistic problem that introduces disharmonious patterns and evident artifacts, making their results easily distinguishable from real paintings. Our method (column 3) synthesizes aesthetically more realistic and pleasing results, greatly narrowing the disparity with real artist-created paintings. }
	
	\label{fig:teaser}
	\vspace{-1em}
\end{figure*} 

As a central problem of style transfer, universal style transfer (UST) seeks to achieve generalization, quality, and efficiency simultaneously~\cite{li2017universal}. Depending on the ways of manipulating content and style features, the existing UST approaches can be roughly divided into two categories, \ie, global statistics-based and local patch-based~\cite{jing2019neural}. As representatives of the former, AdaIN~\cite{huang2017arbitrary}, WCT~\cite{li2017universal}, LST~\cite{li2019learning}, AAST~\cite{hu2020aesthetic}, and ArtFlow~\cite{an2021artflow} transform the content features to match the second-order global statistics of style features. As representatives of the latter, SANet~\cite{park2019arbitrary}, MAST~\cite{deng2020arbitrary}, TPFR~\cite{svoboda2020two}, AdaAttN~\cite{liu2021adaattn}, and IECAST~\cite{chen2021artistic} utilize patch-based decorators or attention mechanisms to locally fuse style features into content features. While achieving favorable results, fast speeds, and good generalizations, these approaches often produce disharmonious patterns and evident artifacts, making their results easily distinguishable from real paintings (see Fig.~\ref{fig:teaser}). On the contrary, an excellent stylized artwork that can be faked as real should be as realistic as possible in human aesthetics, \ie, its artistic characteristics, such as colors, strokes, tones, textures, \etc, all exist in harmony and are visually pleasing. Therefore, we define this problem as the aesthetic-unrealistic problem and provide the following analyses:

For global statistics-based approaches~\cite{huang2017arbitrary,li2017universal,li2019learning,hu2020aesthetic,an2021artflow}, the aesthetic-unrealistic problem appears since the global statistics match the disordered style patterns without considering their local distributions~\cite{gatys2016image}. Thus, they may often integrate messy textures into the content target. While for local patch-based approaches~\cite{park2019arbitrary,deng2020arbitrary,svoboda2020two,liu2021adaattn,chen2021artistic}, the style patterns are integrated to greedily match the local structure distribution of the content image, which is prone to introduce artifacts and disharmonious patterns~\cite{zhang2019multimodal}. Therefore, the leading cause of the aesthetic-unrealistic problem lies in the inappropriate or insufficient considerations on the integration of style patterns.

Based on the above analyses, the current question has been transformed to {\em how to integrate style patterns in an aesthetically realistic and pleasing way?} However, the subjective characteristics of the problem bring great challenges. There are no absolute standards for measuring the aesthetic quality of a painting~\cite{li2009aesthetic}. Different artists can have very different ideas towards aesthetics. Therefore, to sidestep such a dilemma, we explore another perspective to address the question. Our insight is based on Immanuel Kant's famous saying, {\em art is both subjective and \textbf{universal}}. Each individual person may have his/her own taste, but a given piece of art can also appeal to a large number of people across cultures and history~\cite{iigaya2021aesthetic}. We argue that the artist-created paintings may share some notable universal characteristics of human-delightful aesthetics (\eg, harmonious patterns and few artifacts), and we can learn and leverage these aesthetics to produce aesthetically more realistic and pleasing results. That is to say, we can utilize these learned universal aesthetics to guide the integration of style patterns, without defining an elaborate aesthetic criterion.

Motivated by the arguments above, we propose a novel {\em \textbf{Aes}thetic-enhanced \textbf{U}niversal \textbf{S}tyle \textbf{T}ransfer (\textbf{AesUST})} approach that aims to generate aesthetically more realistic and pleasing results for arbitrary styles. Our AesUST is inspired by the recent great success of Generative Adversarial Networks (GANs)~\cite{goodfellow2014generative} in generating domain-realistic results. Specifically, we introduce an aesthetic discriminator to learn the universal human-delightful aesthetic characteristics from the domain of a large corpus of artist-created paintings. These paintings are created by numerous artists, thus containing some universal artist-independent and human-delightful aesthetics. The aesthetic discriminator plays two roles here: (i) an opponent that teaches the generator to produce realistic painting-like results via playing a min-max game, (ii) a feature extractor that extracts the aesthetic features to enhance the style transfer process. Notably, we also propose a novel {\em \textbf{Aes}thetic-aware \textbf{S}tyle-\textbf{A}ttention (\textbf{AesSA})} module to incorporate the aesthetic features with the content and style features. Such an AesSA module enables our approach to efficiently and flexibly integrate the style patterns according to the global aesthetic channel distribution of the style image and the local semantic spatial distribution of the content image, thus helping produce aesthetically more realistic and pleasing results. A remaining issue of our approach is how to perform adversarial training with traditional style transfer training. To this end, we develop a new two-stage transfer training strategy with two aesthetic regularizations to train our model more effectively, further improving stylization performance. We conduct extensive experiments as well as user studies to demonstrate the effectiveness and superiority of our approach. Compared to the state-of-the-art (SOTA) UST algorithms, our AesUST can synthesize aesthetically more harmonious and realistic results, greatly narrowing the disparity with real artist-created paintings.

In summary, our contributions are threefold:

\begin{itemize}

	\item We reveal the aesthetic-unrealistic problem in existing UST algorithms, and propose a novel {\em aesthetic-enhanced universal style transfer} approach, \ie, {\em AesUST}, to generate aesthetically more realistic and pleasing results for arbitrary styles.
	
	\item We introduce a novel {\em aesthetic-aware style-attention (AesSA)} module to achieve global aesthetic-guided and local structure-guided style integration.  
	
	\item A new two-stage transfer training strategy with two aesthetic regularizations is also developed to train the networks more effectively, further improving stylization performance. 
\end{itemize}

\begin{figure*}[t]
	\centering
	\includegraphics[width=0.9\linewidth]{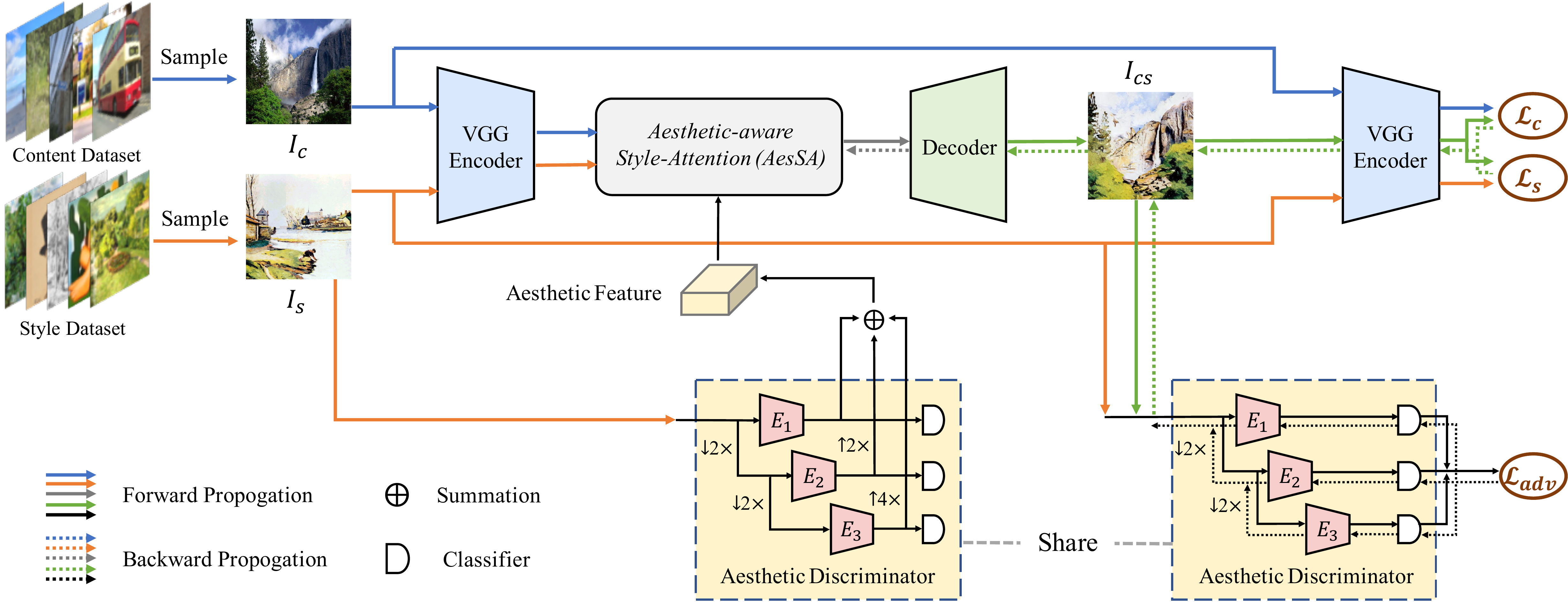}
	\caption{{\bf Overview of our proposed AesUST.} Note that we apply a two-stage transfer training strategy with a transfer learning fashion: (1) At stage I, the aesthetic discriminator only acts as the discriminator, and we pre-train the generator (AesSA module and decoder) using VGG features only. (2) At stage II, the aesthetic discriminator acts as both the discriminator and feature extractor, and we fine-tune the generator using the aesthetic features extracted from the aesthetic discriminator. The aesthetic discriminator is trained when maximizing the adversarial loss $\mathcal{L}_{adv}$ and fixed when minimizing other losses like content loss $\mathcal{L}_{c}$ and style loss $\mathcal{L}_{s}$. For clarity, we omit some stage-specific losses here, \ie, identity loss~\cite{park2019arbitrary} for stage I, and our proposed aesthetic regularization losses (see Fig.~\ref{fig:AR}) for stage II.
	}
	\label{fig:overview}
	\vspace{-1em}
\end{figure*}

\vspace{-0.5em}
\section{Related Work}
{\bf Global Statistics-based Methods.} The seminal work of Gatys~\etal~\cite{gatys2015texture,gatys2016image} opened up the era of neural style transfer~\cite{jing2019neural}. They used the global statistics, \ie, the correlations between features extracted from a pre-trained Deep Convolutional Neural Network (DCNN), to represent the style of an image. Albeit the stunning results, their method incurs a rather slow speed in practical usage. To address this issue, Johnson~\etal~\cite{johnson2016perceptual} and Ulyanov~\etal~\cite{ulyanov2016texture} achieved fast style transfer, but the model is limited to a pre-defined style. Thereafter, many successors were advanced to improve the performance in different aspects, including multi-style transfer~\cite{chen2017stylebank,dumoulin2017learned}, universal style transfer~\cite{huang2017arbitrary,li2017universal,li2019learning,lu2019closed,jing2020dynamic,cheng2021style,lu2022universal}, diversified style transfer~\cite{ulyanov2017improved,li2017diversified,wang2020diversified}, unbiased style transfer~\cite{an2021artflow}, and domain-aware style transfer~\cite{hong2021domain}, \etc. Unlike these methods, the proposed approach aims to achieve aesthetic-enhanced style transfer, which considers the universal human-delightful aesthetics to improve the quality of universal style transfer.

{\bf Local Patch-based Methods.} Li and Wand~\cite{li2016combining,li2016precomputed} first combined Markov Random Fields (MRFs) and DCNNs to achieve local patch-based neural style transfer. They represented the style patterns by a set of neural patches and greedily integrated them to match the local structure prior of the content image. Later, Chen and Schmidt~\cite{chen2016fast} proposed style-swap for fast patch-based style transfer. Based on it, Sheng~\etal~\cite{sheng2018avatar} designed Avatar-Net for zero-shot multi-scale style decoration. Park and Lee~\cite{park2019arbitrary}, Deng~\etal~\cite{deng2020arbitrary}, and Liu~\etal~\cite{liu2021adaattn} utilized attention mechanism to integrate style patterns into content features. Other extensions include image analogy~\cite{liao2017visual,gu2018arbitrary}, semantic style transfer~\cite{champandard2016semantic,wang2020glstylenet,wang2022texture}, diversified style transfer~\cite{wang2022divswapper}, and multi-modal style transfer~\cite{zhang2019multimodal}, \etc. While these methods can integrate semantically more coherent style patterns into the content features, the greedy structure-guided distributions are prone to introduce artifacts and disharmonious patterns. By contrast, our proposed approach considers both structure-guided and aesthetic-guided distributions to integrate the style patterns, leading to aesthetically more harmonious and realistic results.

{\bf Aesthetic-related Methods.} While there have been some efforts~\cite{li2009aesthetic,aydin2014automated,wang2021evaluate} to explore the automatic evaluation of aesthetic quality, they are confined to the limited aspects, and so far, no unified criterion has been reached. Similar to our method, Sanakoyeu~\etal~\cite{sanakoyeu2018style}, Kotovenko~\etal~\cite{kotovenko2019content,kotovenko2019content1}, Chen~\etal~\cite{chen2021dualast,chen2021diverse}, and Zuo~\etal~\cite{zuo2022style} also used GAN-based frameworks for style transfer. However, their discriminators are trained to distinguish the aesthetics of a specific artist, \eg, Claude Monet, while ours captures the artist-independent aesthetics. Moreover, their models can only transfer the style of a pre-defined artist, while ours can achieve universal style transfer. Recently, Svoboda~\etal~\cite{svoboda2020two} provided a two-stage peer-regularization model based on graph attention and GAN for arbitrary style transfer, but it suffers from the style-uncontrollable problem~\cite{chen2021dualast}. Chen~\etal~\cite{chen2021artistic} used internal-external learning and contrastive learning with GAN to improve the performance of SANet~\cite{park2019arbitrary}, but it may produce implausible stylizations for complex textures and sometimes still introduces noticeable artifacts. Furthermore, Hu~\etal~\cite{hu2020aesthetic} proposed aesthetic-aware style transfer, but their aesthetics are defined as a combination of only color and texture. By contrast, we provide a more general definition of aesthetics, \ie, the universal human-delightful characteristics learned from a large corpus of artist-created paintings.

\vspace{-0.5em}
\section{Approach}
This part first introduces the overall pipeline of our proposed approach in Sec.~\ref{pipe}, and then follows it up by providing the details of each component in Sec.~\ref{compo}. The two-stage transfer training strategy is specified in Sec.~\ref{two-stage}.

\vspace{-0.5em}
\subsection{Overview}
\label{pipe}

Let $\Phi_c$ and $\Phi_s$ be the datasets of content photographs and real artist-created paintings, respectively. Our goal is to learn a style transformation that can transfer the style of an arbitrary painting $I_s\in \Phi_s$ to a content target $I_c\in \Phi_c$. The key insight is utilizing the aesthetic characteristics learned from $\Phi_s$ to enhance the style transformation to synthesize aesthetically more realistic and pleasing results. To achieve this goal, we propose a novel aesthetic-enhanced universal style transfer framework, termed AesUST.

As shown in Fig.~\ref{fig:overview}, our AesUST consists of four main components: (1) A pre-trained VGG~\cite{simonyan2014very} encoder $E_{vgg}$ that projects images into multi-level feature embeddings. (2) An aesthetic-aware style-attention module $AesSA$ that appropriately integrates the style patterns into the content features under the guidance of aesthetic and structure distributions. (3) A decoder network $D$ that recovers the feature embeddings to the stylized images. (4) An aesthetic discriminator $\mathcal{D}_a$ that teaches the decoder to generate realistic painting-like results and also extracts the aesthetic features to enhance the style transfer process. The overall pipeline\footnote{\noindent The pipeline will vary slightly at different training stages in later Sec.~\ref{two-stage}.} is as follows:
\begin{enumerate}
	\renewcommand{\labelenumi}{(\theenumi)}
	\vspace{-0.3em}
	\setlength{\itemsep}{2pt}
	\setlength{\parsep}{0pt}
	\setlength{\partopsep}{0pt}
	\setlength{\parskip}{0pt}
	
	\item From a content image $I_c$ and style image $I_s$ pair, we first extract the VGG content feature $F_c:=E_{vgg}(I_c)$ and style feature $F_s:=E_{vgg}(I_s)$ at a certain layer (\eg, $Relu\_4\_1$) of the encoder $E_{vgg}$.
	
	\vspace{-0.2em}
	\item We then extract the aesthetic feature of the style image $I_s$ using the aesthetic discriminator $\mathcal{D}_a$, denoted as $F_a:=\mathcal{D}_a(I_s)$.
	
	\vspace{-0.2em}
	\item After obtaining the feature maps $F_c$, $F_s$, and $F_a$, we feed them into an $AesSA$ module, producing the output feature map $F_{cs}:=AesSA(F_c, F_s, F_a)$.
	
	\vspace{-0.2em}
	\item Finally, the stylized output image $I_{cs}$ is synthesized by feeding $F_{cs}$ into the decoder $D$, \ie, $I_{cs}:=D(F_{cs})$.
\end{enumerate}

\vspace{-0.5em}
\subsection{Component Details}
\label{compo}

{\bf Encoder-Decoder Module.} Following~\cite{huang2017arbitrary}. We employ the pre-trained VGG-19 network~\cite{simonyan2014very} as our encoder $E_{vgg}$ and fix it all the time. The decoder $D$ is trainable, which mostly mirrors the encoder, but with all pooling layers replaced by the nearest up-sampling.

{\bf Aesthetic Discriminator.} Inspired by the recent remarkable success of GANs~\cite{goodfellow2014generative} in generating domain-realistic results, we introduce an aesthetic discriminator $D_a$ to learn the universal human-delightful aesthetic features from a large corpus of artist-created paintings. Basically, the aesthetic discriminator $D_a$ is to distinguish between the fake paintings and the real paintings. To achieve so, $D_a$ should perform a sort of encodings of the input paintings before it can judge what paintings are real and what are fake~\cite{radford2015unsupervised,chen2020reusing}. These encodings, therefore, can be deemed as the notable universal aesthetic features that determine the realism and delicacy of the paintings. Motivated by this, we contend two roles of the aesthetic discriminator: (i) an opponent that teaches the generator to produce realistic painting-like results via playing a min-max game, and (ii) a feature extractor that extracts the aesthetic features to enhance the style transfer process. Such designations bring two main advantages: (1) The framework can be more informative since the adversarial training helps it fuse the domain knowledge of the artist-created paintings. (2) The style transfer networks can be trained more effectively, as the aesthetic discriminator provides additional guidance for better style transfer.

In detail, our aesthetic discriminator follows the multi-scale architecture of~\cite{wang2018high}, which consists of three identical encoders ($E_1$, $E_2$, and $E_3$) and classifiers ($\mathcal{C}_1$, $\mathcal{C}_2$, and $\mathcal{C}_3$). Specifically, we first downsample the real and fake paintings by a factor of 2 and 4 to create an image pyramid of 3 scales. Then, each pair of encoders and classifiers (\eg, $E_1$ and $\mathcal{C}_1$) is used to differentiate real and fake paintings at one scale. This multi-scale mechanism helps our aesthetic discriminator capture the aesthetic features from different scales, and also guide the style transfer networks to produce more delicate results with coarse-to-fine style patterns. The multi-scale aesthetic discriminator is illustrated in Fig.~\ref{fig:overview}, and the detailed architecture can be found in {\em appendix}. We sum the encodings of $E_1$, $E_2$, and $E_3$ to obtain the aesthetic features as follows:
\begin{equation}
	F_a := \mathcal{D}_a(I_s) := E_1(I_s) \oplus E_2 (I_s^{\downarrow^2}) ^{\uparrow^2} \oplus E_3 (I_s^{\downarrow^4}) ^{\uparrow^4},
\end{equation}
where $\oplus$ denotes element-wise summation, $\downarrow^i$ ($\uparrow^i$) denotes down-sampling (up-sampling) operation with factor $i$.

\begin{figure}[t]
	\centering
	\includegraphics[width=1\linewidth]{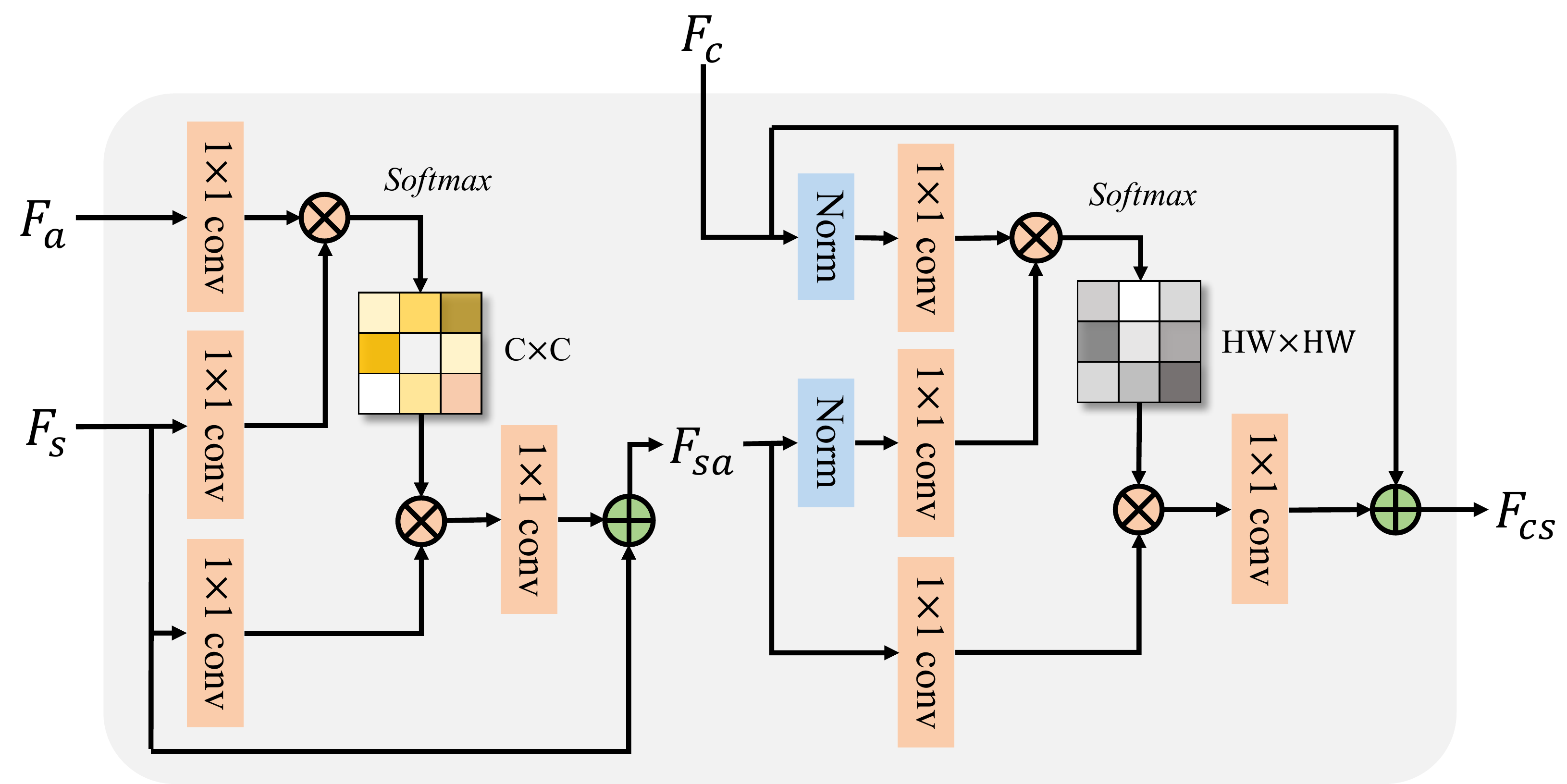}
	\vspace{-1em}
	\caption{{\bf Aesthetic-aware style-attention (AesSA) module.} $F_c$, $F_s$, and $F_a$ are content, style, and aesthetic features, respectively. ``$Norm$'' denotes the mean-variance channel-wise normalization.
	}
	\label{fig:aesSA}
	\vspace{-1em}
\end{figure}

{\bf Aesthetic-aware Style-Attention (AesSA) Module.} To incorporate the aesthetic features with the VGG content and style features, we propose the AesSA module, which can adaptively integrate the style patterns into the content features via considering both global aesthetic and local structure features with the attention mechanism. As shown in Fig.~\ref{fig:aesSA}, AesSA works in two steps: (1) globally enhancing the style features according to the aesthetic channel distributions of the aesthetic features, and (2) locally integrating the enhanced style features according to the semantic spatial distributions of the content features.

{\em Step I: Global aesthetic-guided style enhancement.} We first enhance the style features with the aesthetic features extracted from the style images. Inspired by~\cite{gatys2016image,fu2019dual,deng2020arbitrary} that the channel-wise inner product between the vectorized features can represent the global style well, and the channel attention can effectively improve the feature representation; we introduce the aesthetic attention to globally enhance the VGG style features according to the channel distributions of the aesthetic features. Specifically, given the style feature $F_s \in \mathbb{R}^{C\times H_s \times W_s}$ from VGG and the aesthetic feature $F_a \in \mathbb{R}^{C\times H_s \times W_s}$ from discriminator (where $H_s$ and $W_s$ are the height and width, $C$ is the number of channels), we first obtain their transformed and vectorized features:
\begin{equation}
	\begin{aligned}
		&\hat{F}_a := \Gamma(f_a (F_a)) \in \mathbb{R}^{C\times H_sW_s}, \\
		&\hat{F}_s^1 := \Gamma(f_s^1 (F_s)) \in \mathbb{R}^{C\times H_sW_s}, \\
		&\hat{F}_s^2 := \Gamma(f_s^2 (F_s)) \in \mathbb{R}^{C\times H_sW_s}, 
	\end{aligned}
\end{equation}
where $\Gamma$ denotes the feature vectorization operation, $f_a$, $f_s^1$, and $f_s^2$ are $1\times1$ learnable convolutions. 

Then, we calculate the aesthetic attention between $\hat{F}_a$ and $\hat{F}_s^1$ as follows:
\begin{equation}
	A_a := Softmax (\hat{F}_a \otimes (\hat{F}_s^1)^T) \in \mathbb{R}^{C\times C},
\end{equation}
where $\otimes$ is matrix multiplication, $T$ is transpose operation.

Finally, we enhance the style feature $F_s$ through a matrix multiplication and an element-wise summation:
\begin{equation}
	F_{sa} := f_{out}^1 (\hat{\Gamma}(A_a \otimes \hat{F}_s^2))  \oplus F_s \in \mathbb{R}^{C\times H_s \times W_s}  ,
\end{equation}
where $\hat{\Gamma}$ denotes the reverse operation of $\Gamma$ that reshapes the vectorized features to the original size, $f_{out}^1$ is $1\times1$ learnable convolution.

{\em Step II: Local structure-guided style integration.} After enhancing the style features under the aesthetic guidance, we want to integrate them into the content features, so as to achieve aesthetic-aware style transfer. Following SANet~\cite{park2019arbitrary}, we utilize the style attention to locally integrate the enhanced style patterns according to the structure distributions of the content features. Given the aesthetic-enhanced style feature $F_{sa}\in \mathbb{R}^{C\times H_s \times W_s}$ and content feature $F_c \in \mathbb{R}^{C\times H_c \times W_c}$, we first obtain their transformed and vectorized features as follows:
\begin{equation}
	\begin{aligned}
		&\hat{F}_c := \Gamma(f_c (Norm(F_c))) \in \mathbb{R}^{C\times H_cW_c}, \\
		&\hat{F}_{sa}^1 := \Gamma(f_{sa}^1 (Norm(F_{sa}))) \in \mathbb{R}^{C\times H_sW_s}, \\
		&\hat{F}_{sa}^2 := \Gamma(f_{sa}^2 (F_{sa})) \in \mathbb{R}^{C\times H_sW_s}, 
	\end{aligned}
\end{equation}
where $Norm$ denotes the mean–variance channel-wise normalization, $f_c$, $f_{sa}^1$, and $f_{sa}^2$ are $1\times1$ learnable convolutions. 

Then, we calculate the style attention between $\hat{F}_c$ and $\hat{F}_{sa}^1$ as follows:
\begin{equation}
	A_s := Softmax ((\hat{F}_c)^T \otimes \hat{F}_{sa}^1) \in \mathbb{R}^{H_cW_c\times H_sW_s}.
\end{equation}

Finally, the output feature $F_{cs}$ is achieved by:
\begin{equation}
	F_{cs} := f_{out}^2 (\hat{\Gamma}(\hat{F}_{sa}^2 \otimes (A_s)^T) ) \oplus F_c \in \mathbb{R}^{C\times H_c \times W_c},
\end{equation}
where $f_{out}^2$ is $1\times1$ learnable convolution.

{\em Discussions.} In summary, our AesSA performs style integration considering both global aesthetic guidance and local structure guidance. It is closely related to the style attention mechanisms of SANet~\cite{park2019arbitrary} and MAST~\cite{deng2020arbitrary}. There are, however, two main differences: (1) The inputs of AesSA include not only the VGG content feature $F_c$ and style feature $F_s$ but also the aesthetic feature $F_a$ from the discriminator, integrating different sorts of encoding information. (2) The AesSA is optimized using not only the traditional style transfer training but also the adversarial training with two new aesthetic regularizations. Therefore, our AesSA can appropriately enhance the style patterns and embed them into the content features by mapping the global channel-wise relationship between the style and aesthetic features and the local point-wise relationship between the content and enhanced style features through learning, helping produce aesthetically more realistic and pleasing results.

\vspace{-0.5em}
\subsection{Two-Stage Transfer Training}
\label{two-stage}

Our framework involves two training processes: adversarial training and traditional style transfer training. The adversarial training is to pursue domain transfer, which encourages that the generated results look like real artist-created paintings. The traditional style transfer training is to achieve universal style transfer, which aims to transfer arbitrary given styles to the content target.

As we introduced in the previous sections, the encodings of the aesthetic discriminator $\mathcal{D}_a$ are utilized to enhance the style transfer process. However, since the ability of the discriminator is gradually improved in the confrontation with the generator, the features it extracts are not always meaningful, especially at the beginning of the adversarial training. The meaningless features will deteriorate the style integration and affect the style transfer training. To overcome this defect, we develop a two-stage transfer training strategy to train the proposed framework in a transfer learning fashion which contains a pre-training stage and a fine-tuning stage. The details of each stage and loss functions are provided below.

{\bf Stage I: Pre-training.} Since the features extracted from the aesthetic discriminator $\mathcal{D}_a$ are meaningless at the beginning of the adversarial training, we do not inject them into the AesSA module at this stage to guide the style transfer. Instead, we directly exploit the VGG style feature $F_s$ as the aesthetic feature $F_a$ to feed into the AesSA module. In this way, we can pre-train the AesSA module and the decoder to provide a good initialization for the fine-tuning of the next stage. The aesthetic discriminator $\mathcal{D}_a$ here only acts as the discriminator, which plays the min-max game as follows:
\begin{equation}
	\begin{aligned}
		\max \limits_{\mathcal{D}_a}&  \min \limits_{G} \mathcal{L}_{adv}^1 := \underset{I_s \sim \Phi_s} {\mathbb{E}}[log (\mathcal{D}_a (I_s))] \\
		& + \underset{I_c \sim \Phi_c, I_s \sim \Phi_s} {\mathbb{E}} [log(1- \mathcal{D}_a(G(I_c, I_s)))], \\
		G(I_c, I_s)  &= D(AesSA(E_{vgg}(I_c), E_{vgg}(I_s), E_{vgg}(I_s))), 
	\end{aligned}
\end{equation}
where our generator $G$ consists of a fixed VGG encoder $E_{vgg}$, a decoder $D$, and a feature transform module $AesSA$.

For universal style transfer, similar to~\cite{huang2017arbitrary,park2019arbitrary}, we use the pre-trained VGG encoder $E_{vgg}$ to compute the content loss $\mathcal{L}_c$ and style loss $\mathcal{L}_s$ as follows:
\begin{equation}
	\begin{aligned}
		\mathcal{L}_c := \sum_{i=1}^{L_c} &\parallel Norm(\phi_i(I_{cs})) - Norm(\phi_i(I_{c})) \parallel_2, \\
		\mathcal{L}_s := \sum_{i=1}^{L_s} & (\parallel \mu(\phi_i(I_{cs})) - \mu(\phi_i(I_{s})) \parallel_2  \\
		& + \parallel \sigma(\phi_i(I_{cs})) - \sigma(\phi_i(I_{s})) \parallel_2) ,
	\end{aligned}
	\label{eq:csloss}
\end{equation}
where $\mu$ and $\sigma$ are the channel-wise mean and standard deviation, respectively. $\phi_i$ denotes the $i^{th}$ layer of VGG encoder $E_{vgg}$. We use layers \{$Relu4\_1$, $Relu5\_1$\} to compute the content loss, and layers \{$Relu1\_1$, $Relu2\_1$, $Relu3\_1$, $Relu4\_1$, $Relu5\_1$\} to compute the style loss.

Moreover, inspired by~\cite{park2019arbitrary,deng2020arbitrary}, we also use an identity loss $\mathcal{L}_{id}$ at this stage to constrain the identity mapping relations between content features and style features, which helps better pre-train the generator to maintain the content structure and style characteristics simultaneously~\cite{park2019arbitrary}.
\begin{equation}
	\begin{aligned}
		\mathcal{L}_{id} & := \parallel I_{cc} - I_c\parallel_2 + \parallel I_{ss} - I_s\parallel_2,
	\end{aligned}
	\label{eq:idloss}
\end{equation}
where $I_{cc}$ denotes the generated results using a photograph $I_c \in \Phi_c$ as both content and style images, and $I_{ss}$ denotes the generated results using a painting $I_s \in \Phi_s$ as both content and style images.

{\bf Full Objective of Stage I.} At this stage, the aesthetic discriminator’s final objective is
\begin{equation}
	\begin{aligned}
		\max \limits_{\mathcal{D}_a} \mathcal{L}_{adv}^1;
	\end{aligned}
\end{equation}
while the generator's final objective is
\begin{equation}
	\begin{aligned}
		\min \limits_{G}\lambda_1 \mathcal{L}_{adv}^1 + \lambda_2 \mathcal{L}_c + \lambda_3 \mathcal{L}_s + \lambda_4 \mathcal{L}_{id},
	\end{aligned}
    \label{eq12}
\end{equation}
where $\lambda_1$, $\lambda_2$, $\lambda_3$, and $\lambda_4$ are the trade-off weights which are set to $\lambda_1=5$, $\lambda_2=1$, $\lambda_3=1$, and $\lambda_4=50$.

{\bf Stage II: Fine-tuning.} After rounds of adversarial training at stage I, the aesthetic discriminator $D_a$ is capable of extracting the meaningful aesthetic encodings of input images. Thus, at this stage, we inject the aesthetic feature of style image into the AesSA module to provide the global aesthetic guidance for style transfer. The aesthetic discriminator $\mathcal{D}_a$ here acts as both the discriminator and the feature extractor, which plays a new min-max game as follows:
\begin{equation}
	\begin{aligned}
		\max \limits_{\mathcal{D}_a}&  \min \limits_{G} \mathcal{L}_{adv}^2 := \underset{I_s \sim \Phi_s} {\mathbb{E}}[log (\mathcal{D}_a (I_s))] \\
		& + \underset{I_c \sim \Phi_c, I_s \sim \Phi_s} {\mathbb{E}} [log(1- \mathcal{D}_a(G(I_c, I_s)))], \\
		G(I_c, I_s)  &= D(AesSA(E_{vgg}(I_c), E_{vgg}(I_s), \mathcal{D}_a(I_s))), 
	\end{aligned}
\end{equation}
here the generator $G$ consists of $E_{vgg}$, $D$, $AesSA$, and the aesthetic discriminator $D_a$. $D_a$ is trained when maximizing $\mathcal{L}_{adv}$, and fixed when minimizing $\mathcal{L}_{adv}$.

\begin{figure}[t]
	\centering
	\includegraphics[width=0.9\linewidth]{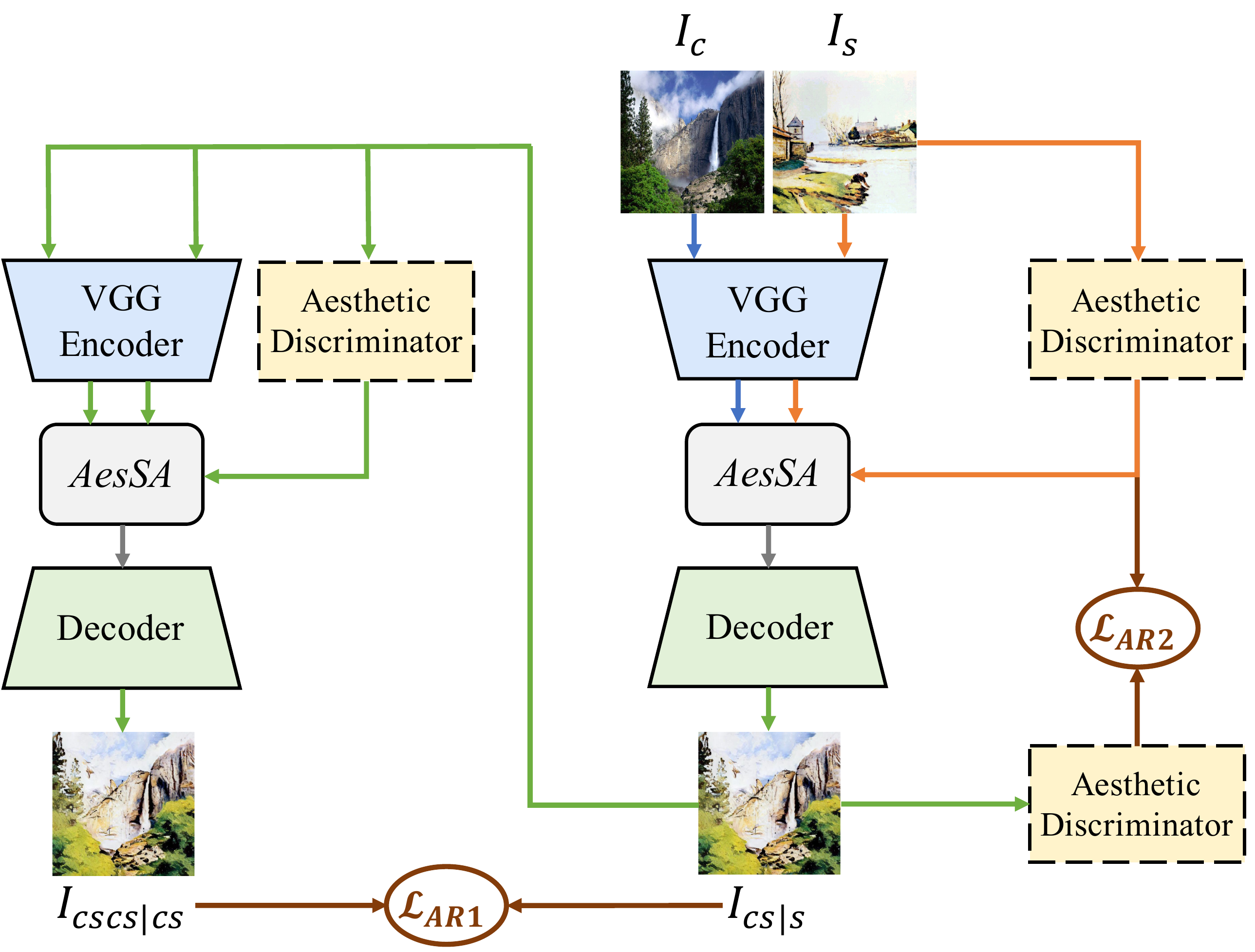}
	\vspace{-1em}
	\caption{{\bf Our proposed two new aesthetic regularizations}.
	}
	\label{fig:AR}
	\vspace{-1em}
\end{figure}

We also optimize the content loss $\mathcal{L}_c$ and style loss $\mathcal{L}_s$ defined in Eq.~(\ref{eq:csloss}) to maintain the generator's generalization ability for universal style transfer. Moreover, to train the networks more effectively and further elevate the stylization performance, we introduce two new aesthetic regularizations at this stage, as shown in Fig.~\ref{fig:AR}.

The form of the first aesthetic regularization $\mathcal{L}_{AR1}$ is similar to the identity loss defined in Eq.~(\ref{eq:idloss}), but we utilize the generated result as content image, style image, and aesthetic guidance simultaneously.
\begin{equation}
	\mathcal{L}_{AR1} := \parallel I_{cs|s} -  I_{cscs|cs}\parallel_2,
\end{equation}
where $I_{cs|s}$ denotes the generated result using a photograph $I_c \in \Phi_c$ as content image and a painting $I_s \in \Phi_s$ as both style image and aesthetic guidance. $I_{cscs|cs}$ denotes the generated result using $I_{cs|s}$ as content image, style image, and aesthetic guidance simultaneously. Through this way, we can not only constrain the identity mapping relations like identity loss, but also pull the content feature, style feature, and aesthetic feature extracted from the generated result $I_{cs|s}$ close to those extracted from the content image $I_c$ and style image $I_s$, thus improving the performance.

To further encourage the guidance of aesthetic features and prevent the networks from ignoring the aesthetic signals, we propose the second aesthetic regularization $\mathcal{L}_{AR2}$ to explicitly enforce the one-to-one mapping between the channel-wise aesthetic features extracted from the style image $I_s$ and the generated result $I_{cs|s}$.
\begin{equation}
	\begin{aligned}
		\mathcal{L}_{AR2} := &\parallel \mu(\mathcal{D}_a(I_s)) -  \mu(\mathcal{D}_a(I_{cs|s}))\parallel_2  \\
		&+ \parallel \sigma(\mathcal{D}_a(I_s)) -  \sigma(\mathcal{D}_a(I_{cs|s}))\parallel_2.
	\end{aligned}
\end{equation}
This kind of aesthetic regularization can help the model preserve aesthetically more pleasing style patterns, which will be clarified by our experiments in later Sec.~\ref{ablation}.

\begin{figure*}
	\centering
	\setlength{\tabcolsep}{0.02cm}
	\renewcommand\arraystretch{0.4}
	\begin{tabular}{cccccccc}
    \includegraphics[width=0.12\linewidth]{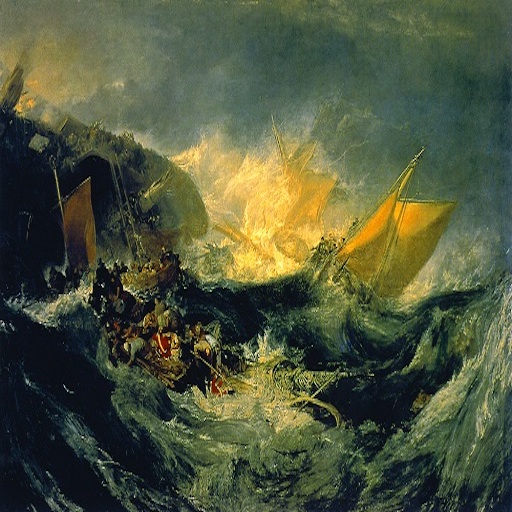}&
    \includegraphics[width=0.12\linewidth]{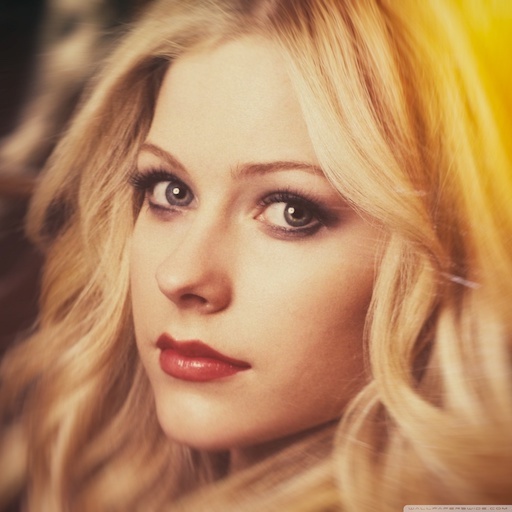}&
     \includegraphics[width=0.12\linewidth]{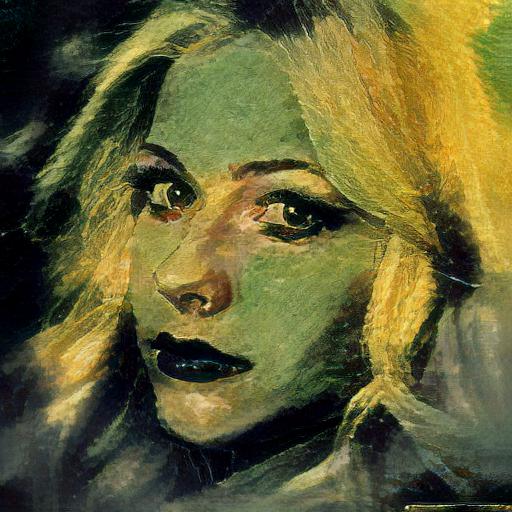} &
    \includegraphics[width=0.12\linewidth]{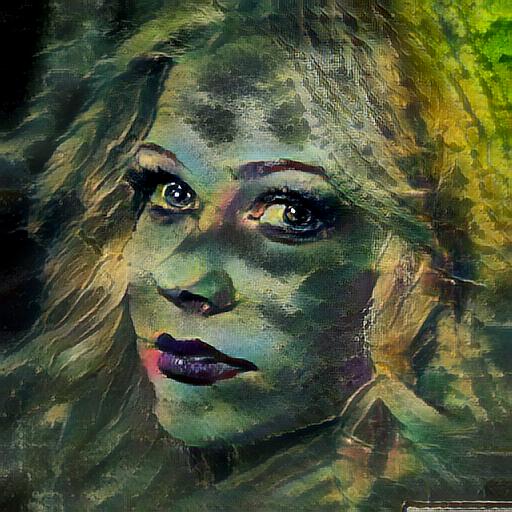}&
    \includegraphics[width=0.12\linewidth]{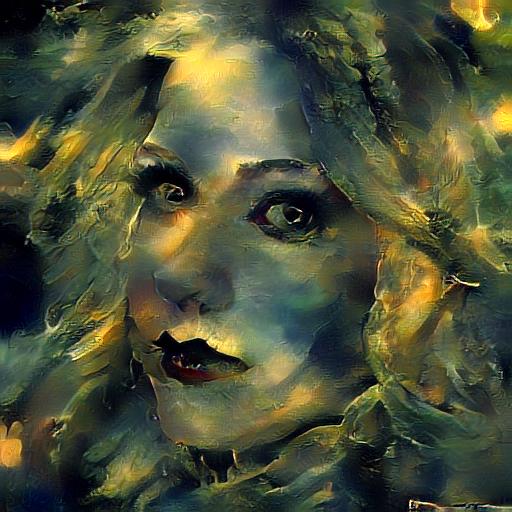}&
    \includegraphics[width=0.12\linewidth]{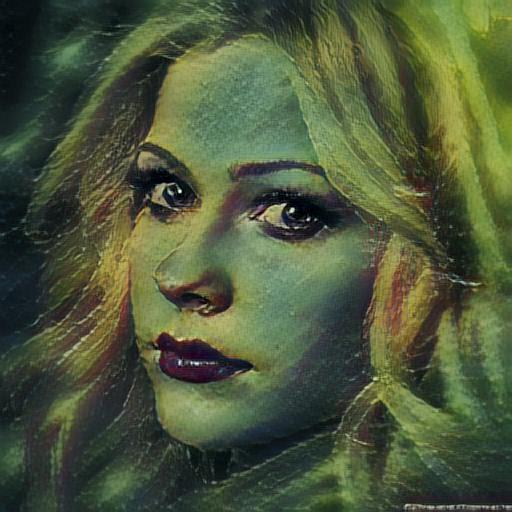}&
    \includegraphics[width=0.12\linewidth]{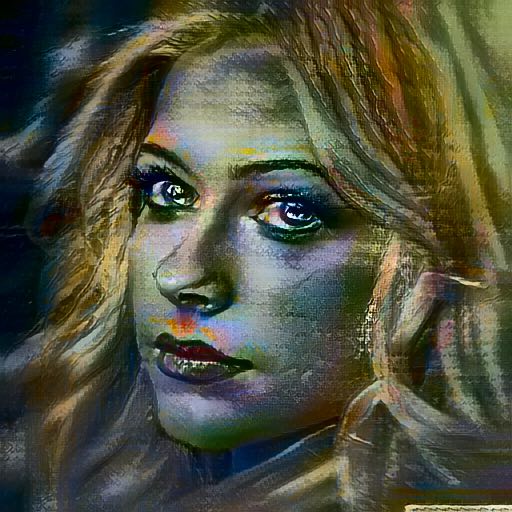}&
    \includegraphics[width=0.12\linewidth]{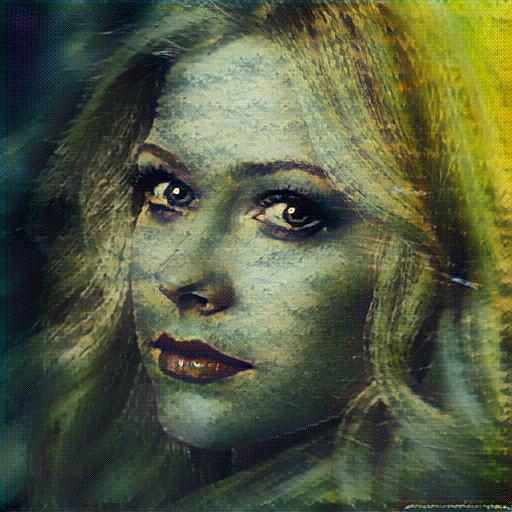}
    \\
    
    \includegraphics[width=0.12\linewidth]{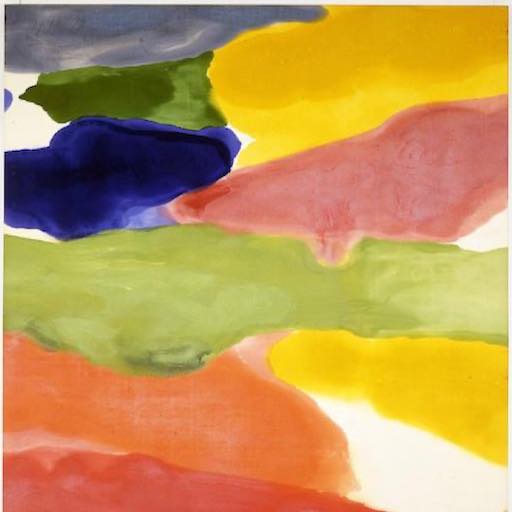}&
    \includegraphics[width=0.12\linewidth]{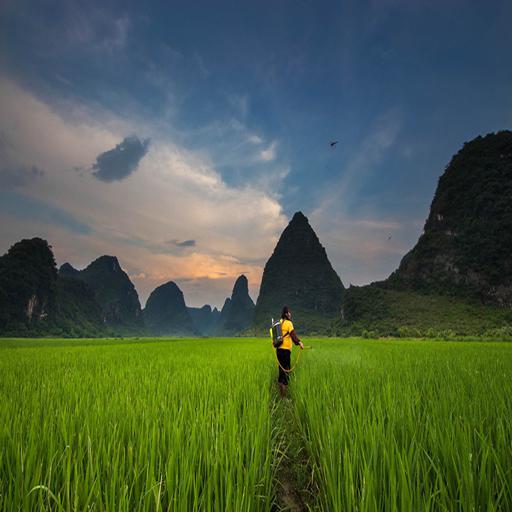}&
    \includegraphics[width=0.12\linewidth]{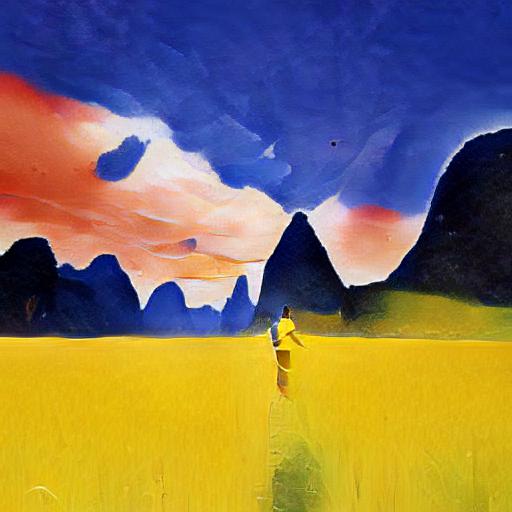} &
    \includegraphics[width=0.12\linewidth]{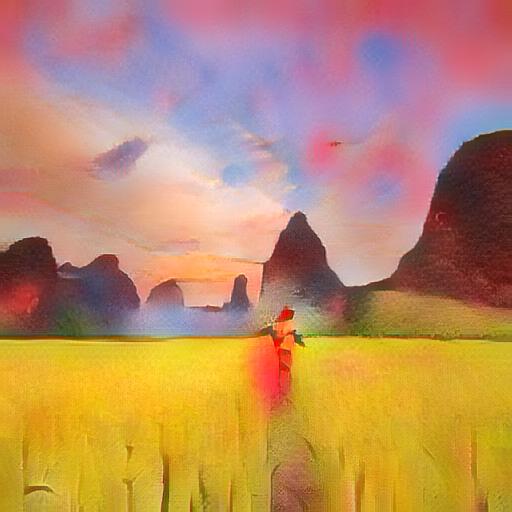}&
    \includegraphics[width=0.12\linewidth]{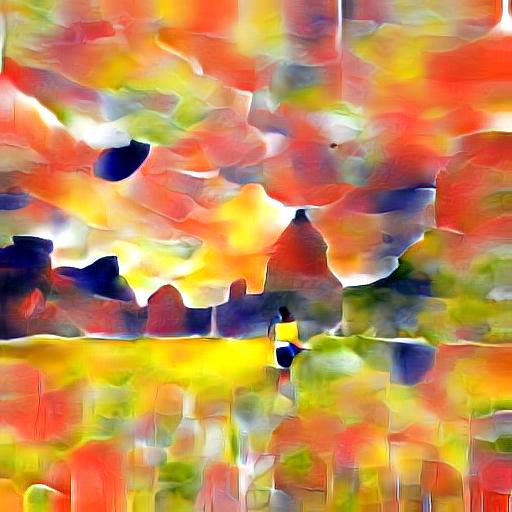}&
    \includegraphics[width=0.12\linewidth]{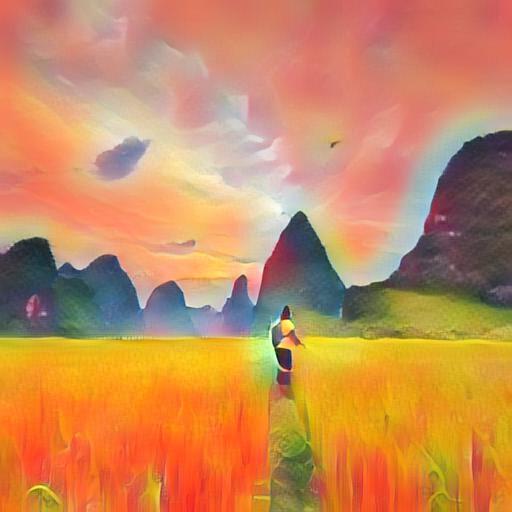}&
    \includegraphics[width=0.12\linewidth]{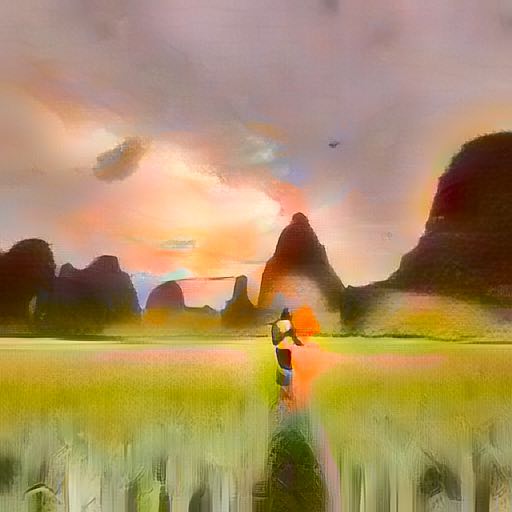}&
    \includegraphics[width=0.12\linewidth]{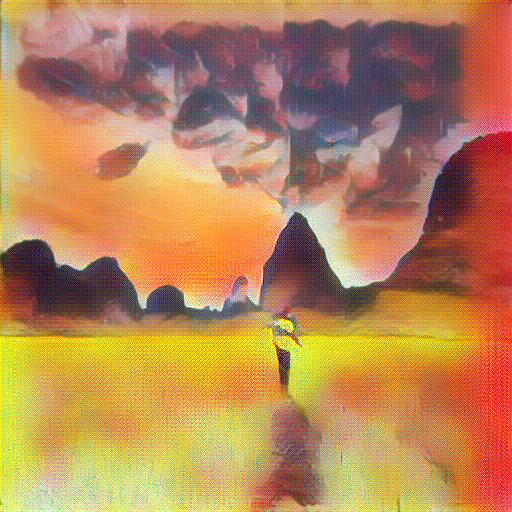}
    \\
    
    \includegraphics[width=0.12\linewidth]{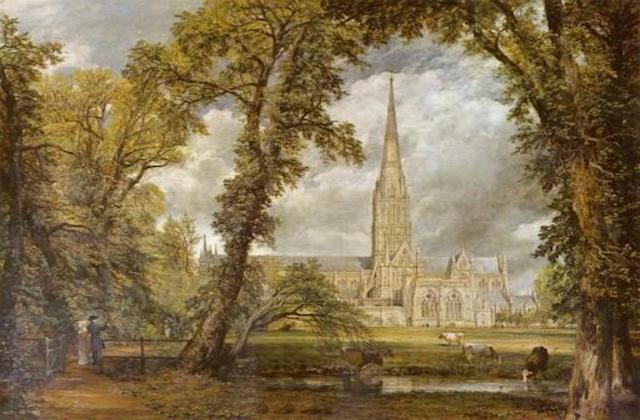}&
    \includegraphics[width=0.12\linewidth]{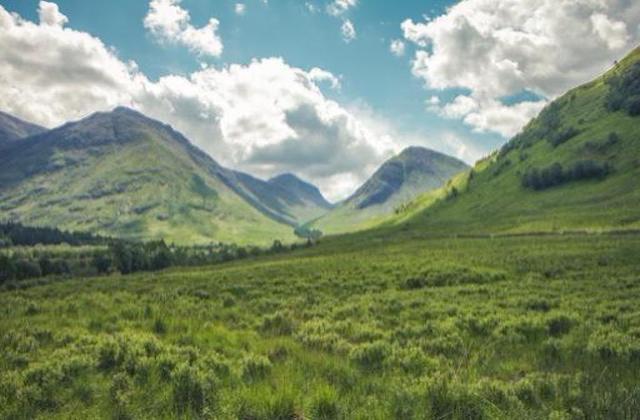}&
    \includegraphics[width=0.12\linewidth]{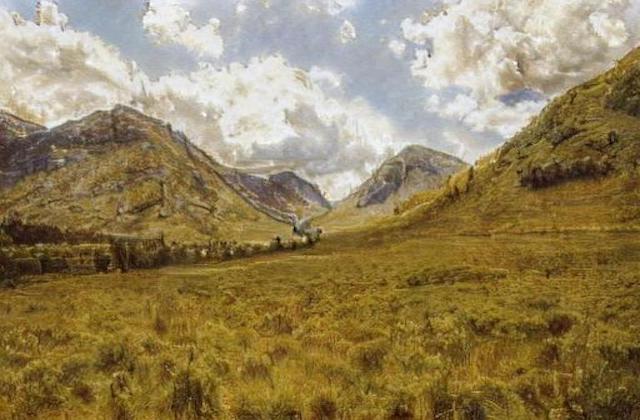} &
    \includegraphics[width=0.12\linewidth]{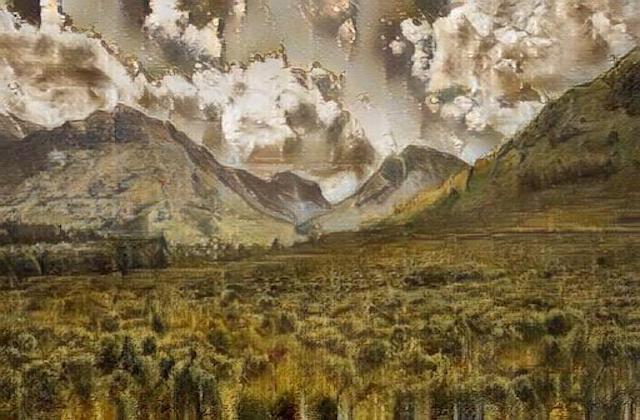}&
    \includegraphics[width=0.12\linewidth]{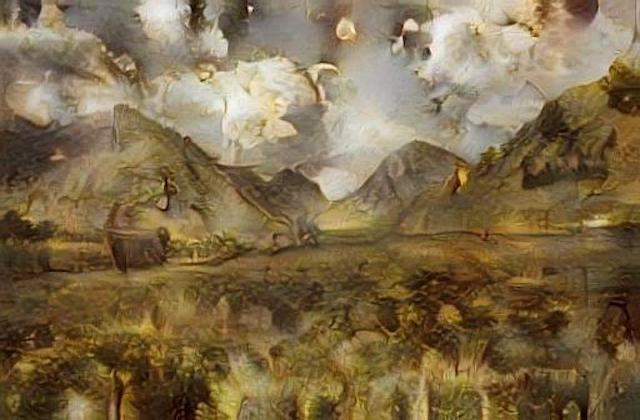}&
    \includegraphics[width=0.12\linewidth]{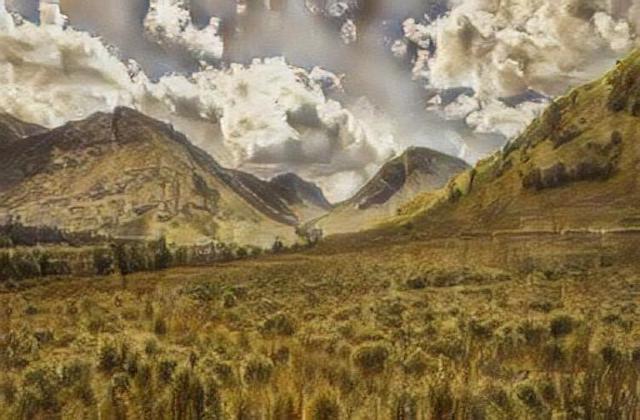}&
    \includegraphics[width=0.12\linewidth]{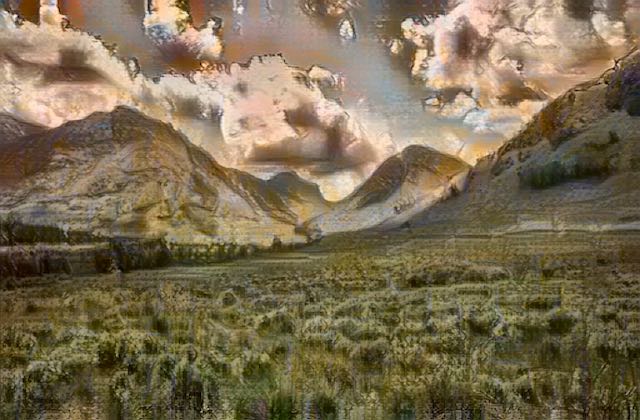}&
    \includegraphics[width=0.12\linewidth]{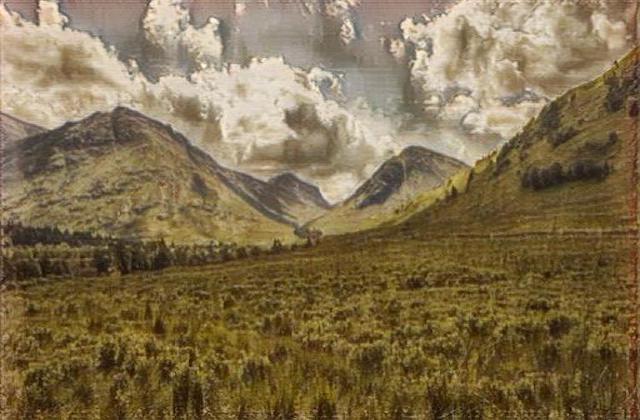}
    \\
    
    Style & Content & {\bf Ours}   &  AdaIN & WCT &  LST & AAST &  ArtFlow
    
    \\
    
    \includegraphics[width=0.12\linewidth]{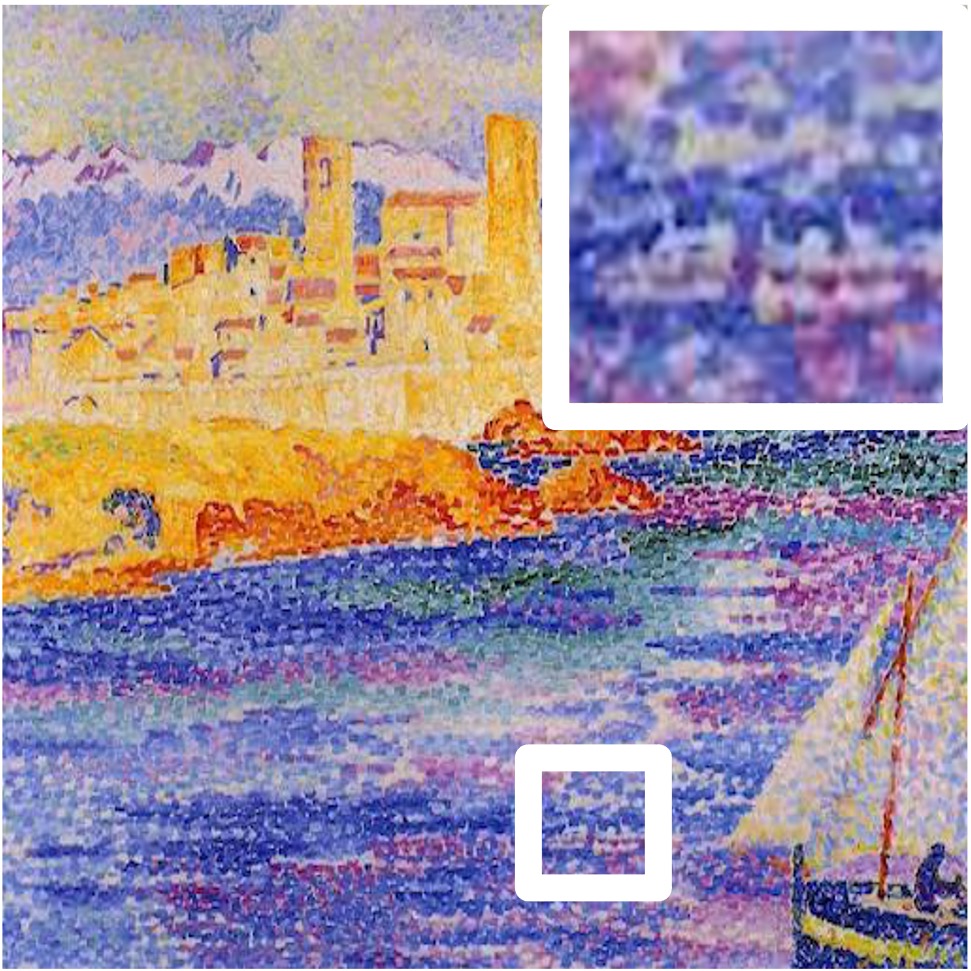}&
    \includegraphics[width=0.12\linewidth]{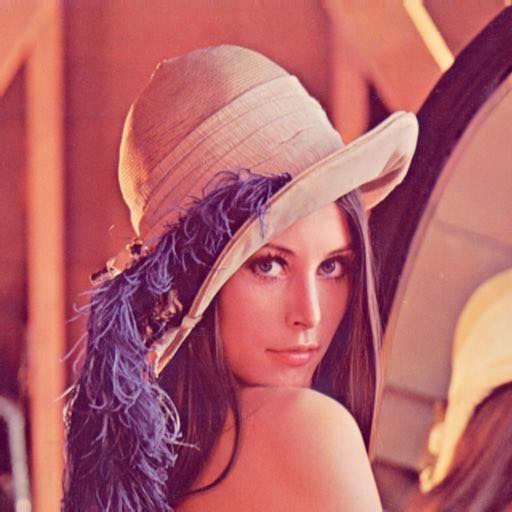}&
    \includegraphics[width=0.12\linewidth]{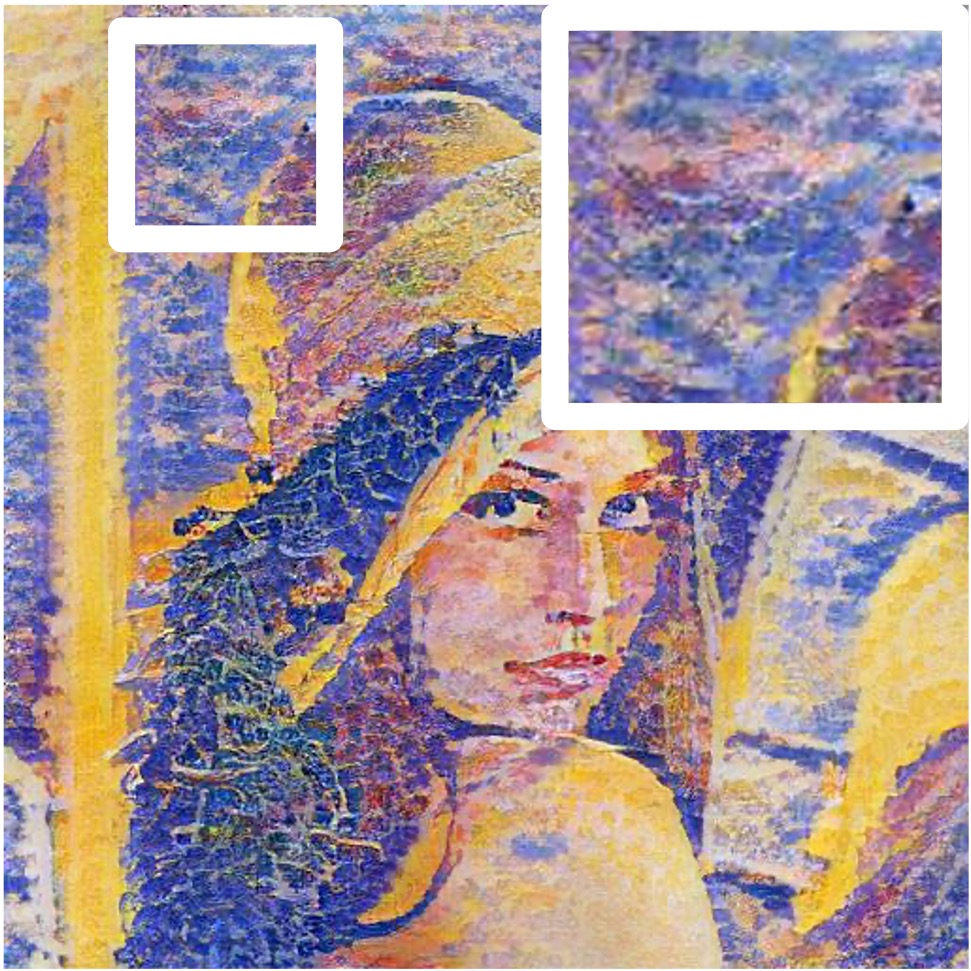} &
    \includegraphics[width=0.12\linewidth]{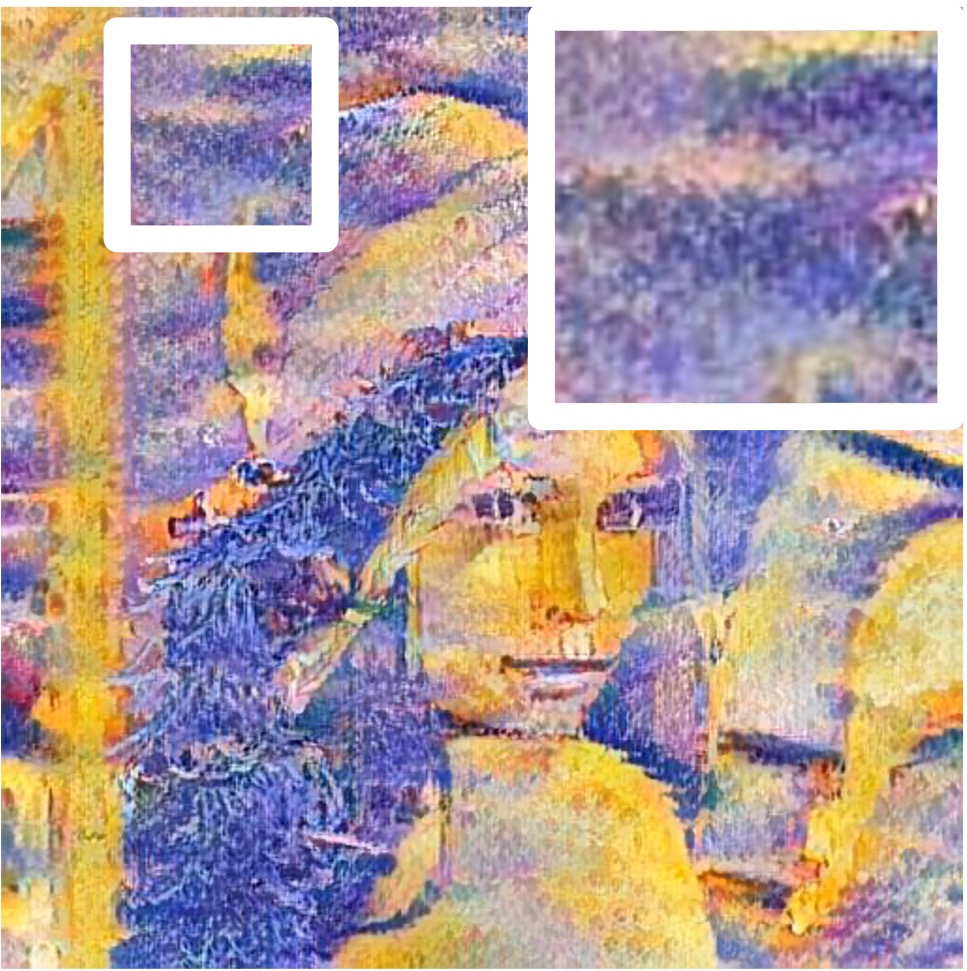}&
    \includegraphics[width=0.12\linewidth]{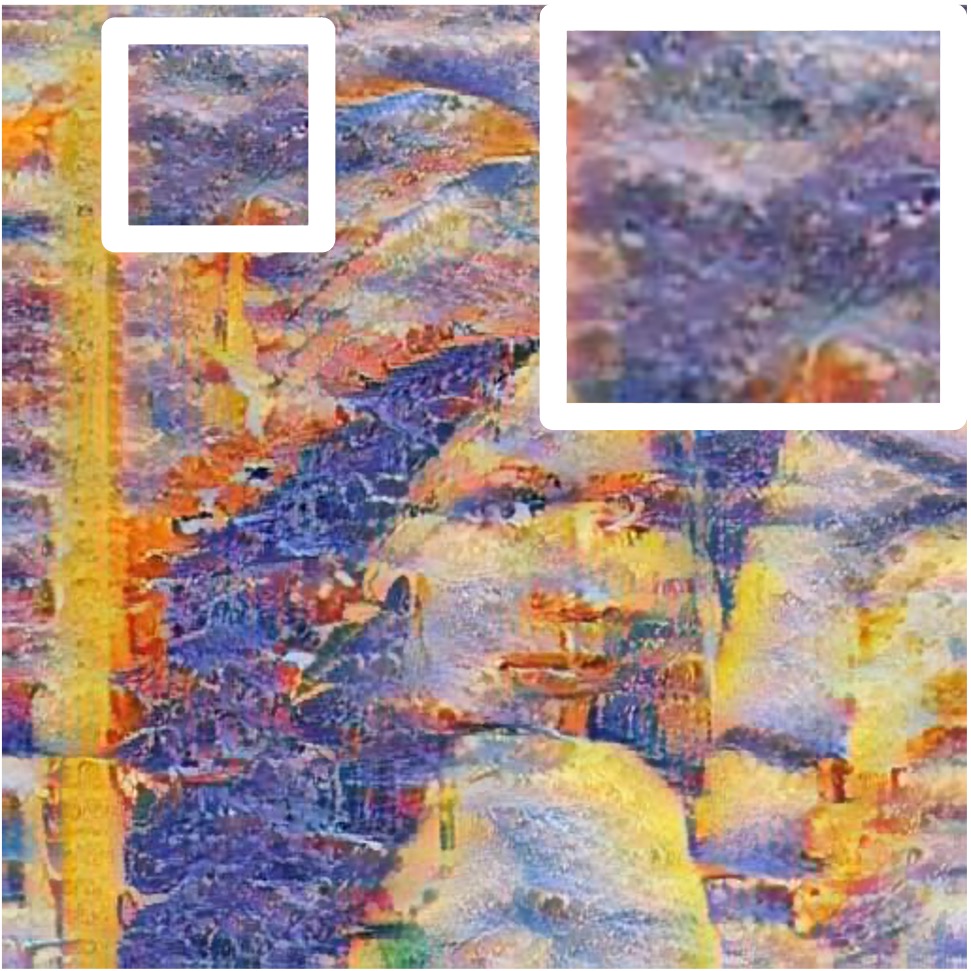}&
    \includegraphics[width=0.12\linewidth]{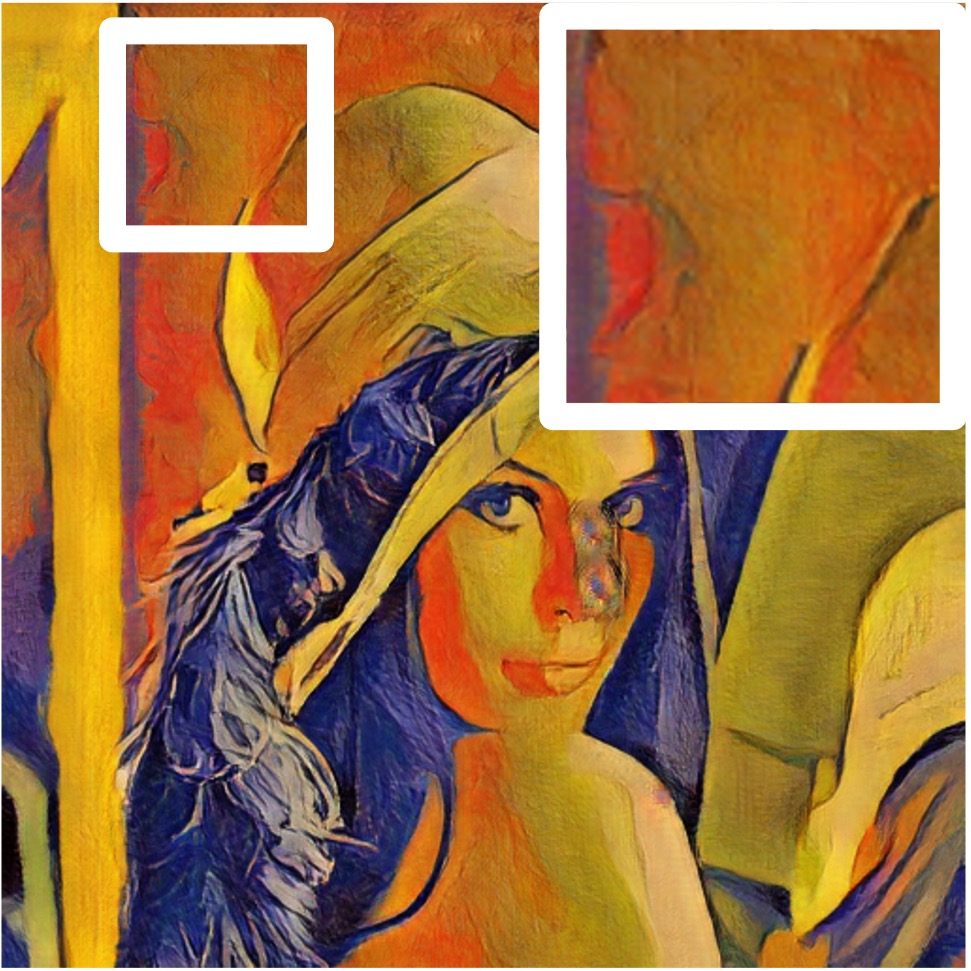}&
    \includegraphics[width=0.12\linewidth]{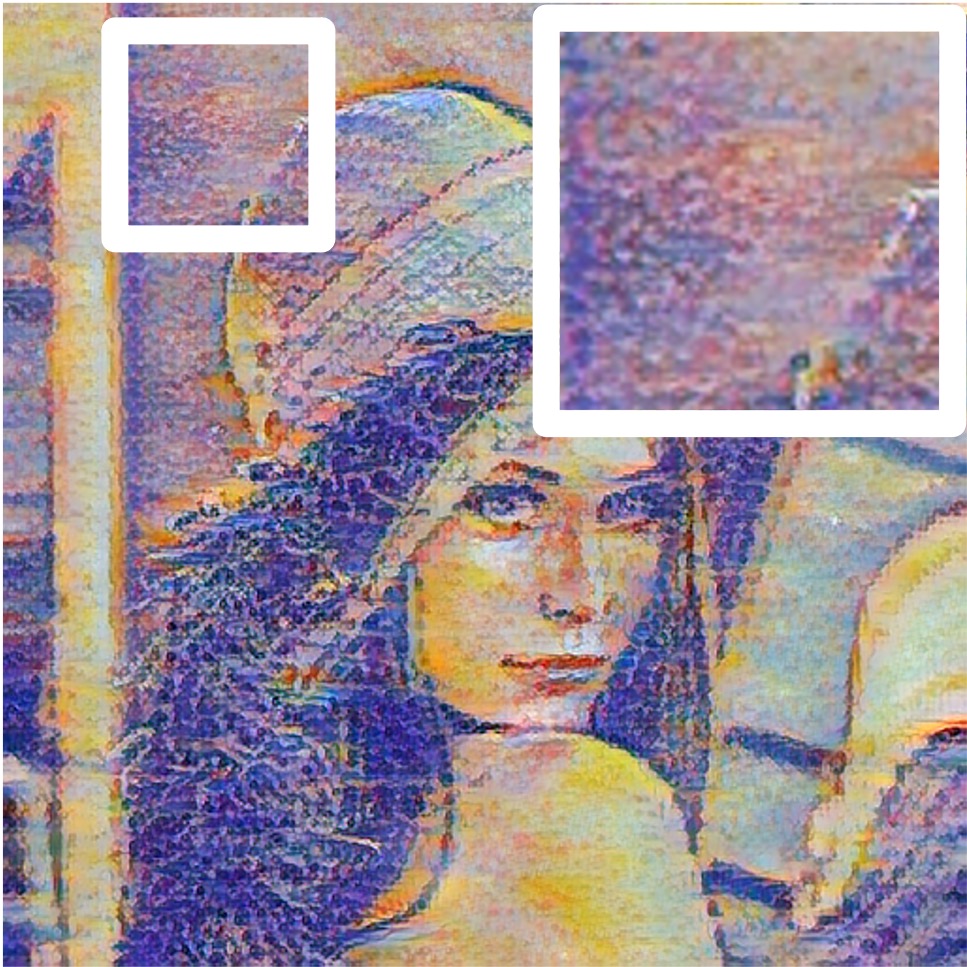}&
    \includegraphics[width=0.12\linewidth]{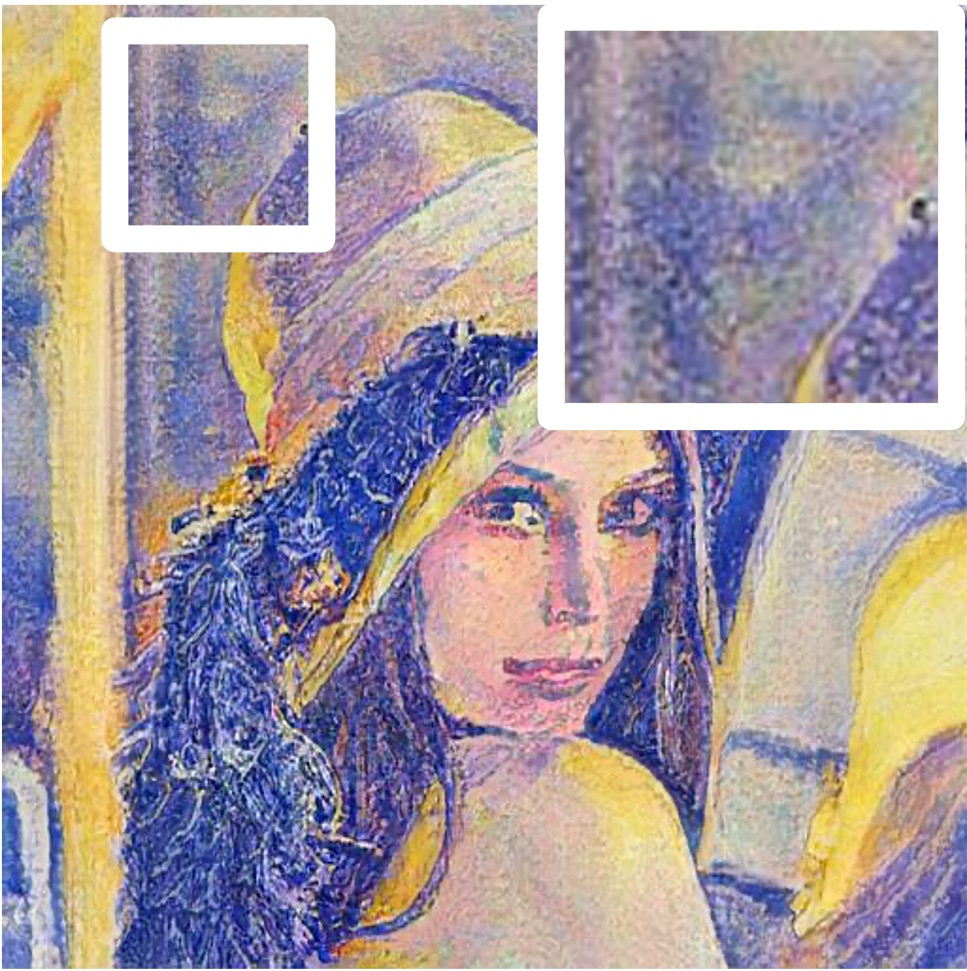}
    \\
    
    \includegraphics[width=0.12\linewidth]{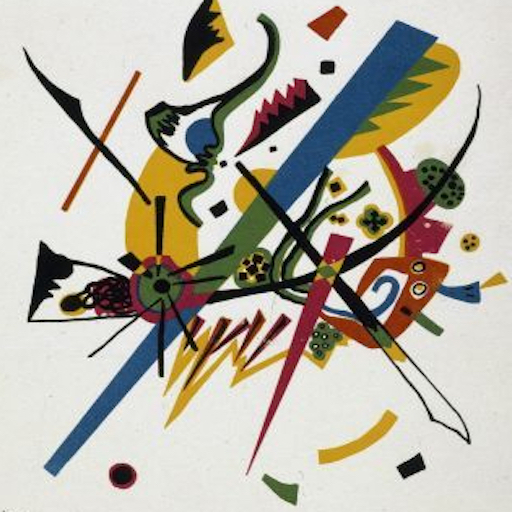}&
    \includegraphics[width=0.12\linewidth]{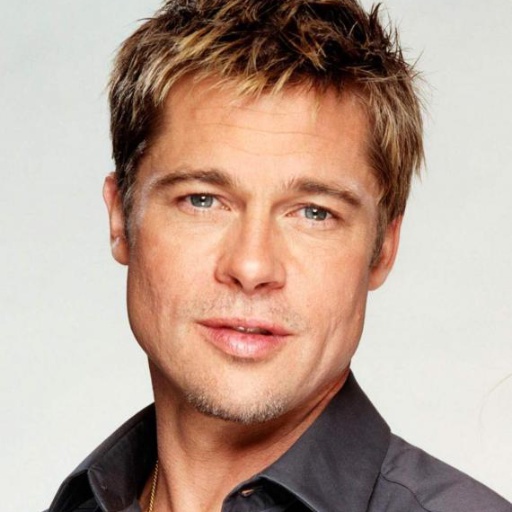}&
    \includegraphics[width=0.12\linewidth]{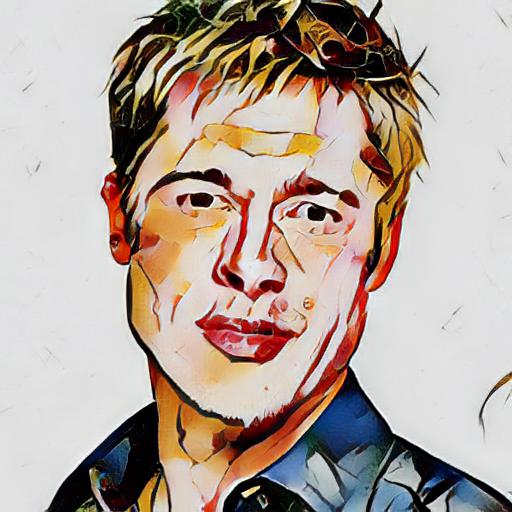} &
    \includegraphics[width=0.12\linewidth]{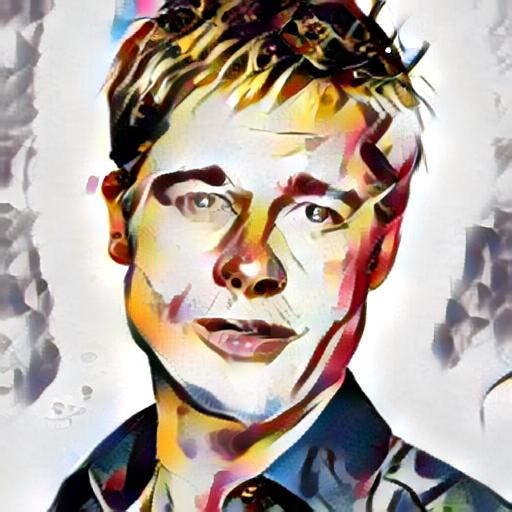}&
    \includegraphics[width=0.12\linewidth]{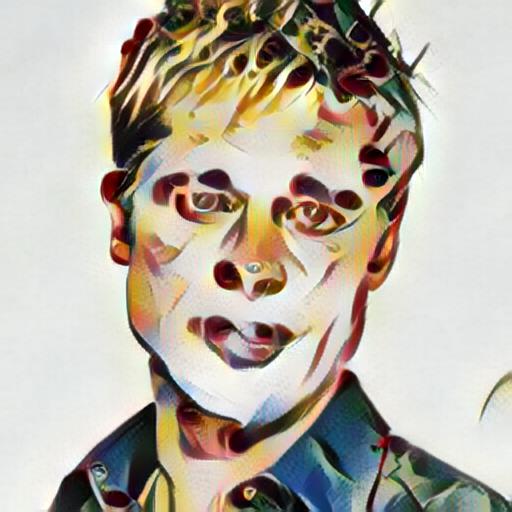}&
    \includegraphics[width=0.12\linewidth]{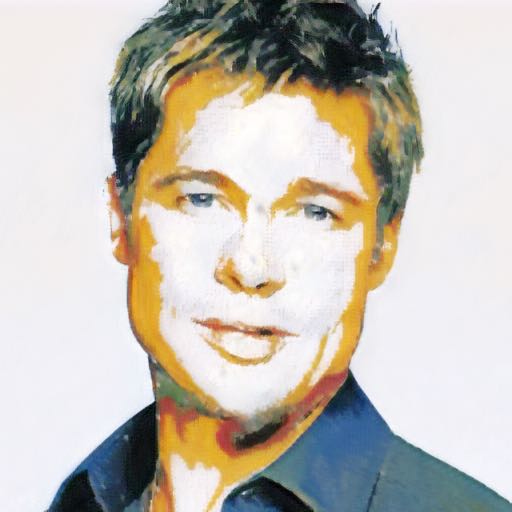}&
    \includegraphics[width=0.12\linewidth]{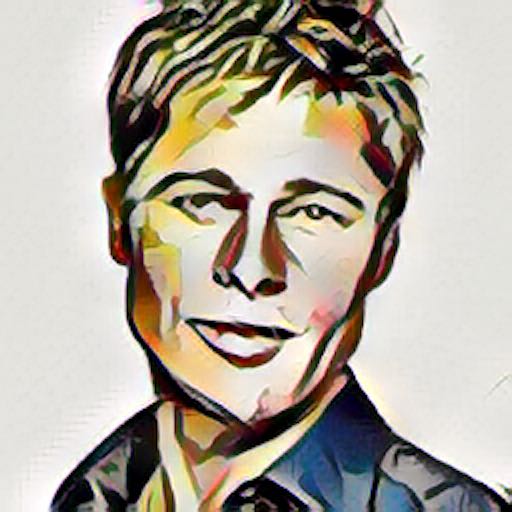}&
    \includegraphics[width=0.12\linewidth]{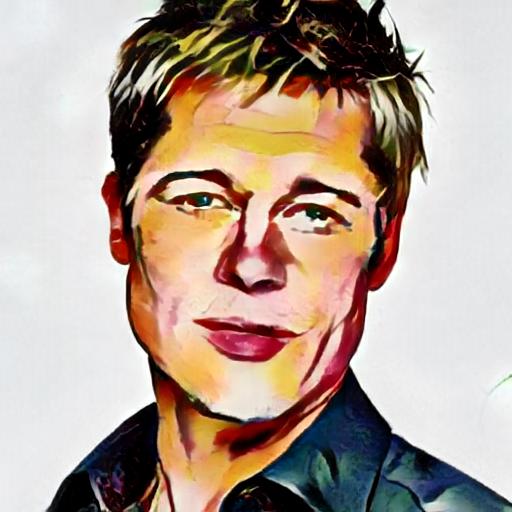}
    \\
    
    \includegraphics[width=0.12\linewidth]{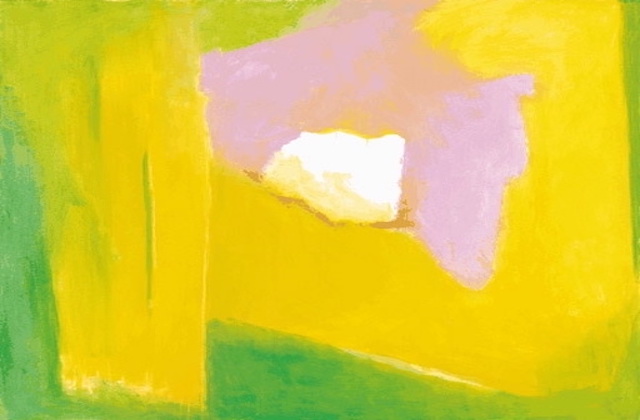}&
    \includegraphics[width=0.12\linewidth]{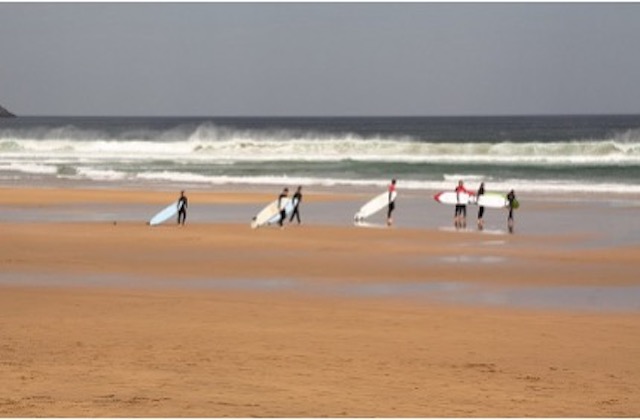}&
    \includegraphics[width=0.12\linewidth]{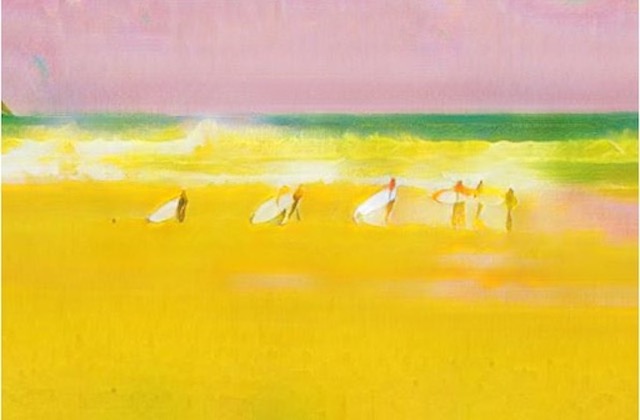} &
    \includegraphics[width=0.12\linewidth]{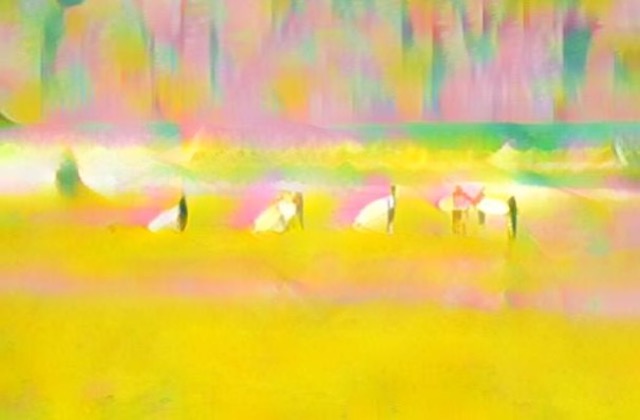}&
    \includegraphics[width=0.12\linewidth]{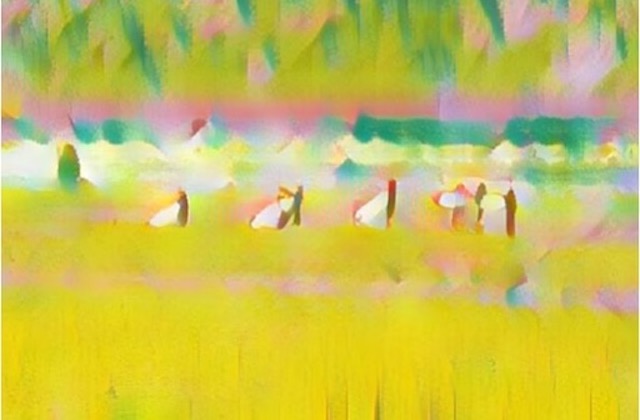}&
    \includegraphics[width=0.12\linewidth]{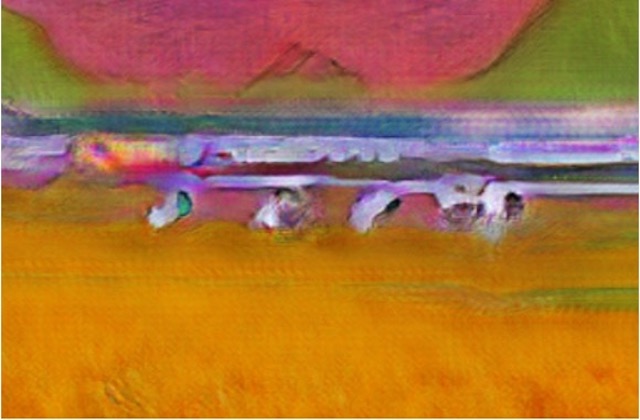}&
    \includegraphics[width=0.12\linewidth]{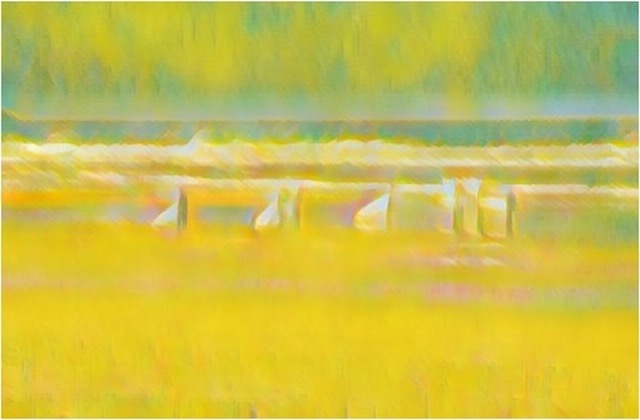}&
    \includegraphics[width=0.12\linewidth]{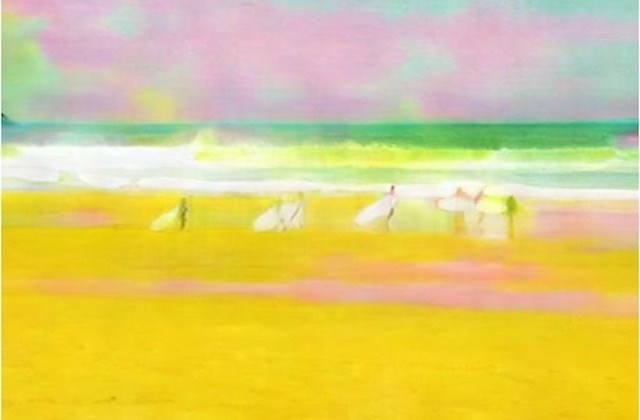}
    \\

    Style & Content & {\bf Ours}   &  SANet & MAST &  TPFR  & AdaAttN &  IECAST
		
	\end{tabular}
    \vspace{-1em}
	\caption{ {\bf Qualitative comparisons} with SOTA UST approaches. See more in Fig.~\ref{fig:teaser}. Zoom-in for better comparison.}
	\label{fig:quality}
\end{figure*} 

\renewcommand\arraystretch{1.1}
\begin{table*}[t]
	\centering
	\setlength{\tabcolsep}{0.2cm}
	\caption{{\bf Quantitative comparisons} with SOTA UST approaches.}
	\vspace{-1em}
	\begin{tabular}{c|c|ccccccccccc}
		\hline
		&  WikiArt &  \bf Ours &  AdaIN &  WCT &  LST &  AAST &  ArtFlow & SANet &  MAST &  TPFR &  AdaAttN &  IECAST
		\\
		\hline
		 CF &  - & \bf 0.420  &  0.383  & 0.346 &  0.408  & 0.368 & 0.391  & 0.366  &  0.360 & 0.401 & 0.380   & 0.412
		\\
		 GE + LP &  - & 1.524 & 1.480 & 1.540 &  1.459 &  1.461 & 1.492  & \bf 1.552 & 1.511  & 1.220 & 1.430  & 1.405
		\\
		 Deception &  0.784 & \bf 0.568 &0.241 &  0.172 & 0.408 & 0.307 &  0.356  & 0.346 & 0.324  & 0.414 & 0.372  & 0.476
		\\
		 Preference & - &  -  & 0.225 & 0.183 &  0.302 & 0.263 & 0.350   &0.287 & 0.254 & 0.247 &  0.383 & 0.391
		\\
		\hline
		 Time (sec.) & - & 0.066 & 0.045  &  1.163  &  0.036 & 1.153 & 0.382  & 0.064 &  0.124  &  0.582  &  0.142   & 0.064
		\\
		\hline
		
	\end{tabular}
	\label{tab:quantity}
\end{table*}

{\bf Full Objective of Stage II.} At this stage, the aesthetic discriminator’s final objective is
\begin{equation}
	\begin{aligned}
		\max \limits_{\mathcal{D}_a} \mathcal{L}_{adv}^2;
	\end{aligned}
\end{equation}
while the generator's final objective is
\begin{equation}
	\begin{aligned}
		\min \limits_{G}\lambda_5 \mathcal{L}_{adv}^2 + \lambda_6 \mathcal{L}_c + \lambda_7 \mathcal{L}_s + \lambda_8 \mathcal{L}_{AR1} + \lambda_9 \mathcal{L}_{AR2},
	\end{aligned}
    \label{eq17}
\end{equation}
where $\lambda_5$, $\lambda_6$, $\lambda_7$, $\lambda_8$, and $\lambda_9$ are the trade-off weights which are set to $\lambda_5=5$, $\lambda_6=1$, $\lambda_7=1$, $\lambda_8=0.5$, and $\lambda_9=500$.

\vspace{-0.5em}
\section{Experimental Results}

\subsection{Implementation Details}
We follow the multi-level strategy of~\cite{park2019arbitrary} by integrating two AesSA modules on $Relu4\_1$ and $Relu5\_1$ layers of VGG-19~\cite{simonyan2014very}, respectively. The network is trained by using MS-COCO~\cite{lin2014microsoft} as the content photograph dataset $\Phi_c$, and WikiArt~\cite{phillips2011wiki} as the artist-created painting dataset $\Phi_s$. Both datasets contain roughly 80000 training images. We use the Adam optimizer~\cite{kingma2014adam} with a learning rate of 0.0001 and a mini-batch size of 4 content-style image pairs at both two training stages. Each stage is trained for 80000 iterations. During training, all images are loaded with the smaller dimension rescaled to 512 while preserving the aspect ratio, and then randomly cropped to 256$\times$256 pixels. Since our network is fully convolutional, it can handle arbitrary input size during testing. All experiments are conducted on an NVIDIA RTX 2080 8GB GPU.

\vspace{-0.5em}
\subsection{Comparisons}
We compare our proposed AesUST against ten SOTA UST approaches including five global statistics-based methods (AdaIN~\cite{huang2017arbitrary}, WCT~\cite{li2017universal}, LST~\cite{li2019learning}, AAST~\cite{hu2020aesthetic}, and ArtFlow~\cite{an2021artflow}) and five local patch-based methods (SANet~\cite{park2019arbitrary}, MAST~\cite{deng2020arbitrary}, TPFR~\cite{svoboda2020two}, AdaAttN \cite{liu2021adaattn}, and IECAST~\cite{chen2021artistic}).

\begin{figure*}
	\centering
	\setlength{\tabcolsep}{0.07cm}
	\renewcommand\arraystretch{0.4}
	\begin{tabular}{ccp{0.05cm}|p{0.05cm}ccccp{0.05cm}|p{0.05cm}cc}
		\includegraphics[width=0.11\linewidth]{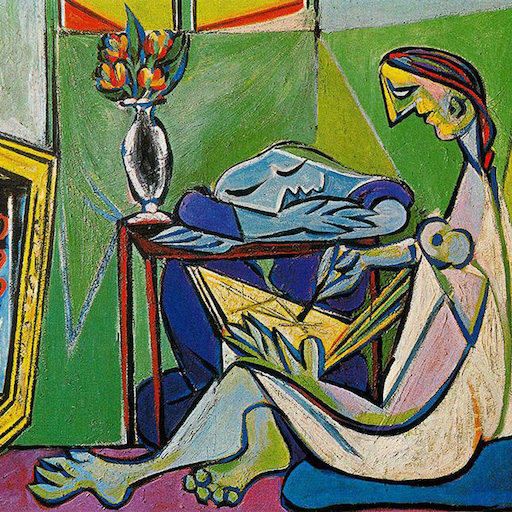}&
		\includegraphics[width=0.11\linewidth]{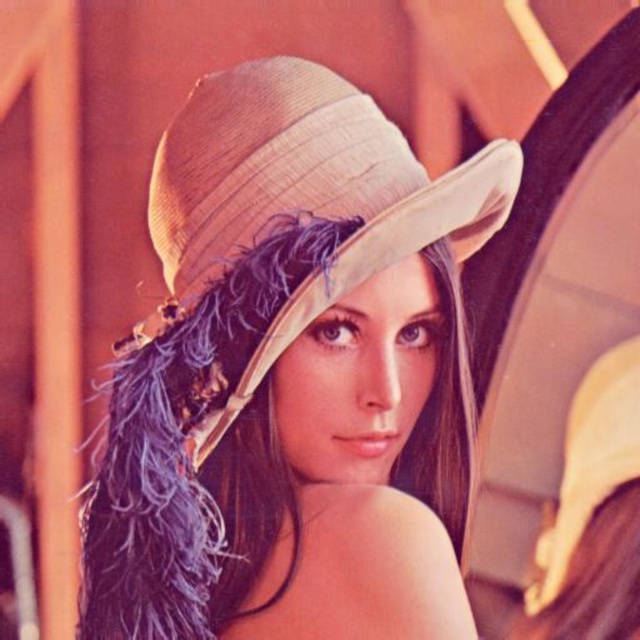}& &&
		\includegraphics[width=0.11\linewidth]{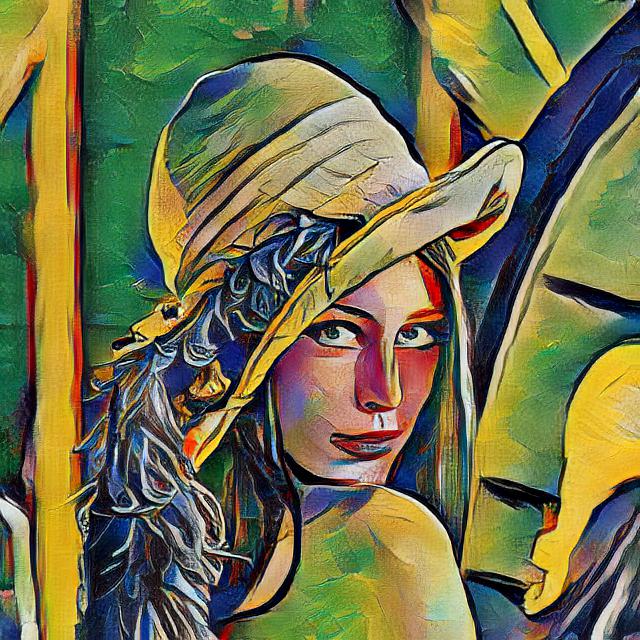}&
		\includegraphics[width=0.11\linewidth]{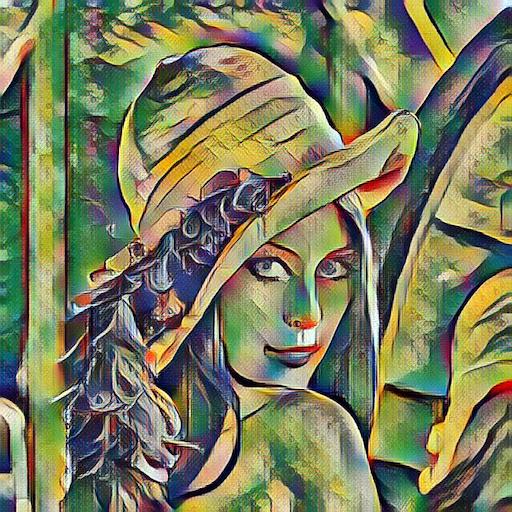}&
		\includegraphics[width=0.11\linewidth]{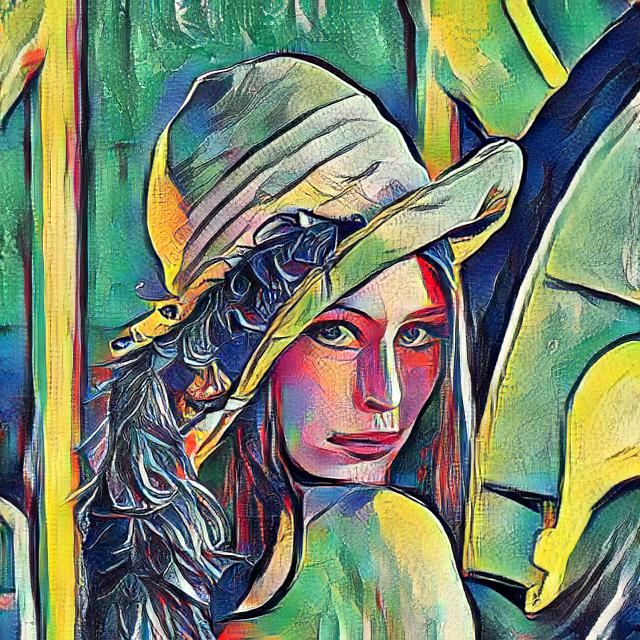}&
		\includegraphics[width=0.11\linewidth]{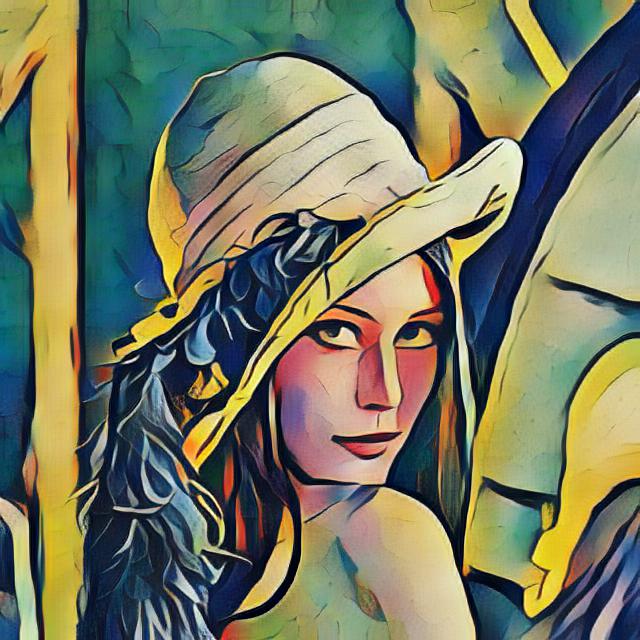}& &&
		\includegraphics[width=0.11\linewidth]{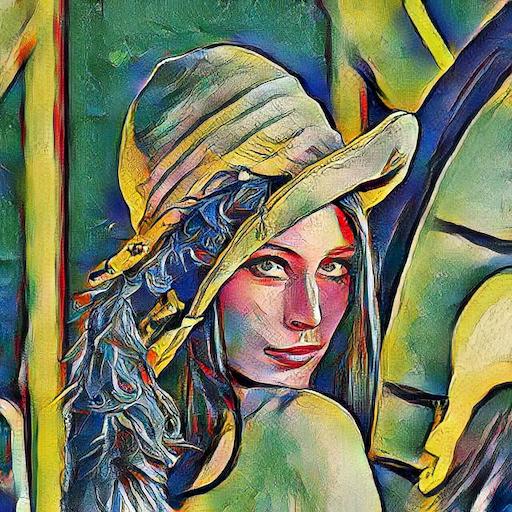}&
		\includegraphics[width=0.11\linewidth]{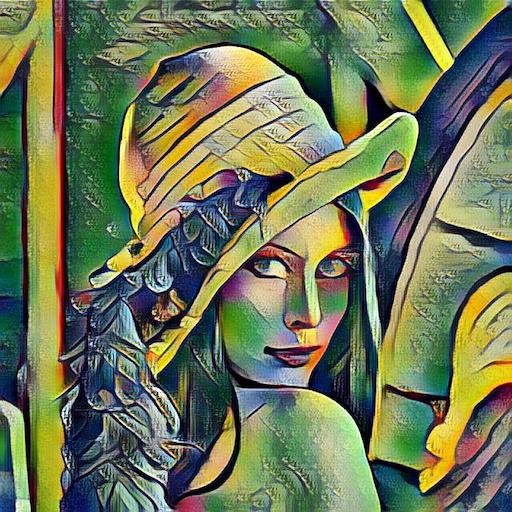}
		
		\\
		\includegraphics[width=0.11\linewidth]{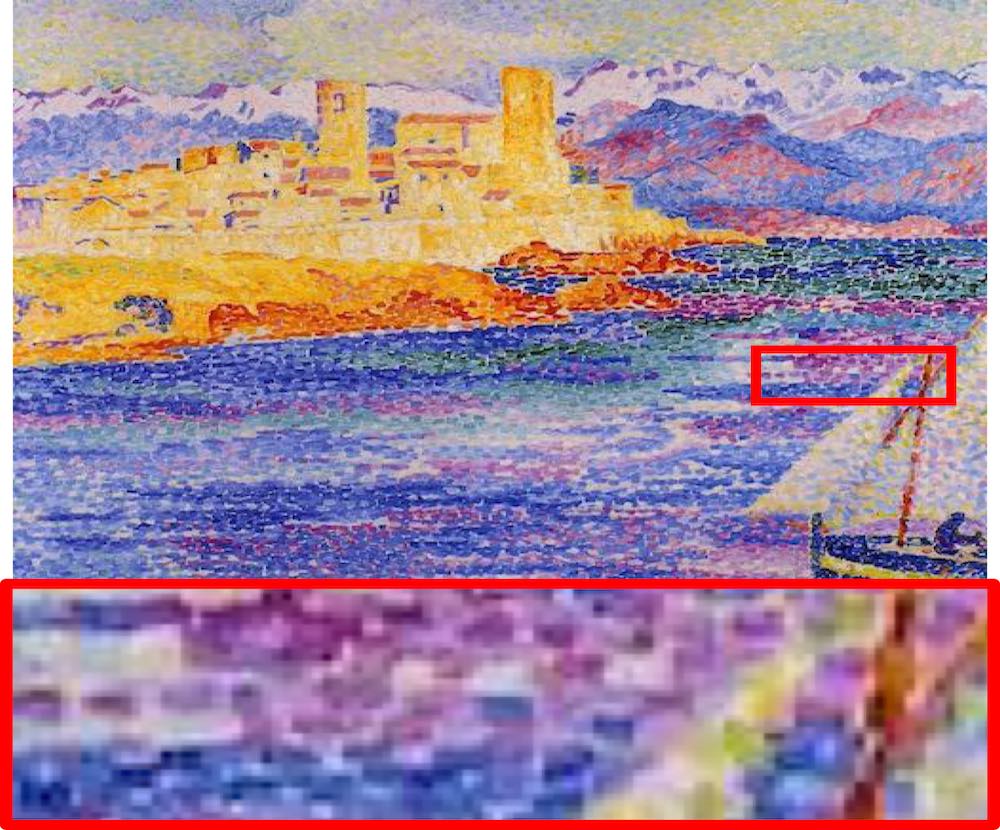}&
		\includegraphics[width=0.11\linewidth]{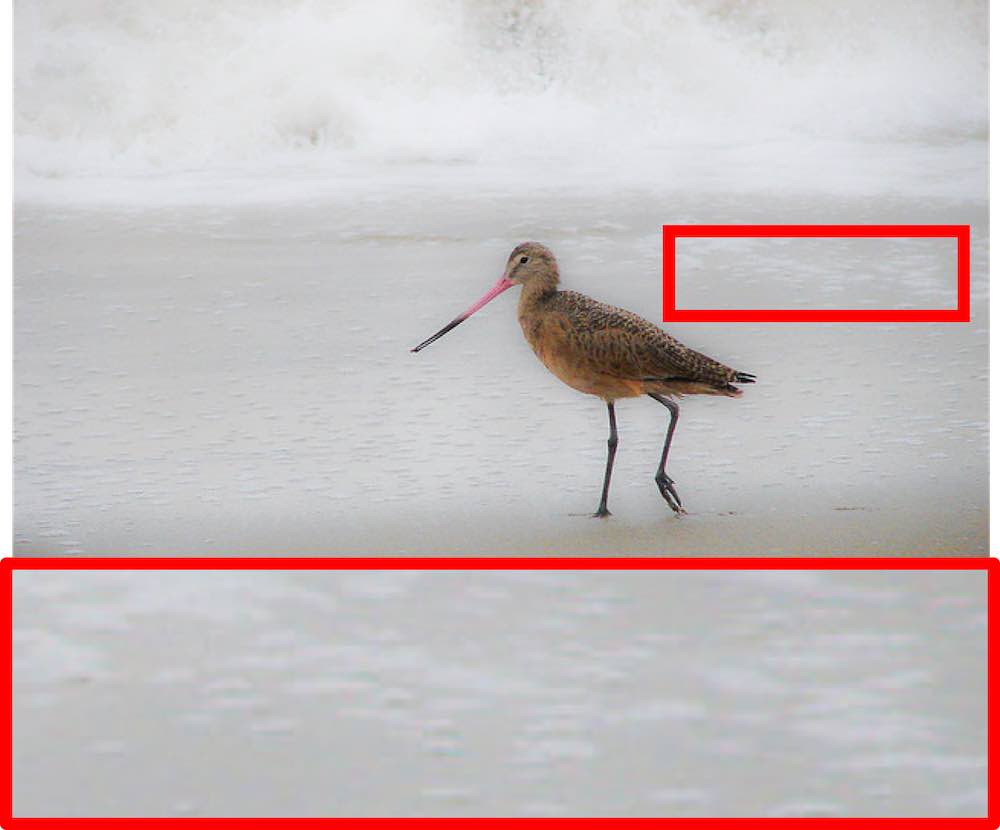}& &&
		\includegraphics[width=0.11\linewidth]{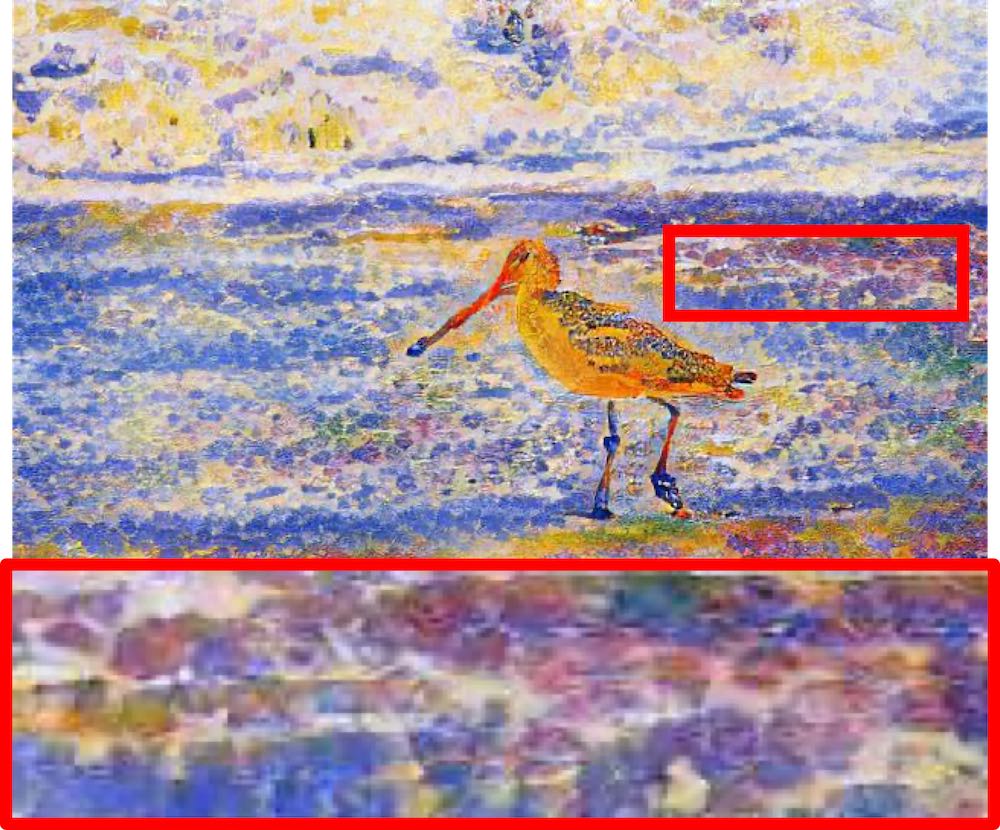}&
		\includegraphics[width=0.11\linewidth]{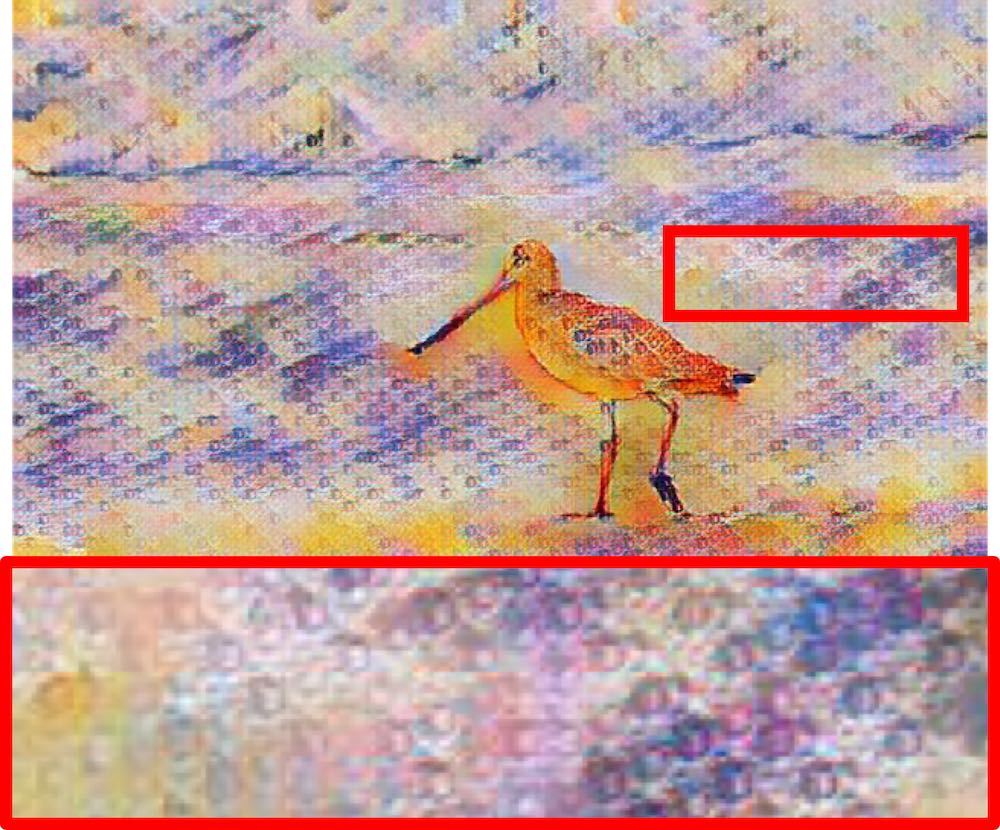}&
		\includegraphics[width=0.11\linewidth]{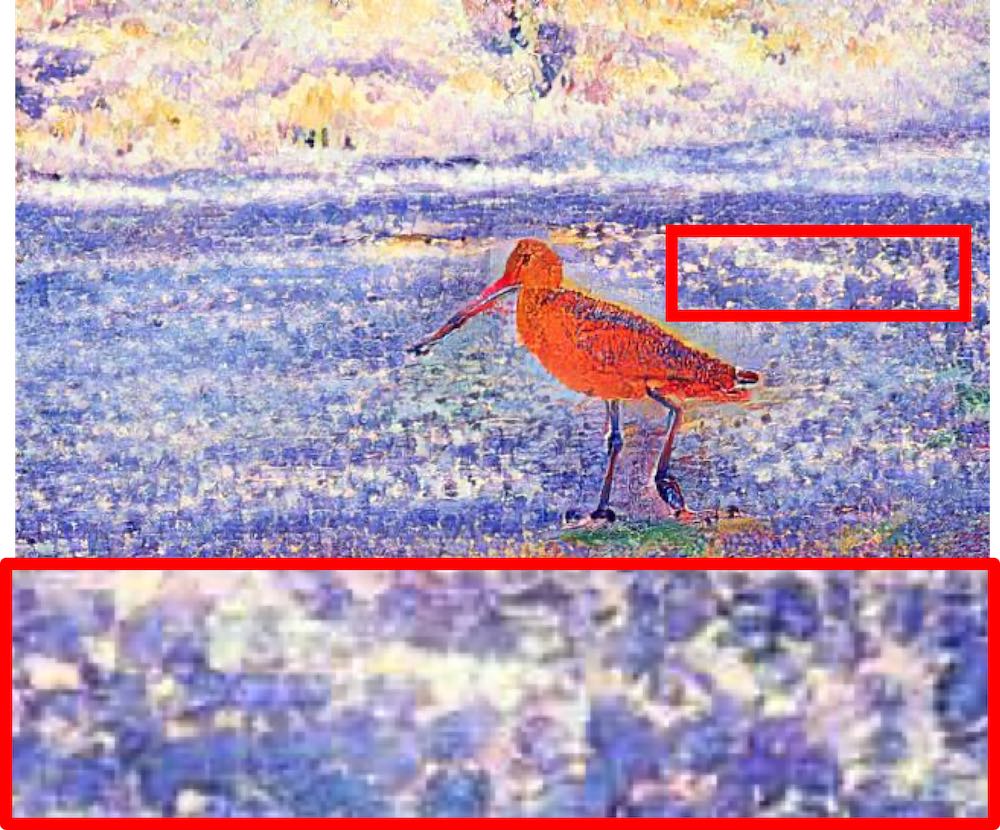}&
		\includegraphics[width=0.11\linewidth]{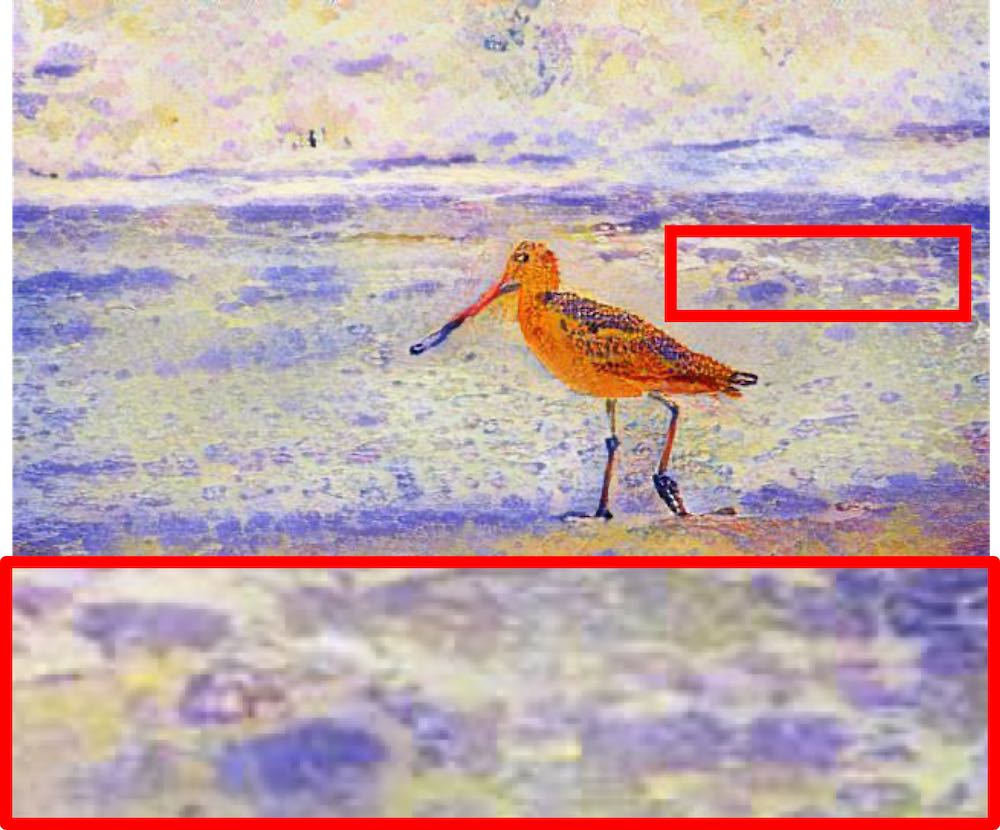}& &&
		\includegraphics[width=0.11\linewidth]{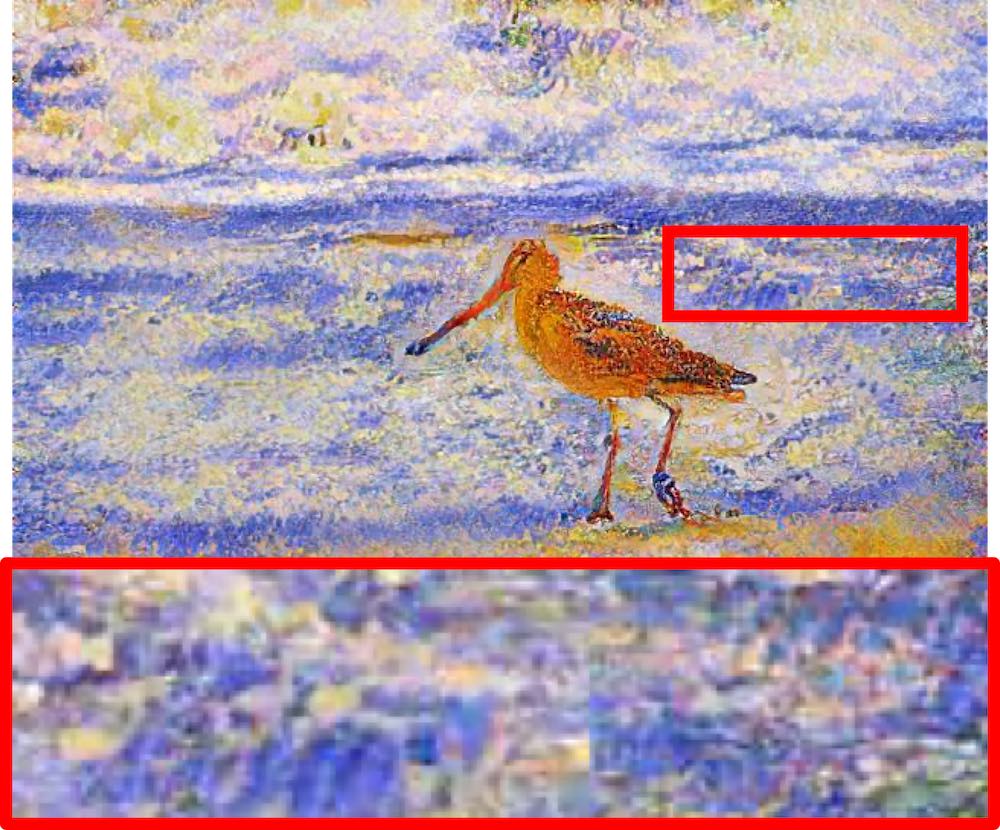}&
		\includegraphics[width=0.11\linewidth]{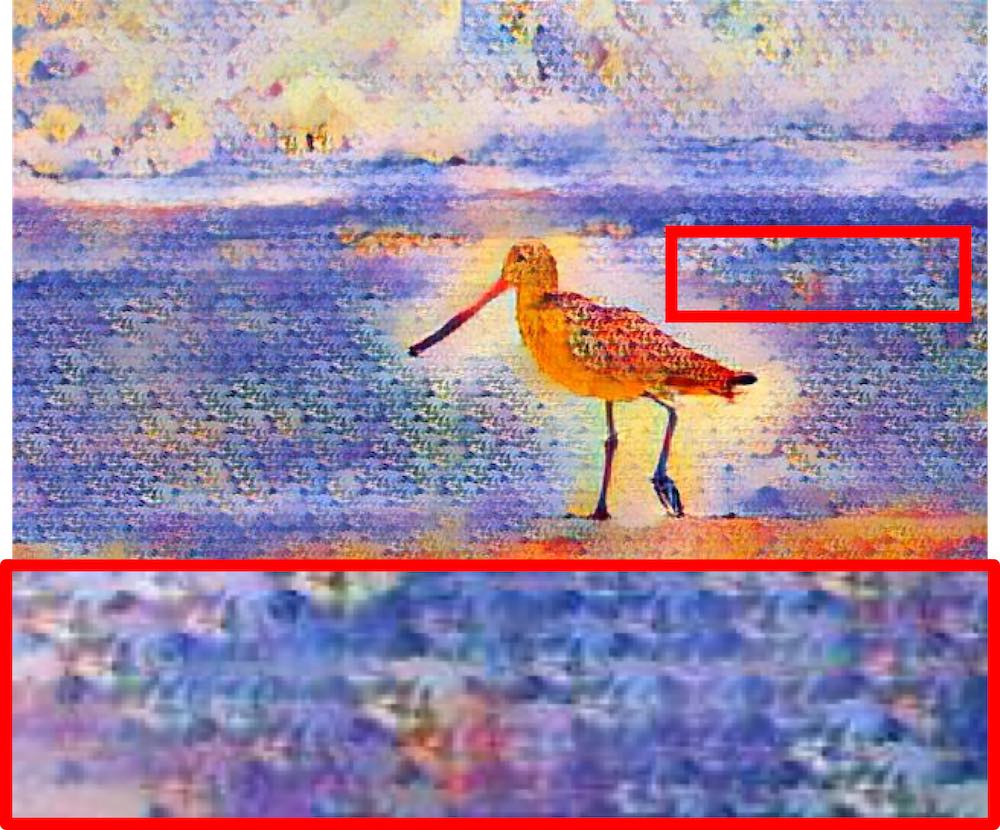}
		\\
		
		\vspace{-0.2em}
		\\
		 (a) Style &  (b) Content &&&  (c) Full Model  & 	 (d) w/o $\mathcal{L}_{adv}$ & 	 (e) w/o $\mathcal{L}_{AR1}$ & 	 (f) w/o $\mathcal{L}_{AR2}$ &&&  (g) Only Stage I  &  (h) Only Stage II  
		\\
		\vspace{-0.2em}
		\\
		\hline
		\vspace{-0.2em}
		\\
		\multicolumn{2}{c}{ CF / (GE + LP) :} &&& 0.420 / 1.524 & 0.352 / 1.501 & 0.389 / 1.482 & 0.436 / 1.438 & && 0.415 / 1.455 &  0.374 / 1.516

		\vspace{0.5em}

	\end{tabular}
	\vspace{-1.5em}
	\caption{ {\bf Ablation study} on loss functions (d-f) and training strategy (g-h). Zoom-in for better comparison. }
	\label{fig:ablation}
	\vspace{-0.6em}
\end{figure*} 

{\bf Qualitative Comparisons.} We first present qualitative comparison results in Fig.~\ref{fig:quality}. On the one hand, for global statistics-based methods (top three rows), AdaIN and WCT are prone to produce messy stylized results with conspicuous artifacts and distorted structures. LST, AAST, and ArtFlow can produce cleaner results with better-preserved content structures. However, there are still evident artifacts and disharmonious patterns (\eg, the mixed colors in row 2 and the spurious shadows in row 3). On the other hand, for local patch-based methods (bottom three rows), SANet and MAST often generate distorted contents (\eg, row 4) or structural artifacts (\eg, rows 5-6), making their results unrealistic. TPFR suffers from the style-uncontrollable problem, where the stylized results deviate from the style references. AdaAttN and IECAST perform well in content preservation but cannot transfer complex textures (\eg, rows 4-5) and may produce blended colors (\eg, row 6).

Through considering the universal human-delightful aesthetics, our AesUST synthesizes aesthetically more harmonious and satisfactory results. The transferred textures are more plausible (\eg, the yellow brushes in row 1, the punctate strokes in row 4, and the line abstraction in row 5), the semantic alignment is more accurate (\eg, the sky in row 3), and the style patterns are more harmonious with fewer artifacts (\eg, rows 2 and 6). It shows that AesUST achieves the best performance in stylization effects, significantly narrowing the disparity with real artist-created paintings.

{\bf Quantitative Comparisons.} We also resort to some quantitative metrics to better evaluate the proposed method.

{\em CF, GE, and LP Scores.} Recently, Wang~\etal~\cite{wang2021evaluate} proposed three quantifiable factors to evaluate the quality of style transfer, \ie, content fidelity (CF), global effects (GE), and local patterns (LP). In detail, CF measures the faithfulness to content characteristics; GE assesses the stylization quality in terms of the global effects like global colors and holistic textures; LP assesses the stylization quality in terms of the similarity and diversity of the local style patterns. All factors are {\em the higher the better}. We collect 5 content images and 10 style images to synthesize 50 stylized images for each method and show their average CF, GE, and LP scores in Tab.~\ref{tab:quantity}. As marked in bold, our AesUST achieves the highest CF score and the third-highest GE+LP score (slightly lower than SANet~\cite{park2019arbitrary} and WCT~\cite{li2017universal}). It signifies that AesUST achieves the best balance between style transformation and content preservation. 

{\em Deception Score.} Like~\cite{kotovenko2019content}, to measure the aestheticism and realism of the fake paintings against the real paintings, we conduct a user study to calculate the deception scores as the fraction of times the synthesized images are guessed as ``real''. We randomly select 20 synthesized images for each method and ask 50 subjects to guess if it is a real painting or not. The results are shown in Tab.~\ref{tab:quantity} (``Deception'' row). For comparison, we also report the deception score of the real artist-created paintings from WikiArt~\cite{phillips2011wiki}. As is evident, the deception score of our AesUST is closest to that of the real artist-created paintings, which further validates the effectiveness of our method.

{\em Preference Score.} We conduct A/B Test user studies to compare the stylization effects of our method with the SOTA methods. We randomly select 100 content-style pairs for each subject. In each pair, except for the content and style images, two stylized results generated by our method and a randomly selected SOTA method are displayed in random order. The subjects are asked to choose their preferred outcomes in terms of content preservation and style transformation. We obtain 5000 votes from 50 subjects and show the percentage of votes in Tab.~\ref{tab:quantity} (``Preference'' row). The results indicate that our AesUST achieves the best stylization effects.

{\em Efficiency.} The bottom row of Tab.~\ref{tab:quantity} shows the run time comparisons on images with a scale of 512$\times$512 pixels. The speed of our AesUST is comparable with SOTA methods such as AdaIN~\cite{huang2017arbitrary}, LST~\cite{li2019learning}, SANet~\cite{park2019arbitrary}, and IECAST~\cite{chen2021artistic}. Thus, it can practicably synthesize stylized images in real time.

\vspace{-0.7em}
\subsection{Ablation Study}
\label{ablation}

{\bf Loss Analyses.} We present ablation study results in Fig.~\ref{fig:ablation} (d-f) to verify the effectiveness of each loss term used for training AesUST. {\bf (1)} Without adversarial loss $\mathcal{L}_{adv}$ (column d), the results exhibit disharmonious patterns and obvious artifacts like the repetitive textures, leading to significant quality degradation. It verifies that adversarial training can help networks synthesize aesthetically more harmonious and realistic results. {\bf (2)} Without the first aesthetic regularization $\mathcal{L}_{AR1}$ (column e), the content preservation and style transformation decline simultaneously. There are some noticeable structural and noisy artifacts, and the style patterns, like the punctate strokes in the bottom row, are not well transferred. {\bf (3)} The second aesthetic regularization $\mathcal{L}_{AR2}$ (column f) can help preserve more pleasing style patterns. Removing it leads to less stylized results. All above analyses are also supported by the quantitative scores below each column of Fig.~\ref{fig:ablation}.

{\bf Two-Stage Transfer Training.} Fig.~\ref{fig:ablation} (g-h) show the results obtained by each training stage, respectively. {\bf (1)} Training with only stage I (column g) produces less satisfactory results where the colors are not vivid enough (\eg, the top row) and the style patterns are implausible (\eg, the punctate strokes in the bottom row). It suggests that the aesthetic guidance and regularizations of stage II are necessary and effective for producing aesthetically more realistic and pleasing results. {\bf (2)} Directly training with only stage II (column h), yet, also cannot produce satisfactory results. It is because the aesthetic features extracted from the discriminator are meaningless at the beginning of the adversarial training, thus deteriorating the style integration and affecting the style transfer training. Therefore, the pre-training of stage I is also necessary and can force networks to transfer style patterns better while maintaining the content structures.

\vspace{-0.6em}
\section{Conclusion}
In this paper, we reveal the aesthetic-unrealistic problem in the SOTA UST algorithms. Upon analyzing the leading cause of this problem, we propose a novel aesthetic-enhanced UST framework, termed AesUST. Our AesUST introduces an aesthetic discriminator and contends two roles for it, \ie, the discriminator and feature extractor, which learns the universal human-delightful aesthetic features from a large corpus of artist-created paintings. Then, a novel AesSA module is introduced to incorporate the aesthetic features with the content and style features, thus achieving more reasonable and flexible style integration. Furthermore, we also develop a new two-stage transfer training strategy with two aesthetic regularizations to train our model more effectively, further improving stylization performance. Experiments validate that our AesUST can synthesize aesthetically more harmonious and realistic results, greatly narrowing the disparity with real artist-created paintings.

\vspace{-0.6em}
\begin{acks}
	This work was supported in part by the National Key Research and Development Program of China (2020YFC1522704), the National Natural Science Foundation of China (62172365),  Zhejiang Elite Program (2022C01222), Key Scientific Research Base for Digital Conservation of Cave Temples of State Administration for Cultural Heritage, and MOE Frontier Science Center for Brain Science \& Brain-Machine Integration (Zhejiang University).
\end{acks}

\bibliographystyle{ACM-Reference-Format}
\balance
\bibliography{acmart}

\clearpage
\nobalance
\appendix

\section{Detailed Architecture of Aesthetic Discriminator}
\label{arch}
The detailed architecture of our aesthetic discriminator is summarized in Tab.~\ref{tab:arch}, which is adapted from~\cite{wang2018high}.

\newcommand{\tabincell}[2]{\begin{tabular}{@{}#1@{}}#2\end{tabular}}
\renewcommand\arraystretch{1.3}
\begin{table*}[h]
	\centering
	\setlength{\tabcolsep}{0.26cm}
	\caption{Architecture of aesthetic discriminator.}
	\vspace{-0.5em}
	\begin{tabular}{c|ccccccc}
		\hline
		\hline
		Part & Layer & Kernel Size & Stride & Padding & In Channel & Out Channel & Negative Slope
		\\
		\hline
		\multirow{11}{*}{$E_1$ / $E_2$ / $E_3$} & Conv & 4 &  2 & 1 & 3 & 64 & -
		\\
		& LeakyReLU & - &  - & - & - & - & 0.2
		\\
		
		& Conv & 4 &  2 & 1 & 64 & 128 & -
		\\
		& InstanceNorm & - & - & - & 128 & 128 & -
		\\
		& LeakyReLU & - &  - & - & - & - & 0.2
		\\
		
		& Conv & 4 &  2 & 1 & 128 & 256 & -
		\\
		& InstanceNorm & - & - & - & 256 & 256 & -
		\\
		& LeakyReLU & - &  - & - & - & - & 0.2
		\\
		
		& Conv & 4 &  2 & 1 & 256 & 512 & -
		\\
		& InstanceNorm & - & - & - & 512 & 512 & -
		\\
		& LeakyReLU & - &  - & - & - & - & 0.2
		\\
		
		\hline
		$\mathcal{C}_1$ / $\mathcal{C}_2$ / $\mathcal{C}_3$ & Conv & 3 &  1 & 1 & 512 & 1 & -
		\\
		\hline
		Downsample & Avg Pool & 3 & 2 & 1 & - & - & -
		\\
		\hline
		Upsample$^2$ & Nearest & - & 1/2 & - & - & - & -
		\\
		Upsample$^4$ & Nearest & - & 1/4 & - & - & - & -
		\\
		\hline
		\hline
		
	\end{tabular}
	\label{tab:arch}
\end{table*}

\section{Runtime Controls}
\label{control}

Similar to~\cite{huang2017arbitrary,park2019arbitrary}, our AesUST also allows users to control the degree of stylization, interpolate between different styles, transfer styles while preserving colors, and use different styles in different spatial regions. Note that all these controls are only applied at runtime using the same model without re-training.

\subsection{Content-style trade-off}
\begin{figure}[h]
	\centering
	\setlength{\tabcolsep}{0.03cm}
	\begin{tabular}{cccccc}
		\includegraphics[width=0.16\linewidth]{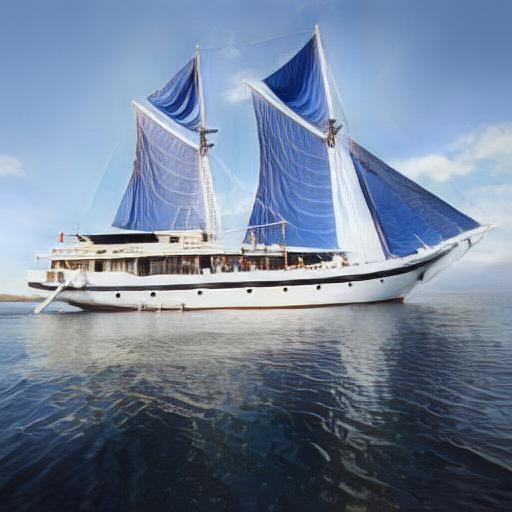}&
		\includegraphics[width=0.16\linewidth]{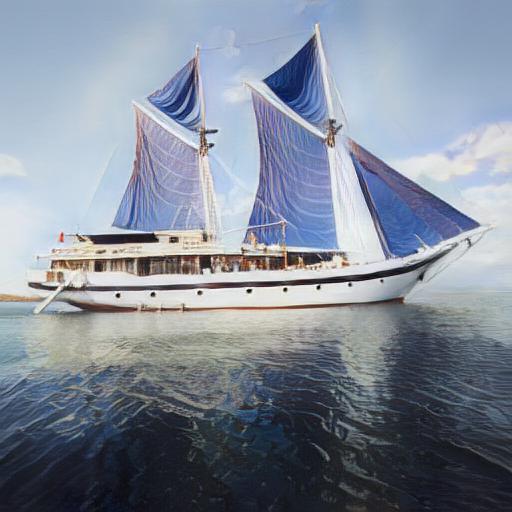}&
		\includegraphics[width=0.16\linewidth]{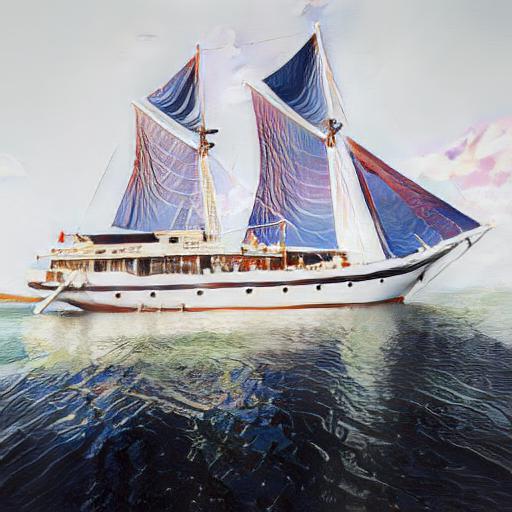}&
		\includegraphics[width=0.16\linewidth]{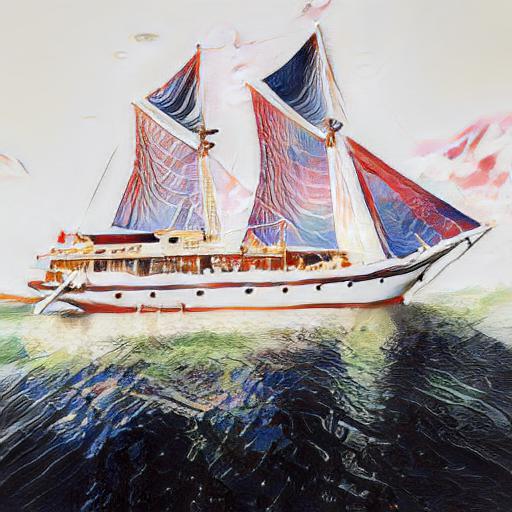}&
		\includegraphics[width=0.16\linewidth]{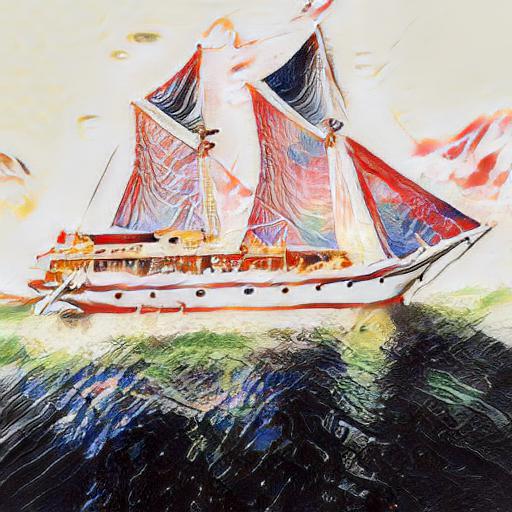}&
		\includegraphics[width=0.16\linewidth]{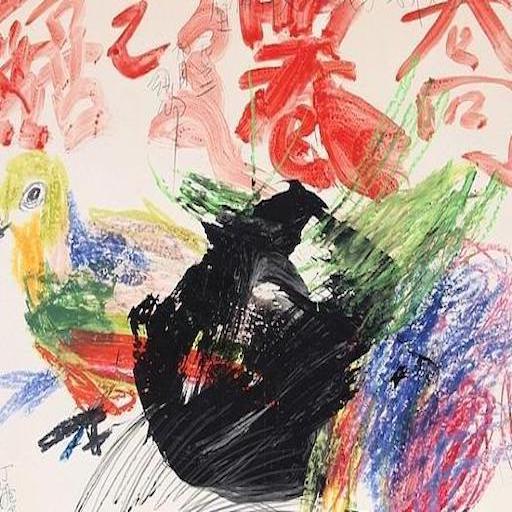}
		
		\\
		$\alpha=0$ & $\alpha=0.25$ & $\alpha=0.5$ &  $\alpha=0.75$ &  $\alpha=1$ &  Style
	\end{tabular}
    \vspace{-0.5em}
	\caption{ Content–style trade-off during runtime. 
	}
	\label{fig:style-content1}
	\vspace{-0.5em}
\end{figure}

We can control the degree of stylization by adjusting the style weights $\lambda_3$ in Eq. (\ref{eq12}) and $\lambda_7$ in Eq. (\ref{eq17}) in our main paper during {\bf train} time, or interpolating between feature maps that are fed into the decoder during {\bf test} time.  We empirically determine the $\lambda$-values in Eq. (\ref{eq12}) and Eq. (\ref{eq17}) following the previous experience of~\cite{huang2017arbitrary,park2019arbitrary}. Users can adjust them to control the degree of stylization but need to re-train the model, which is inconvenient. In order to achieve runtime control, like~\cite{huang2017arbitrary,park2019arbitrary}, we introduce $\alpha\in [0,1]$ to control the degree of stylization during test time. In detail, given the content image $I_c$ and style image $I_s$, we first obtain the stylized feature $F_{cs}$ by taking the content image as content input and style image as style input for our model. Then we obtain the feature $F_{cc}$ by taking the content image as both content and style inputs for our model. The $\alpha$ value is used to control the blending of $F_{cs}$ and $F_{cc}$, obtaining the new stylized feature $F_{cs}’= \alpha F_{cs}+(1-\alpha) F_{cc}$. The new stylized feature $F_{cs}’$ is fed to the decoder to obtain the final result. In this way, users can control the degree of stylization without re-training the network. The network tries to reconstruct the content image when $\alpha=0$, and to generate the most stylized image when $\alpha=1$, as shown in Fig.~\ref{fig:style-content1}.

\renewcommand\arraystretch{0.6}

\subsection{Style interpolation}
\begin{figure}[t]
	\centering
	\setlength{\tabcolsep}{0cm}
	\begin{tabular}{ccccccc}
		\includegraphics[width=0.14\linewidth]{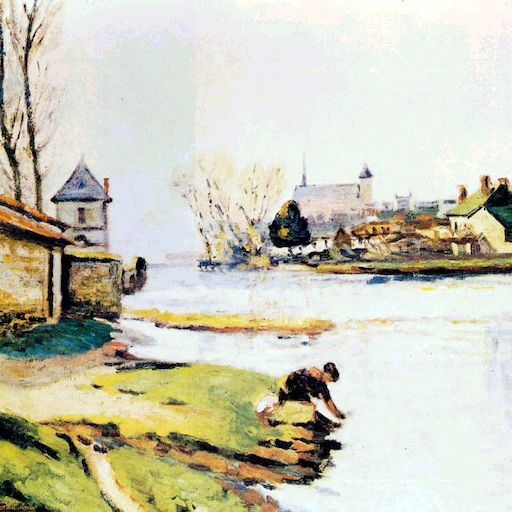}&
		\includegraphics[width=0.14\linewidth]{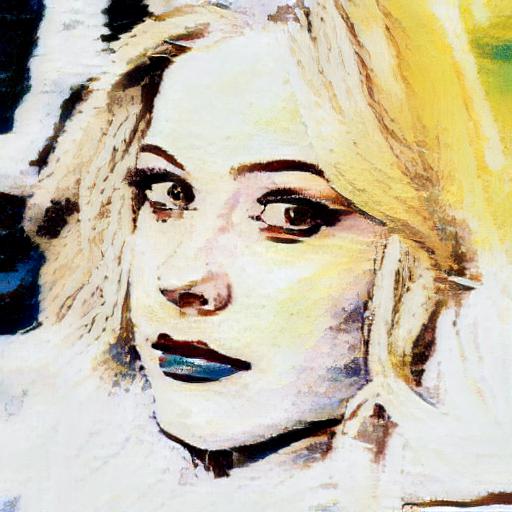}&
		\includegraphics[width=0.14\linewidth]{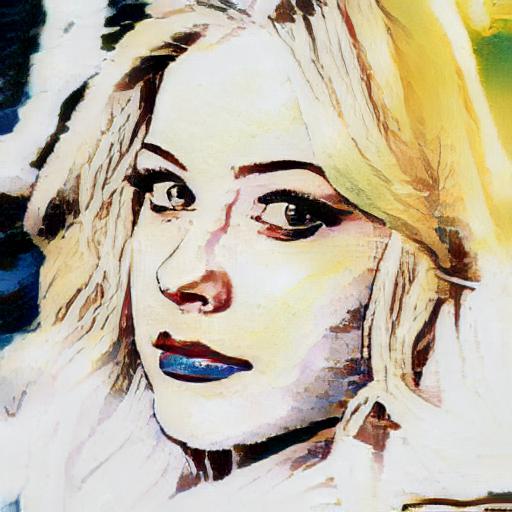}&
		\includegraphics[width=0.14\linewidth]{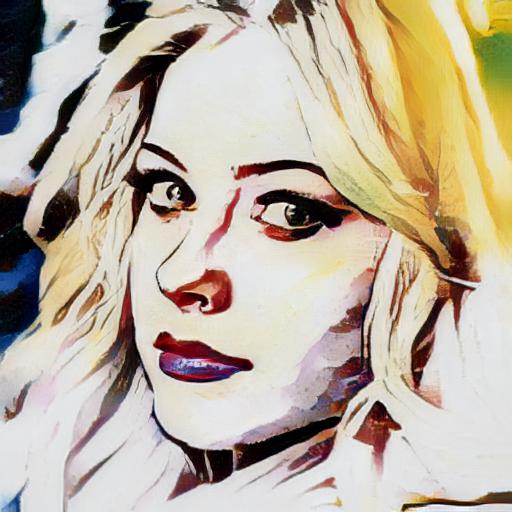}&
		\includegraphics[width=0.14\linewidth]{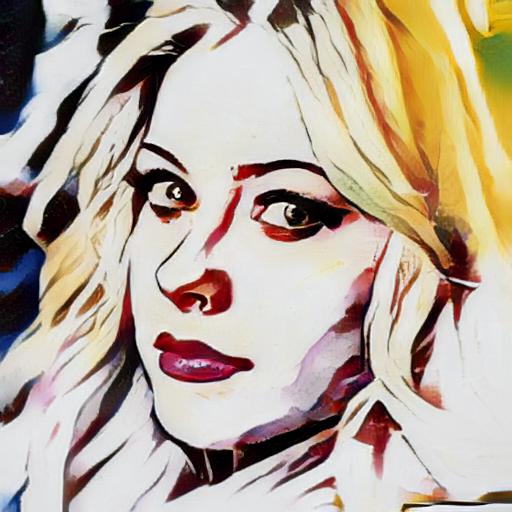}&
		\includegraphics[width=0.14\linewidth]{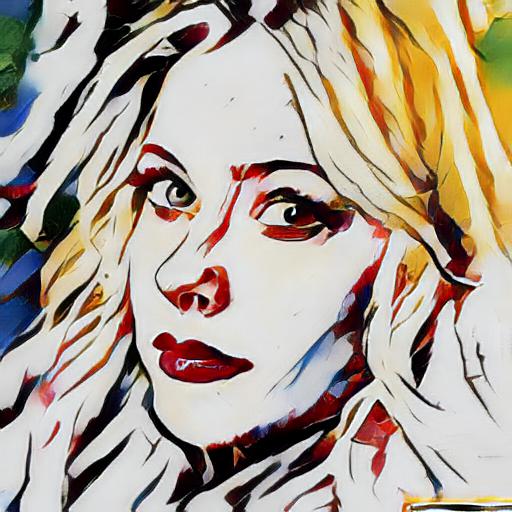}&
		\includegraphics[width=0.14\linewidth]{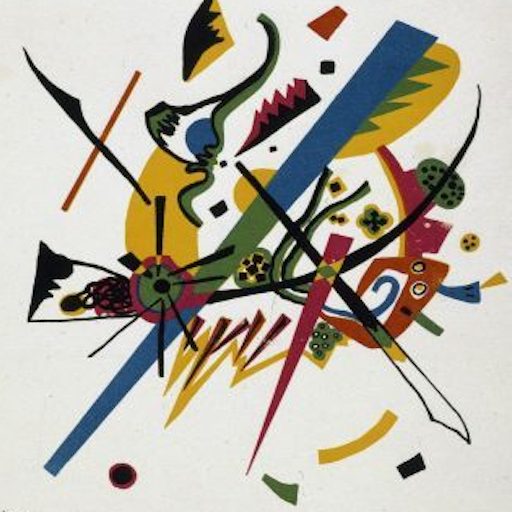}
		
		\\
		&
		\includegraphics[width=0.14\linewidth]{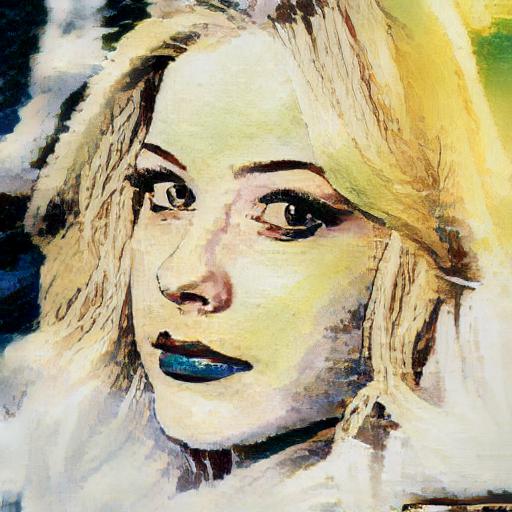}&
		\includegraphics[width=0.14\linewidth]{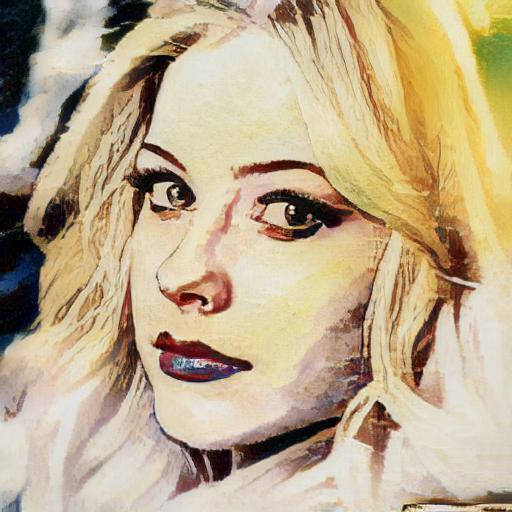}&
		\includegraphics[width=0.14\linewidth]{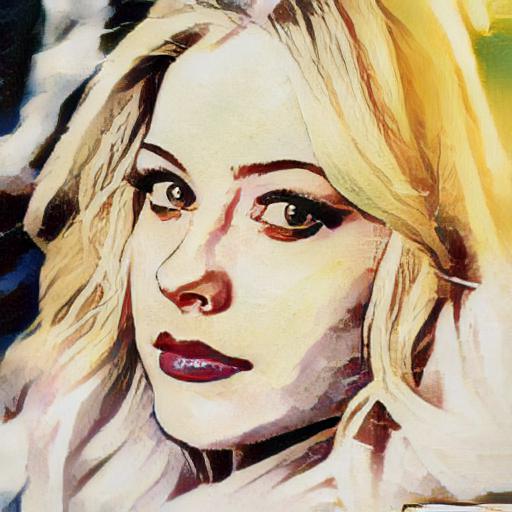}&
		\includegraphics[width=0.14\linewidth]{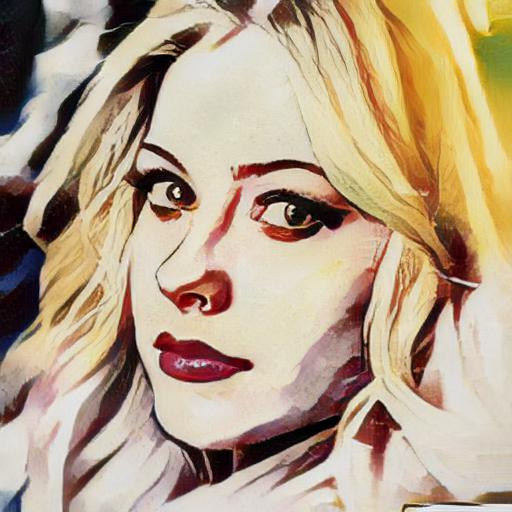}&
		\includegraphics[width=0.14\linewidth]{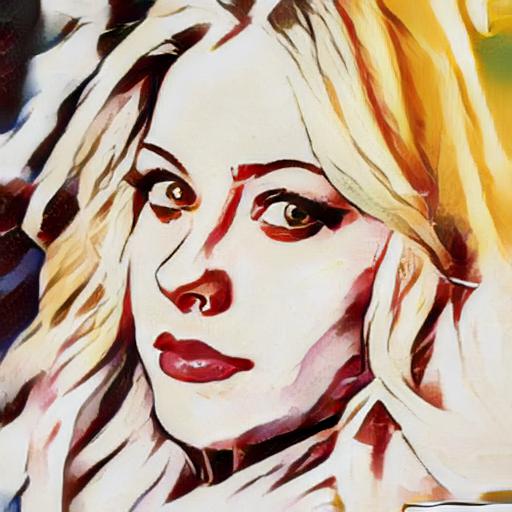}
		
		\\
		&
		\includegraphics[width=0.14\linewidth]{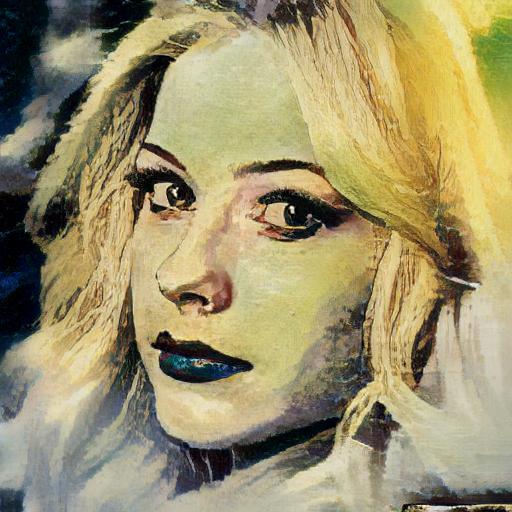}&
		\includegraphics[width=0.14\linewidth]{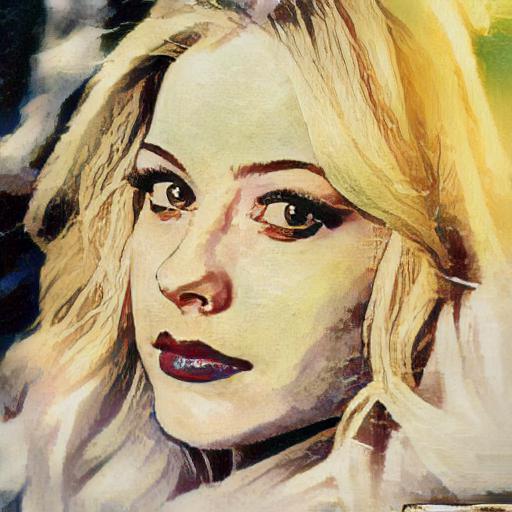}&
		\includegraphics[width=0.14\linewidth]{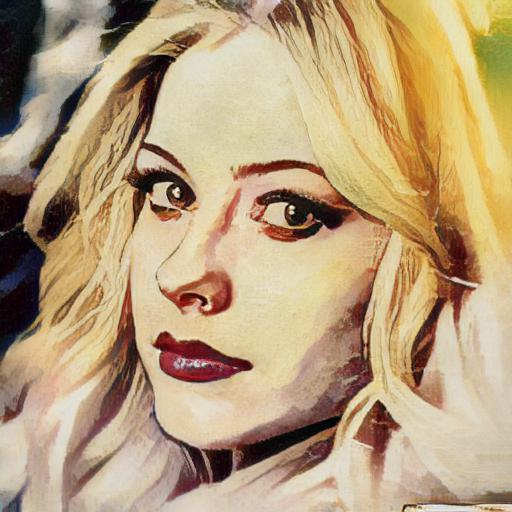}&
		\includegraphics[width=0.14\linewidth]{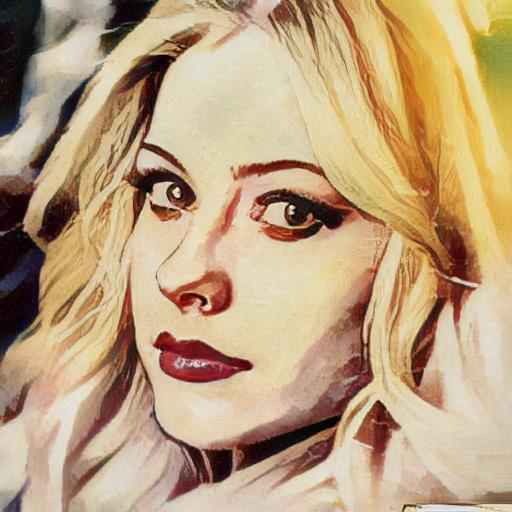}&
		\includegraphics[width=0.14\linewidth]{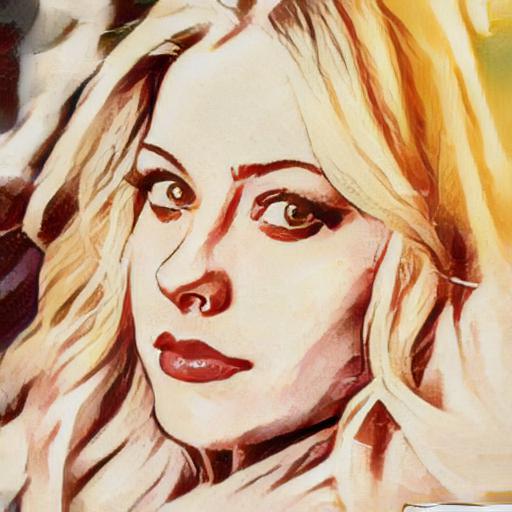}
		
		\\
		&
		\includegraphics[width=0.14\linewidth]{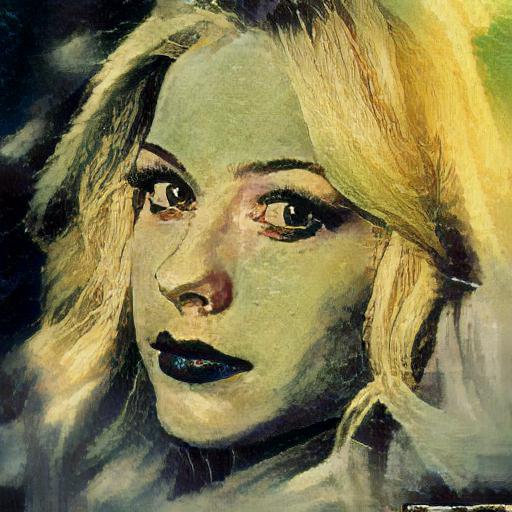}&
		\includegraphics[width=0.14\linewidth]{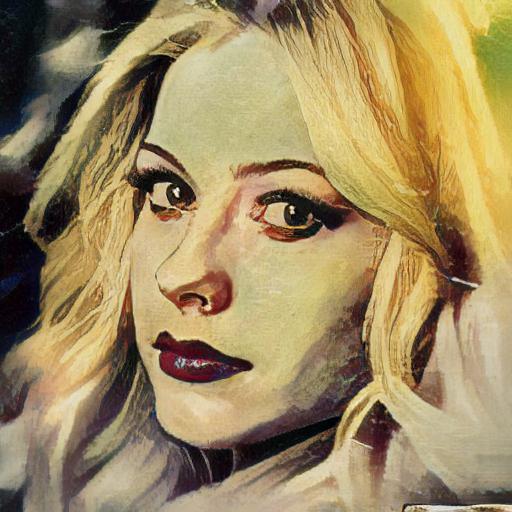}&
		\includegraphics[width=0.14\linewidth]{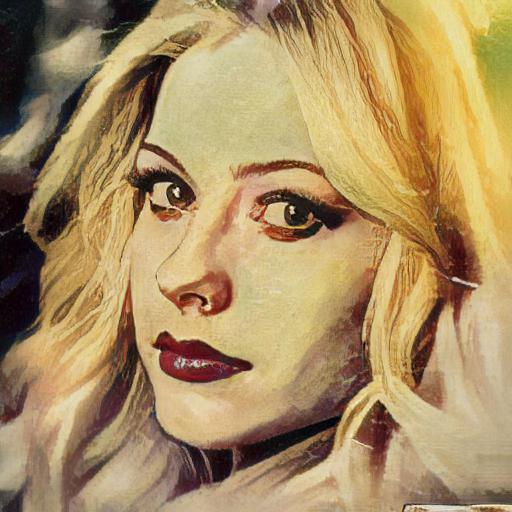}&
		\includegraphics[width=0.14\linewidth]{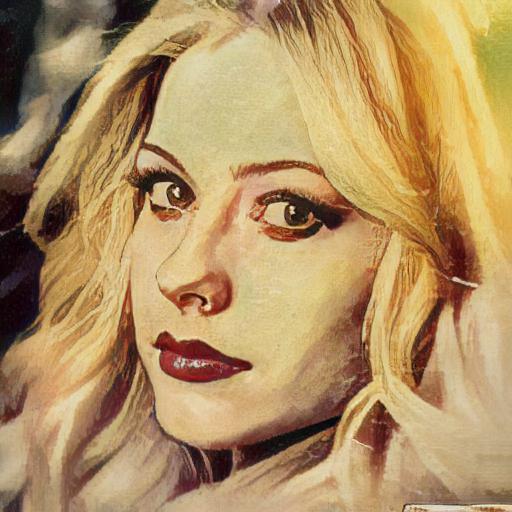}&
		\includegraphics[width=0.14\linewidth]{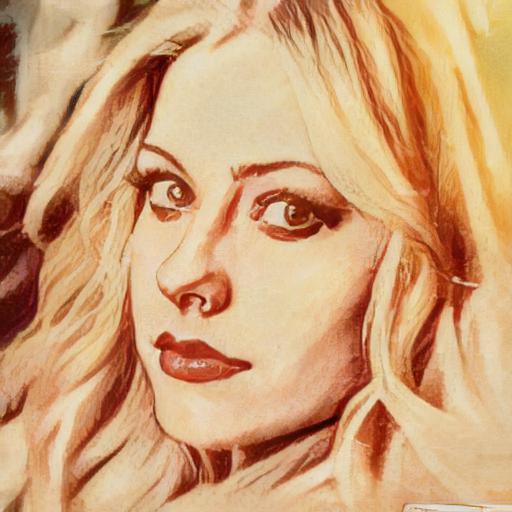}
		
		\\
		\includegraphics[width=0.14\linewidth]{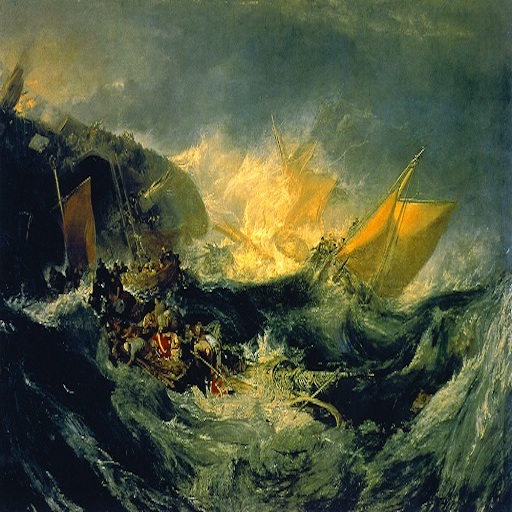}&
		\includegraphics[width=0.14\linewidth]{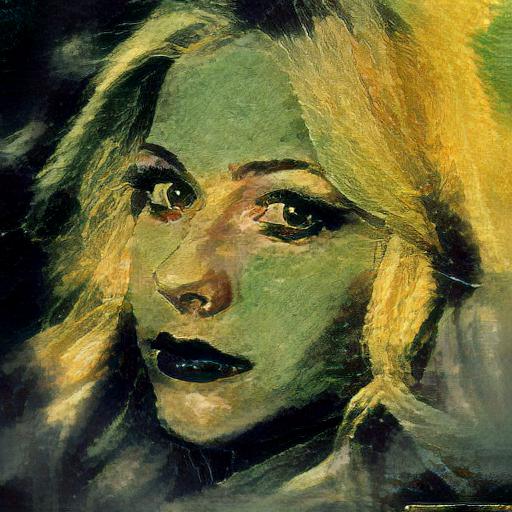}&
		\includegraphics[width=0.14\linewidth]{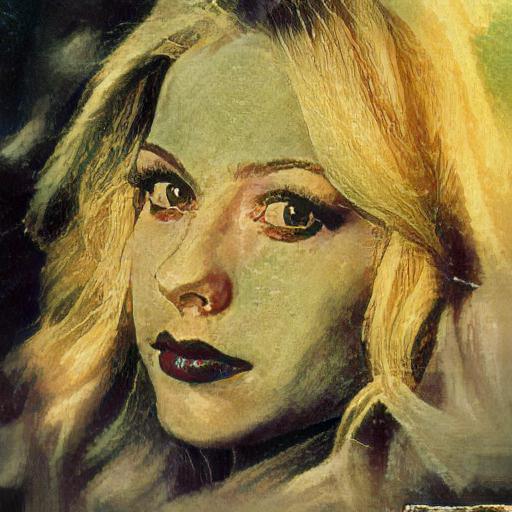}&
		\includegraphics[width=0.14\linewidth]{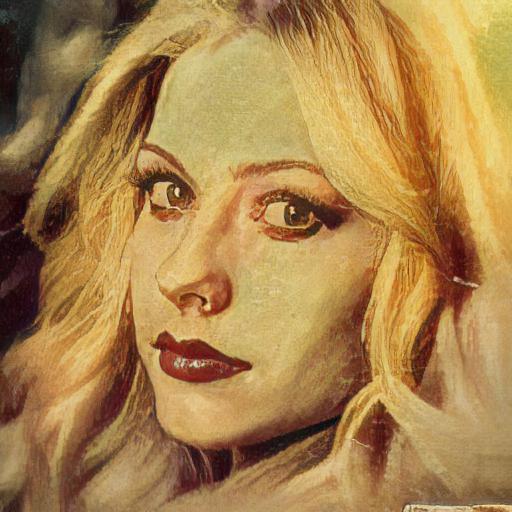}&
		\includegraphics[width=0.14\linewidth]{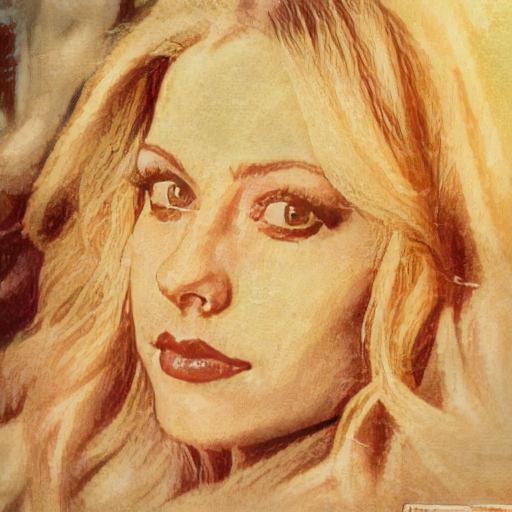}&
		\includegraphics[width=0.14\linewidth]{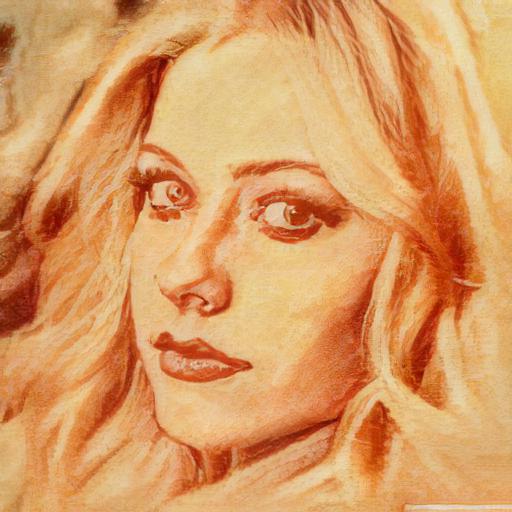}&
		\includegraphics[width=0.14\linewidth]{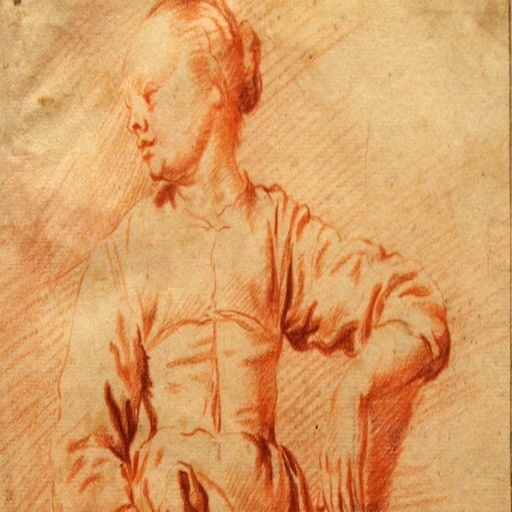}

	\end{tabular}
    \vspace{-0.5em}
	\caption{ Style interpolation between four different styles.
	}
	\label{fig:style-interpolation}
	\vspace{-0.5em}
\end{figure}

We can interpolate between several style images by blending feature maps $F_{cs}$ transferred from different styles and then feeding the blended feature into the decoder. An example of style interpolation between four different styles is shown in Fig.~\ref{fig:style-interpolation}.

\renewcommand\arraystretch{0}

\subsection{Color and spatial control}
\begin{figure}[t]
	\centering
	\setlength{\tabcolsep}{0.03cm}
	\begin{tabular}{ccccc}
		\includegraphics[width=0.195\linewidth]{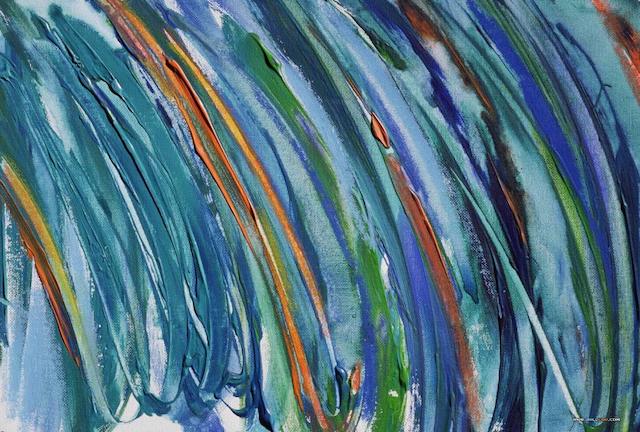}&
		\includegraphics[width=0.195\linewidth]{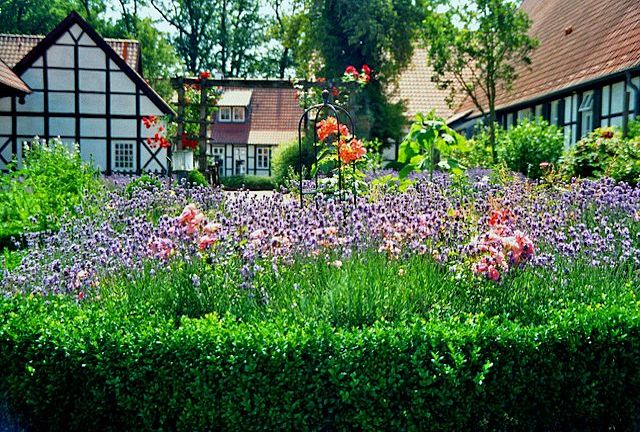}&
		\includegraphics[width=0.195\linewidth]{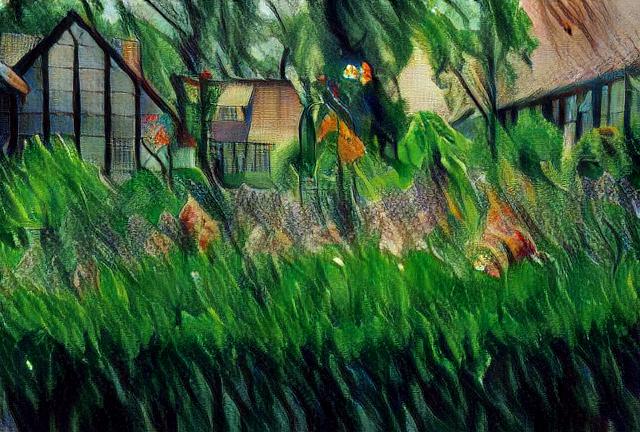}&
		\includegraphics[width=0.195\linewidth]{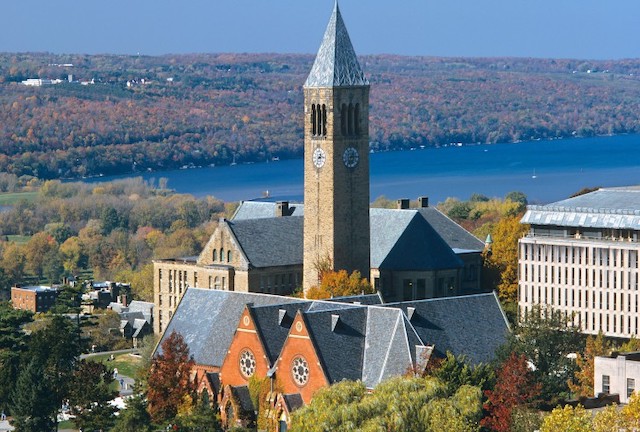}&
		\includegraphics[width=0.195\linewidth]{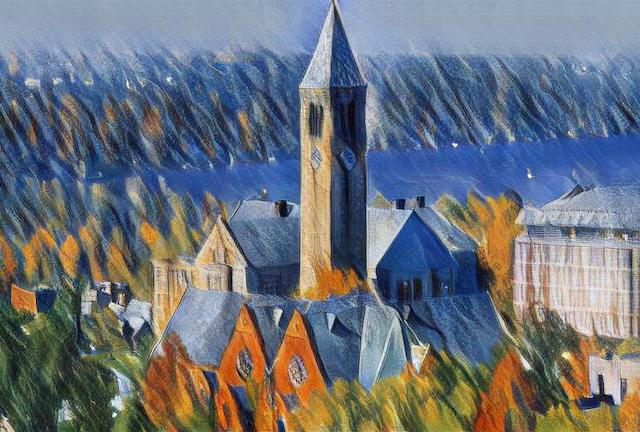}
		\\
		Style & Content 1 & Result 1 &  Content 2 &  Result 2
	\end{tabular}
    \vspace{-0.5em}
	\caption{ Color control. We show color-preserved style transfer results.
	}
	\label{fig:color-preserve}
	\vspace{-0.5em}
\end{figure}

Our method can easily achieve color-preserved style transfer similar to~\cite{gatys2017controlling,huang2017arbitrary}. The color distribution of the style image is first manipulated to match that of the content image, and then we perform the style transfer using the color-aligned style image as the style input. Examples are shown in Fig.~\ref{fig:color-preserve}.

\renewcommand\arraystretch{0.6}

\begin{figure}[t]
	\centering
	
	\includegraphics[width=1\linewidth]{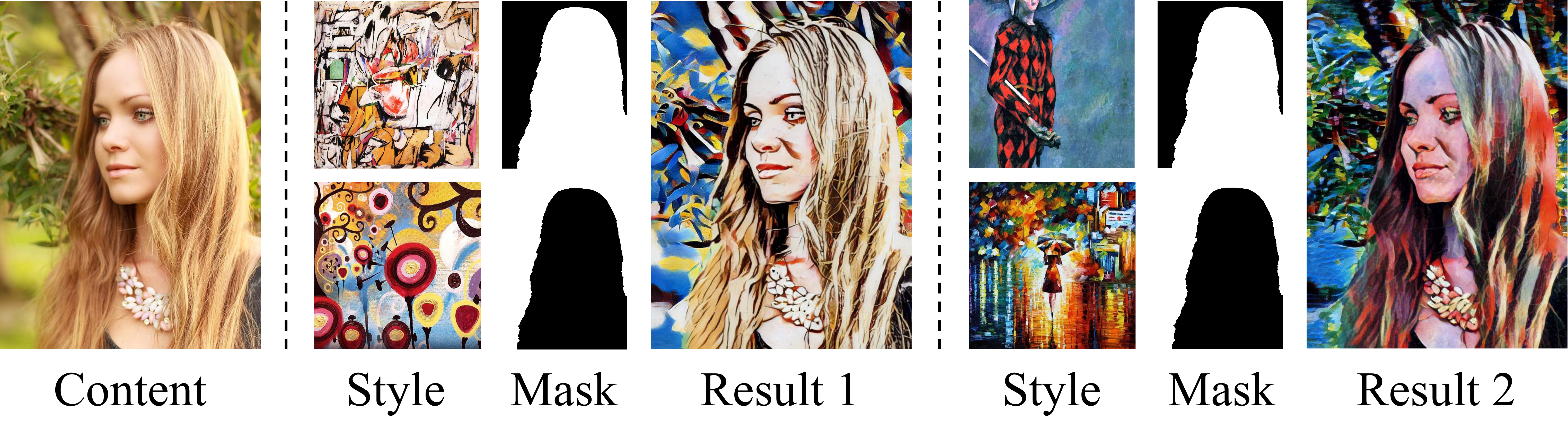}
	\vspace{-2em}
	\caption{ Spatial control examples.
	}
	\label{fig:spatial}
	\vspace{-0.5em}
\end{figure}

Fig.~\ref{fig:spatial} demonstrates that our method can transfer different regions of the content image to different styles. We additionally input a set of masks to map the spatial correspondence between content regions and styles. A simple mask-out operation is applied to assign the different styles in each spatial region.

\renewcommand\arraystretch{0.6}

\section{Limitations and Discussions}
\label{limit}

In this section, we provide some typical limitations of our method, then analyze the reasons behind them, and finally discuss the possible solutions to address them. Given the simplicity of our proposed framework, we believe there is substantial room for improvement. The further improvement of our approach we leave as future work.

\subsection{Mixed textures}
\begin{figure}[t]
	\centering
	\setlength{\tabcolsep}{0.03cm}
	\begin{tabular}{ccccc}
		\includegraphics[width=0.195\linewidth]{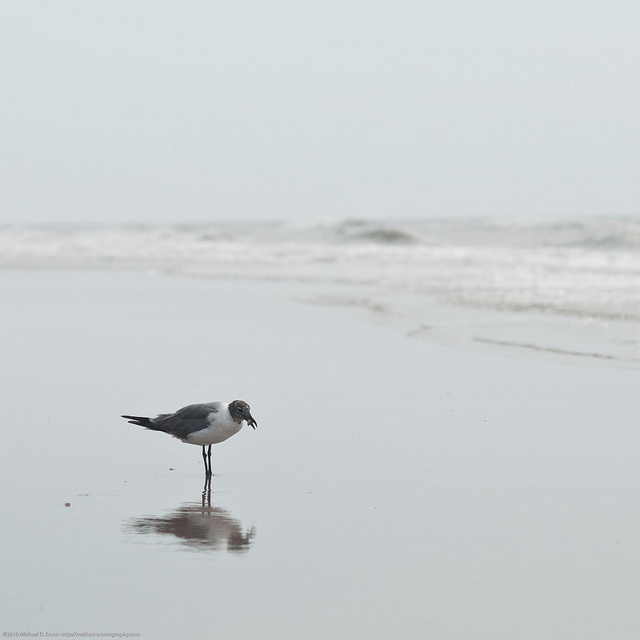}&
		\includegraphics[width=0.195\linewidth]{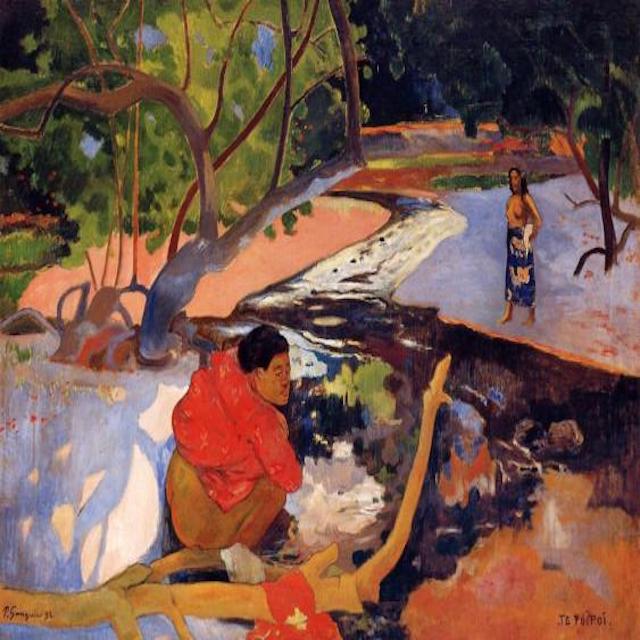}&
		\includegraphics[width=0.195\linewidth]{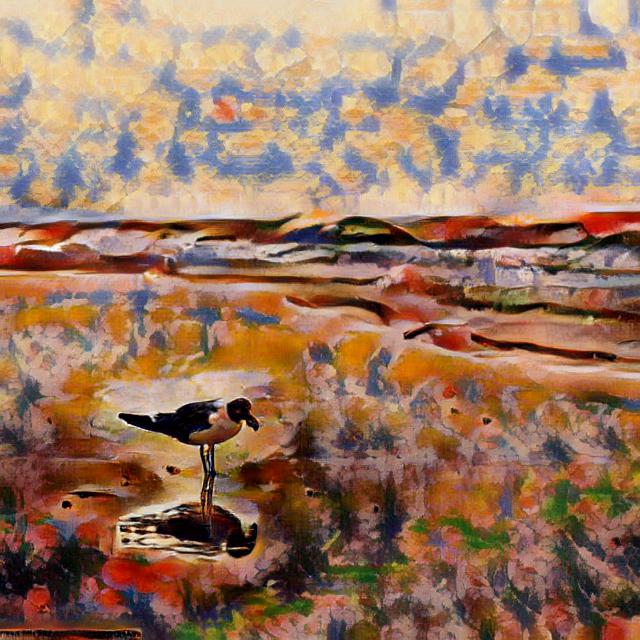}&
		\includegraphics[width=0.195\linewidth]{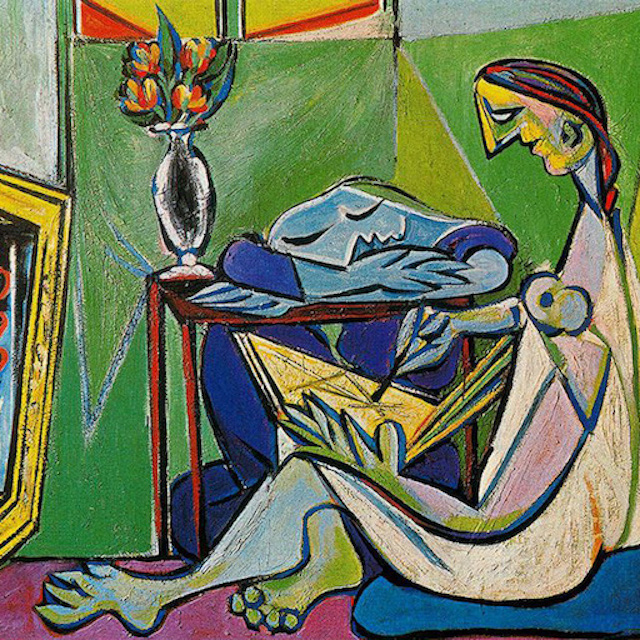}&
		\includegraphics[width=0.195\linewidth]{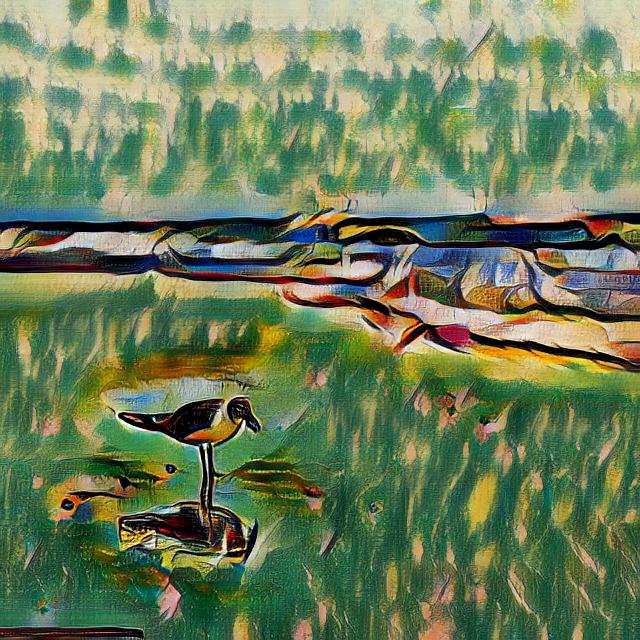}
		
		\\
		Content & Style 1 & Failure 1 &  Style 2 &  Failure 2
	\end{tabular}
    \vspace{-0.5em}
	\caption{ Failure cases of type 1. Our method may produce mixed and messy textures for content areas with low feature activations, like the textureless background.
	}
	\label{fig:failure1}
	\vspace{-0.5em}
\end{figure}

As pointed out in our main paper, our method may produce mixed and messy textures for content areas with low feature activations, like the textureless background shown in Fig.~\ref{fig:failure1}. It can be attributed to the fact that the textureless areas are hard to recognize (with low feature activations) for a pre-trained VGG network~\cite{liao2017visual}; therefore, the style attention~\cite{park2019arbitrary} operation in our AesSA module fails to find the correspondence between these areas and those of the style images. This issue may be addressed by incorporating user guidance like~\cite{champandard2016semantic,wang2022texture} or using some texture enhancement operations to highlight the subtle textures in these areas. Moreover, one may also consider improving the style attention~\cite{park2019arbitrary} operation in our AesSA module to solve this problem.

\renewcommand\arraystretch{0.6}

\subsection{Generalization to out-of-distribution styles}

\begin{figure}[h]
	\centering
	\setlength{\tabcolsep}{0.03cm}
	\begin{tabular}{ccccc}
		\includegraphics[width=0.195\linewidth]{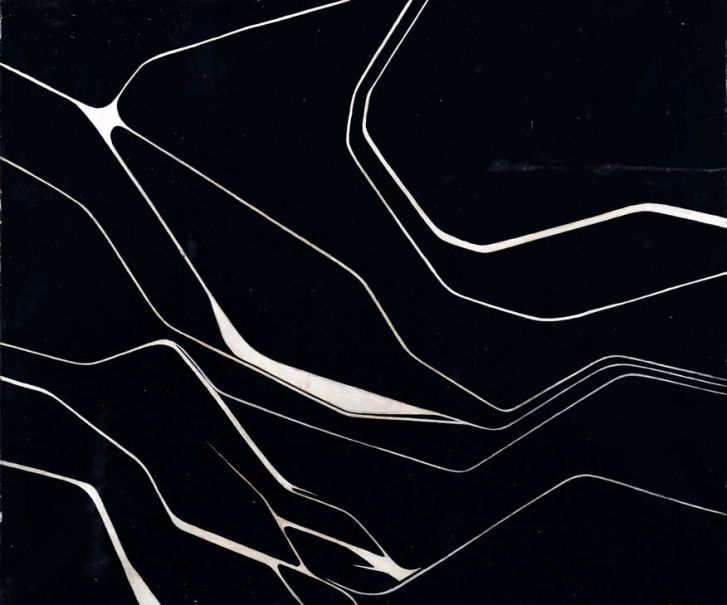}&
		\includegraphics[width=0.195\linewidth]{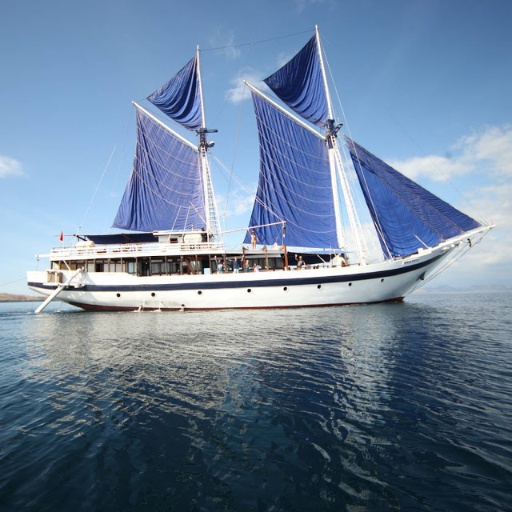}&
		\includegraphics[width=0.195\linewidth]{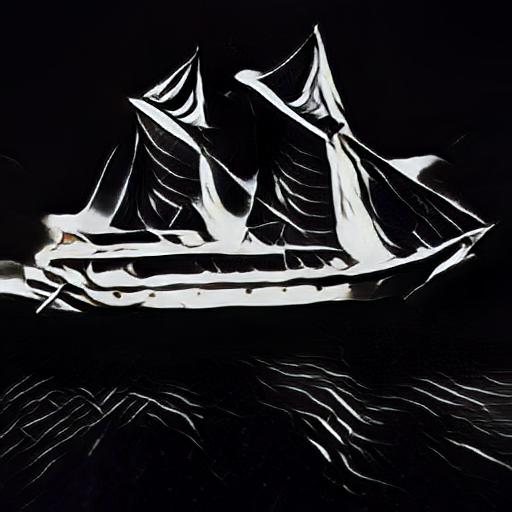}&
		\includegraphics[width=0.195\linewidth]{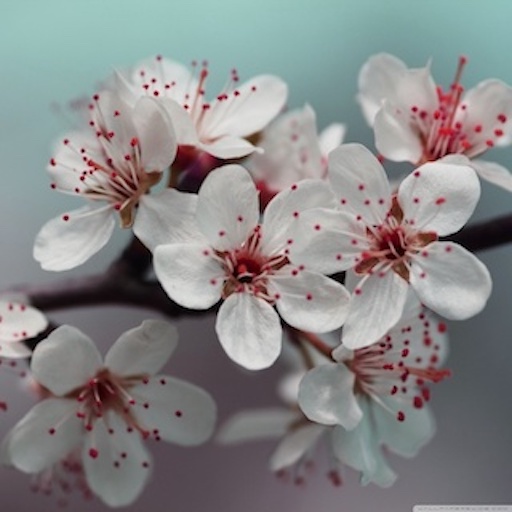}&
		\includegraphics[width=0.195\linewidth]{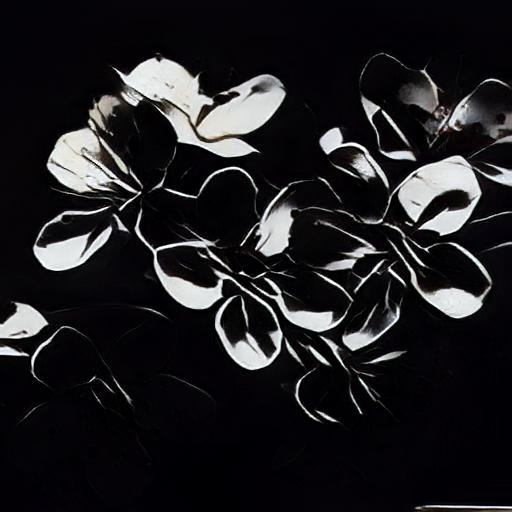}
		
		\\
		Style & Content 1 & Failure 1 &  Content 2 &  Failure 2
	\end{tabular}
    \vspace{-0.5em}
	\caption{ Failure cases of type 2. Our method may not generalize very well to the style images that excessively deviate from the training distribution, like the simple line pattern.
	}
	\label{fig:failure2}
	\vspace{-0.5em}
\end{figure}

Due to the use of GAN loss, our method may not generalize very well to the style images that excessively deviate from the training distribution, \eg, the simple line pattern shown in Fig.~\ref{fig:failure2}. It may be addressed by collecting a more exhaustive training dataset or using some incremental training strategies.

\renewcommand\arraystretch{0.6}

\subsection{Color blending}

\begin{figure}[h]
	\centering
	\setlength{\tabcolsep}{0.03cm}
	\begin{tabular}{cccc}
		\includegraphics[width=0.25\linewidth]{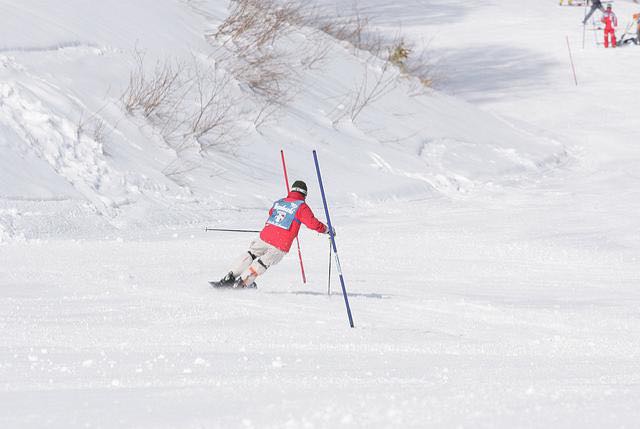}&
		\includegraphics[width=0.207\linewidth]{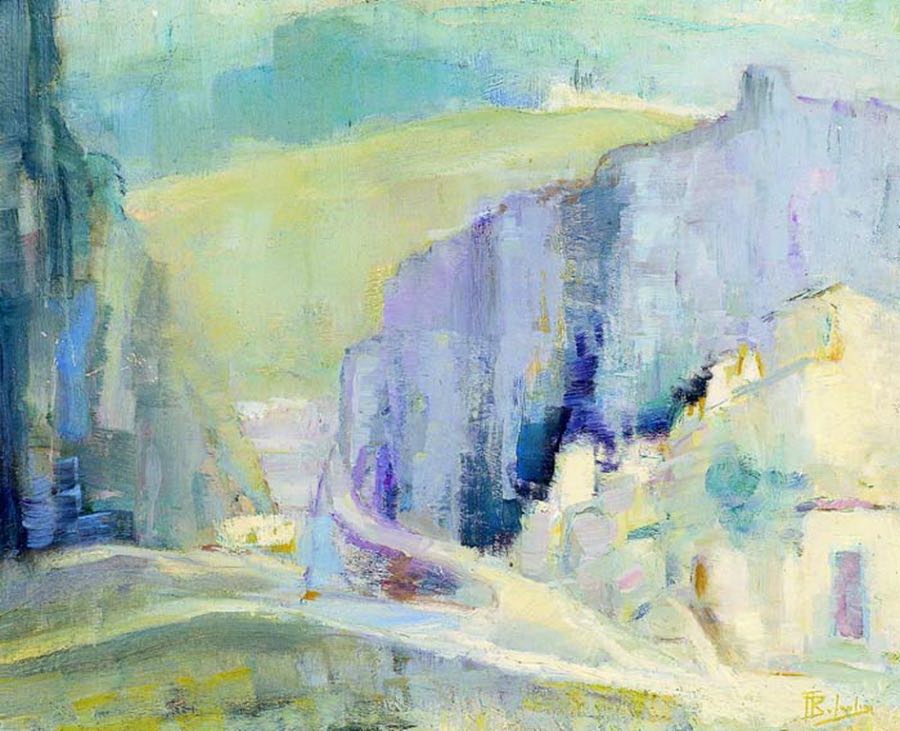}&
		\includegraphics[width=0.25\linewidth]{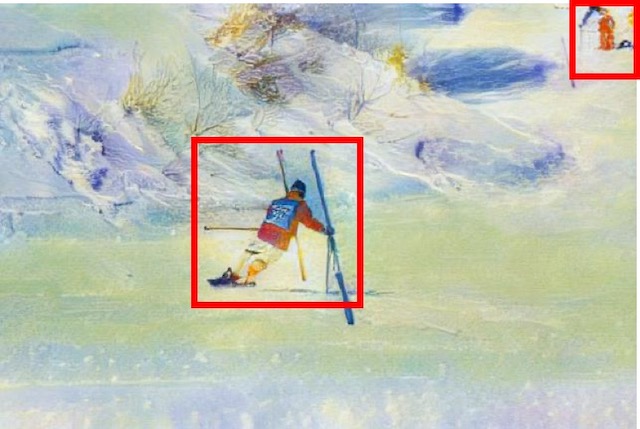}&
		\includegraphics[width=0.25\linewidth]{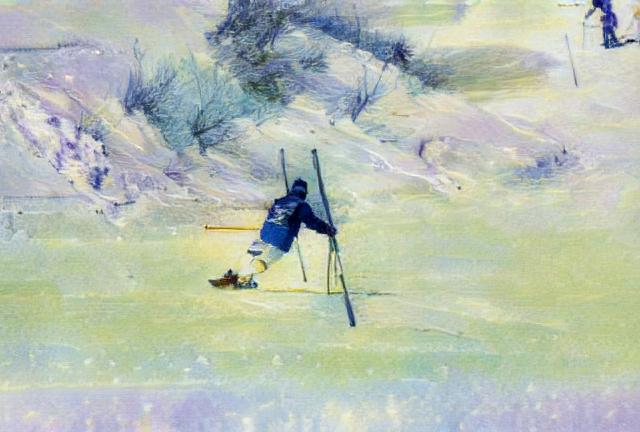}
		
		\\
		Content & Style & Failure  & Remove Color
	\end{tabular}
    \vspace{-0.5em}
	\caption{ Failure cases of type 3. Our method may produce slight color blending for a few cases, like the colors of the clothes (marked by red boxes) in the stylized result.
	}
	\label{fig:failure31}
	\vspace{-0.5em}
\end{figure}

Our method may produce slight color blending for a few cases. As shown in Fig.~\ref{fig:failure31}, the colors of the clothes (marked by red boxes in column 3) seem to be blends of those in the content image and style image. We analyze the reasons may be that the colors in the content image are too vivid compared to those in the style image, thus hard to change via style transfer. The issues can be addressed by removing the colors of the content image as shown in the last column, or training our network with higher style weights $\lambda_3$ in Eq. (\ref{eq12}) and $\lambda_7$ in Eq. (\ref{eq17}) in the main paper.

\renewcommand\arraystretch{0.6}

\end{document}